\documentclass{article}
\usepackage{PRIMEarxiv}
\usepackage[utf8]{inputenc} 
\usepackage[T1]{fontenc}    
\usepackage[utf8]{vietnam}
\usepackage[english]{babel}
\usepackage{tipa}
\usepackage{ragged2e}

\usepackage{hyperref}       
\usepackage{url}            
\usepackage{booktabs}       
\usepackage{amsfonts}       
\usepackage{amsmath}
\usepackage{nicefrac}       
\usepackage{microtype}      
\usepackage{lipsum}
\usepackage{graphicx}
\graphicspath{{figure/}}     
\usepackage{caption}
\usepackage{subcaption}
\usepackage{multirow}
\usepackage{booktabs}

\usepackage[final]{pdfpages}
\usepackage{tabularx}
\usepackage{tkz-tab}
\usepackage{framed,multicol}
\usepackage[framemethod=tikz]{mdframed}
\usepackage{xcolor}

\usepackage{wasysym}

\usepackage{pgfplots}
\pgfplotsset{compat=1.3}
\usetikzlibrary{positioning, fit, calc}  
\tikzset{block/.style={draw, thick, text width=2cm ,minimum height=1.3cm, align=center},   
	line/.style={-latex}     
} 

\tikzset{blocktext/.style={draw, thick, text width=5.2cm ,minimum height=1.3cm, align=center},   
	line/.style={-latex}     
}

\tikzset{font=\footnotesize}

\title{VNHSGE: VietNamese High School Graduation Examination Dataset for Large Language Models
}

\author{ \href{https://orcid.org/0000-0001-5415-7538}{\includegraphics[scale=0.06]{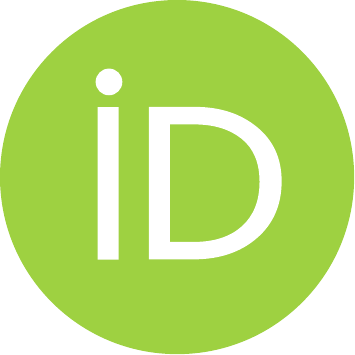}\hspace{1mm}Xuan-Quy Dao$^{1,\ast}$} \\
	\And
	\href{https://orcid.org/0000-0001-7431-0157}{\includegraphics[scale=0.06]{orcid.pdf}\hspace{1mm}Ngoc-Bich Le$^{2,\star}$}\\
	\And
	\href{https://orcid.org/0009-0005-0237-0878}{\includegraphics[scale=0.06]{orcid.pdf}\hspace{1mm}The-Duy Vo$^{1,\ast}$}\\
	\And
	Xuan-Dung Phan$^{1,\ast}$\\
	\AND
	Bac-Bien Ngo$^{1,\dagger}$\\
	\And
	Van-Tien Nguyen$^{1,\ast}$\\
	\And
	Thi-My-Thanh Nguyen$^{1,\ast}$\\
	\And
	Hong-Phuoc Nguyen$^{1,\ast}$\\
	\And
	\texttt{$^{1}$Eastern International University,~$^{2}$International University}\\
	$^{\ast}$\{quy.dao,~duy.vo,~dung.phan,~tien.nguyen,~thimythanh.nguyen,~phuoc.nguyen\}@eiu.edu.vn \\
	$^{\star}$lnbich@hcmiu.edu.vn, $^{\dagger}$ngobacbienspk@gmail.com 
}

\begin{document}
	\maketitle

	\begin{abstract}
		
		The VNHSGE (VietNamese High School Graduation Examination) dataset, developed exclusively for evaluating large language models (LLMs), is introduced in this article. The dataset, which covers nine subjects, was generated from the Vietnamese National High School Graduation Examination and comparable tests. 300 literary essays have been included, and there are over 19,000 multiple-choice questions on a range of topics. The dataset assesses LLMs in multitasking situations such as question answering, text generation, reading comprehension, visual question answering, and more by including both textual data and accompanying images. Using ChatGPT and BingChat, we evaluated LLMs on the VNHSGE dataset and contrasted their performance with that of Vietnamese students to see how well they performed. The results show that ChatGPT and BingChat both perform at a human level in a number of areas, including literature, English, history, geography, and civics education. They still have space to grow, though, especially in the areas of mathematics, physics, chemistry, and biology. The VNHSGE dataset seeks to provide an adequate benchmark for assessing the abilities of LLMs with its wide-ranging coverage and variety of activities. We intend to promote future developments in the creation of LLMs by making this dataset available to the scientific community, especially in resolving LLMs' limits in disciplines involving mathematics and the natural sciences. 
		
	\end{abstract}

	\section{Introduction}
	
	Artificial intelligence (AI) has the potential to revolutionize the educational system. According to Chassignol et al.~\cite{chassignol2018artificial}, four areas—customized educational content, cutting-edge teaching strategies, technology-enhanced evaluation, and communication between students and teachers—are where AI can revolutionize the educational environment. An overview of AI applications in higher education has been offered by Zawacki-Richter et al.~\cite{zawacki2019systematic}, spanning profiling and prediction, evaluation and assessment, adaptive systems and personalization, and intelligent tutoring systems. Potential research subjects in AI applications for education have been suggested by Hwang et al.~\cite{hwang2020vision}. In order to enable effective administrative operations, content modification, and enhanced learning quality, Chen et al.~\cite{chen2020artificial} have concentrated on the use of AI in administration, instruction, and learning. The potential of generative AI in education to lessen workload and increase learner engagement in online learning has been highlighted by Dao et al.~\cite{Dao2021}. Finally, Nguyen et al.~\cite{Nguyen2021} have suggested a platform for online learning that incorporates a Vietnamese virtual assistant to help teachers present lectures to students and to make editing simple without the requirement for video recording.
	
	AI can already understand and communicate with humans, thanks to recent advancements in large language models (LLMs), which open up opportunities for its use in education. LLMs have shown great potential in the fields of education, content development, and language translation. The two primary architectures of LLMs are BERT (Bidirectional Encoder Representations from Transformers) and GPT (Generative Pre-trained Transformer).  In 2018, Google introduced BERT~\cite{devlin2018bert}, which has excelled in various natural language processing (NLP) tasks. The GPT algorithm, developed by OpenAI~\cite{alec2018improving}, was trained on extensive unlabeled text datasets. Facebook's RoBERTa~\cite{liu2019roberta} continued Google's research, and Google released T5~\cite{raffel2020exploring} in 2019. In 2020, OpenAI created GPT-3~\cite{brown2020language}, which demonstrated exceptional performance in various NLP tasks. Recently, OpenAI developed GPT-4~\cite{OpenAI_gpt_4_report}, a text-to-text machine learning system capable of processing both text and image inputs. GPT-4 has shown human-level performance in many professional and academic criteria, although it may not perform as well as humans in other contexts.
	
	With the introduction of {\href{https://chat.openai.com/}{ChatGPT} by OpenAI and \href{https://bing.com/chat}{Bing AI Chat (BingChat)} by Microsoft, which have been widely used in professions including marketing, medicine, law, and education, the popularity of AI has increased dramatically. Among the most frequent users of these applications are students. Although it is becoming more common, using LLMs in teaching certainly has drawbacks. Due to their widespread use, LLMs like ChatGPT and BingChat can quickly spread false information, with serious consequences~\cite{thorp2023chatgpt}. Since LLMs must be used in daily life, countermeasures must be put in place~\cite{van2023chatgpt}. A dataset can be created using real-world assessments, such as the Vietnamese National High School Graduation Examination (VNHSGE), to assess the capabilities of LLMs in the field of education. The evaluation's findings will give teachers a basis on which to build instructional plans for AI applications.
		
		In this article, we present a VNHSGE dataset that is built from exams of the VNHSGE and other exams of a similar nature. Mathematics, literature, English, physics, chemistry, biology, history, geography, and civic education are among the nine subjects covered by this dataset. The VNHSGE dataset specifically consists of 300 essays on Literature and 19,000 multiple-choice questions on other topics. We hope that researchers and educators can benefit greatly from the VNHSGE dataset in training and assessing LLMs. Additionally, we will annually update the VNHSGE dataset to reflect the most recent and pertinent data.
	
	\section{Related Work}
	\label{sec:related_work}
	
	\subsection{Datasets for training large language models}
	
	Large datasets are needed for pre-training and fine-tuning LLMs in order to attain their excellent performance in NLP applications. GLUE~\cite{wang2018glue} is made to assess the generalizability of language models while DecaNLP~\cite{mccann2018natural} is used to train multi-task NLP models. CLUE~\cite{xu2020clue} is a comprehensive benchmark that includes tasks like named entity recognition and text classification. MMLU dataset~\cite{hendrycks2020measuring} tests models in zero-shot and few-shot environments to assess their pretraining knowledge acquisition in 57 subjects in STEM, the humanities, social sciences, and more. 
	
	The accuracy and reasoning skills of LLMs are severely tested by datasets. SQuAD~\cite{rajpurkar2016squad} provides a sizable dataset for automatic question-answering research, enhancing the precision of deep learning models with 100,000 labeled questions and answers from many domains. For NLP models, especially next-word prediction models, LAMBADA~\cite{paperno2016lambada} presents challenging questions that call for in-depth comprehension of the text and next-word prediction to deliver an answer. The 96,000 reading comprehension and math problems in DROP~\cite{dua2019drop} demand a mechanism to resolve references inside the question, necessitating a thorough comprehension of paragraph content. HellaSwag~\cite{zellers2019hellaswag} assesses a model's capacity for deductive reasoning in scenarios that go against popular belief. WinoGrande~\cite{sakaguchi2021winogrande} is a benchmark dataset created to assess how well NLP models can resolve complex pronoun references in context using common sense reasoning, which calls for a thorough comprehension of real language and reasoning.
	
	The evaluation of machine comprehension and language understanding across many languages is supported by a number of datasets that concentrate on cross-lingual understanding. MLQA~\cite{lewis2019mlqa} assesses cross-lingual question-answering systems in seven different languages. XQuAD~\cite{artetxe2019cross} is a parallel dataset in eleven languages that assesses cross-lingual question answering performance. MKQA~\cite{longpre2021mkqa} is an open-domain question answering evaluation set that consists of 10k question-answer pairings in twenty-six topologically different languages. XGLUE~\cite{liang2020xglue} is a multilingual benchmark that assesses how well language models perform on a range of cross-lingual understanding tasks such as text classification, question answering, and machine translation in eleven languages. These datasets offer a useful tool for assessing how well machine understanding models perform across languages. MKQA~\cite{longpre2021mkqa} and XGLUE~\cite{liang2020xglue} offer a wider variety of tasks for evaluating cross-lingual understanding, whereas MLQA~\cite{lewis2019mlqa} and XQuAD~\cite{artetxe2019cross} concentrate on measuring cross-lingual question answering performance. The availability of datasets in several languages can help in the creation of more thorough machine comprehension models that can function in a variety of linguistic contexts, bringing up fresh directions for study and development in NLP.
	
	LLMs have difficulties when dealing with a variety of datasets that cover many areas. CoQA~\cite{reddy2019coqa} poses a unique challenge for big language models due to the conversational character of the questions and the responses, which can be free-form text and include texts from seven different domains. PILE~\cite{gao2020pile} is a large dataset of approximately 800GB of text from many sources, including as books, online pages, and scientific publications. ScienceQA~\cite{lu2022learn} is a great tool for creating machine comprehension models for scientific domains because it comprises a variety of natural science, language science, and social science.
	
	For question answering systems, there are numerous datasets available, each concentrating on a distinct subject.  MATH~\cite{hendrycks2021measuring} contains 12,500 difficult competition math problems with detailed solutions that allow models to produce answer derivations and justifications. GSM-8K~\cite{cobbe2021training} focuses on grade-school mathematics and covers a range of mathematical topics. Questions about biomedical research and medical scientific papers are included in BioASQ~\cite{tsatsaronis2015overview} and are categorized by level of difficulty. TQA~\cite{kembhavi2017you} combines the machine comprehension and visual question-answering paradigms for middle school science classes. SWAG~\cite{zellers2018swag} challenges the grounded commonsense inference, combining natural language inference and physically grounded reasoning. PIQA~\cite{bisk2020piqa} was developed as a commonsense reasoning dataset to examine the physical knowledge of current NLP models. PROST~\cite{aroca2021prost} is intended to test both causal and masked language models in a zero-shot environment. JEC-QA~\cite{zhong2020jec} and CaseHOLD~\cite{zheng2021does} are Chinese legal datasets. 
	
	In the discipline of NLP, large datasets are necessary for the development and assessment of machine learning models. The internet is a source of data for many question-answering datasets, particularly for websites like Wikipedia and search engines like Google. WebQuestions~\cite{berant2013semantic} are all defined as Freebase entities, with Freebase serving as the knowledge base. Each question in WikiQA~\cite{yang2015wikiqa} links to a possible related Wikipedia page, and lines from the summary part of the page are utilized as candidate answers. TriviaQA~\cite{joshi2017triviaqa} consists of 950K question-answer pairs drawn from 662K publications on the web and in Wikipedia. Because the context for each question is quite lengthy, span prediction may not be able to reliably produce the answers. One million pairs of questions and passages drawn from actual search queries are provided by the MS MARCO~\cite{nguyen2016ms}, which is updated on a regular basis with fresh search queries. Real-world, user-generated queries from Google.com and related Wikipedia pages are included in the NQ dataset~\cite{kwiatkowski2019natural}. Although there may be potential mistakes and incompleteness of information presented, the accuracy and completeness of the Wikipedia pages determine how accurate and thorough the responses are in these datasets. These datasets offer researchers useful tools for creating and enhancing machine learning models for problem-solving, with a variety of difficulties and chances for advancement in the field.
	
	Overall, these datasets provide valuable resources for evaluating LLMs in various tasks such as question answering, language modeling, text generation, reading comprehension, among others.
	
	\subsection{Datasets from the exams for training large language models}
	
	LLMs are increasingly being used, hence it is critical to assess their dependability and performance. Due to the richness and diversity of language usage in these datasets, language model evaluation using test datasets have acquired significance. Due to the high cost of data generation by human experts, existing exam datasets like the NTCIR QA Lab~\cite{shibuki2014overview}, Entrance Exams task at CLEF QA Track~\cite{penas2014overview}, ~\cite{rodrigo2015overview}, and AI2 Elementary School Science Questions dataset ~\cite{khashabi2016question}, have not been adequate for training advanced data-driven machine reading models. As a result, larger and more varied exam datasets are essential for LLMs training and evaluation. RACE~\cite{Lai2017} is one such dataset that has drawn interest. RACE is a dataset for automated reading comprehension with RACE-M and RACE-H, two subgroups from middle school and high school tests, respectively.
	
	Exam datasets are increasingly being used to evaluate LLMs, and the current datasets present interesting evaluation issues. The creation of novel test datasets, like the proposed Vietnamese High School Graduation Examination Dataset for LLMs, can improve the assessment of LLMs and guarantee their dependability in a variety of contexts. Using test datasets offers a demanding and varied evaluation of LLMs, which is essential for their usage in real-world applications. The creation of fresh test datasets can improve the evaluation procedure and increase the dependability of LLMs across a range of applications.
	
	\subsection{Datasets from high school exams for training large language models}
	
	Despite the fact that there are few datasets that concentrate on using high school topic exams to assess LLMs, there are still some datasets that contain high school exam questions that can be utilized for this purpose. GeoS~\cite{seo2015solving} intended for automatic math problem-solving. It includes SAT plane geometry questions from prior real SAT examinations and practice tests, each with a diagram and multiple-choice answers. Another dataset that includes multiple-choice questions from academic exams that range from grade 3 to grade 9 and need reasoning to answer is ARC~\cite{clark2018think}. The dataset was split into two parts, Easy and Challenge, with the latter comprising trickier problems. A supporting knowledge library of 14.3 million unstructured text passages is also included. SuperGLUE~\cite{wang2019superglue}, a more difficult dataset with tasks involving intricate thinking and common sense, contains many different jobs in it, some of which need you to respond to questions based on science passages from high school.  
	
	These high school datasets can still be utilized to assess language models' capacities to perceive and analyze natural language, despite the fact that there are few datasets explicitly created for testing LLMs using high school subject exams. Researchers can gain a deeper understanding of language models' strengths and limitations and create ways to enhance their performance by evaluating them against high school-level content. So that they can be used to assess LLMs, these datasets offer a variety of tasks and subject areas that are pertinent to high school education.
	
	\subsection{Our proposed dataset}
	
	To begin with, we conducted a search for available datasets in the "texts" category that are relevant to question answering task, as well as datasets that support the Vietnamese language. Our search was carried out on Paperwithcode as well as in previous studies. Table~\ref{tabl:dataset} displays the available datasets. We found that the majority of datasets consist of English texts, with only a few supporting Vietnamese. The most popular subjects are English, mathematics, and physics, while other subjects have relatively fewer related datasets (see Appendix section~\ref{paperwithcode} for further details).
	
	\begingroup
	\renewcommand{\arraystretch}{1.5} 
	\begin{table}[ht!]
		\caption{Related datasets}
		\label{tabl:dataset}
		\begin{tabular}{|c|l|}
			\hline
			\textbf{Subjects}        & \multicolumn{1}{c|}{\textbf{Dataset}}                                                                                                                                                                                                                                                                                                                                                                  \\ \hline
			Mathematics     & Mathematics~\cite{saxton2019analysing}, MMLU~\cite{hendrycks2020measuring}, MATH~\cite{hendrycks2021measuring} and GSM8K~\cite{cobbe2021training}                                                                                                                                                              \\ \hline
			Literature      & SCROLLS~\cite{shaham2022scrolls} and TAPE~\cite{taktasheva2022tape}                                                                                                                                                                                                                                                                                            \\ \hline
			English         & RACE~\cite{Lai2017}, MLQA~\cite{lewis2019mlqa}, SuperGLUE~\cite{wang2019superglue}, and DREAM~\cite{sun2019dream}                                                                                                                                                                                                  \\ \hline
			Physics         & TQA~\cite{kembhavi2017you}, SWAG~\cite{zellers2018swag}, PIQA~\cite{bisk2020piqa}, MMLU~\cite{hendrycks2020measuring}, PROST~\cite{aroca2021prost}, and ScienceQA~\cite{lu2022learn}                                                                                                                                     \\ \hline
			Chemistry       & SciQ~\cite{welbl2017crowdsourcing}, MMLU~\cite{hendrycks2020measuring}, and ScienceQA~\cite{lu2022learn}                                                                                                                                                                                                                            \\ \hline
			Biology         & BioASQ~\cite{tsatsaronis2015overview}, SciQ~\cite{welbl2017crowdsourcing}, MMLU~\cite{hendrycks2020measuring}, and ScienceQA~\cite{lu2022learn}                                                                                                                                                                   \\ \hline
			History         & MMLU~\cite{hendrycks2020measuring} and ScienceQA~\cite{lu2022learn}                                                                                                                                                                                                                                                                                  \\ \hline
			Geography       & MMLU~\cite{hendrycks2020measuring}, GeoTSQA~\cite{li2021tsqa}, and ScienceQA~\cite{lu2022learn}                                                                                                                                          \\ \hline
			Civic Education & JEC-QA~\cite{zhong2020jec}, and CaseHOLD~\cite{zheng2021does} \\ \hline
			Vietnamese      & MLQA~\cite{lewis2019mlqa}, XQuAD~\cite{artetxe2019cross}, and MKQA~\cite{longpre2021mkqa}                                                                                                                                                                                                                                            \\ \hline
		\end{tabular}
	\end{table}
	\endgroup

	It is essential to have datasets that contain questions at a high level of inference and cover a wide variety of topics because LLMs exhibit impressive human-level performance in several domains~\cite{OpenAI_gpt_4_report}. Utilizing real exam data from sources like  MMLU~\cite{hendrycks2020measuring} and ScienceQA~\cite{lu2022learn} is one approach to create these datasets. However, there are just a few datasets available right now that are focused primarily on using actual examinations to assess LLMs. To assess LLMs' capacity to understand and reason about natural language and challenging high school-level problems, the authors of this article created the VNHSGE dataset from the Vietnam National High School Graduation Examination. The employment of LLMs in teaching strategies can be decided upon by educators with the use of this dataset.
	
	There is a chance that LLMs could give students misleading or erroneous information~\cite{thorp2023chatgpt},~\cite{borji2023categorical}, or~\cite{dwivedi2023so} as they become more prominent in our daily lives. To solve this, it is essential that educators have access to databases that can accurately assess LLM models' capabilities and help them decide whether to use or reject them in their teaching strategies~\cite{rudolph2023chatgpt}. The VNHSGE dataset is created with this goal in mind, ensuring that LLMs give students reliable and secure information.
	
	\section{Vietnamese National High School Graduation Examination}
	\label{sec:Exam}
	
	The official and illustrative exam questions from the VNHSGE exams are all included in the VNHSGE dataset, which was compiled by high school instructors and the Vietnamese Ministry of Education and Training (VMET). It includes test questions from 2019–2023, covering a wide range of disciplines like mathematics, literature, English, physics, chemistry, biology, history, geography, and civic education. 
	
	Based on Bloom's taxonomy, the VNHSGE tests have varying degrees of difficulty, from knowledge-based questions that test fundamental comprehension to high-application questions that gauge one's capacity for in-depth analysis and information synthesis in the context of solving challenging situations. Knowledge (easy), comprehension (intermediate), application (difficult), and high application (extremely tough) are the four levels of complexity. We may learn more about LLM's capabilities for complicated reasoning as well as its strengths and shortcomings in dealing with various high school levels by evaluating its performance over a range of difficulty levels.
	
	The exam's three primary subjects—mathematics, literature, and English—as well as two combinations—the natural science combination of physics, chemistry, and biology, and the social science combination of history, geography, and civic education—make up the exam's framework.
	
	Table~\ref{tabl:MCQ} displays the multiple choice question subjects. Each exam contains 40 questions in each of the other topics in addition to 50 questions in mathematics and English. The dataset encompasses a wide range of disciplines and calls for a variety of abilities, from arithmetic to sophisticated reasoning.
	
	\begingroup
	\renewcommand{\arraystretch}{1.5} 
	\begin{table}[ht!]
		\caption{Subjects use multiple-choice questions}
		\label{tabl:MCQ}
		\centering
		\begin{tabularx}{\textwidth}{|c|X|}
			\hline
			\textbf{Subjects}         &  \multicolumn{1}{c|}{\textbf{Topics}}      \\ 
			\hline
			Mathematics     & spatial geometry, number series (arithmetic progression, geometric progression), combinations and probability, derivatives and applications, exponential and logarithmic functions, primitives and integrals, complex numbers, polyhedrons, rotating blocks, and Oxyz spatial calculus \\ 
			\hline
			English & pronunciation and stress, grammar, vocabulary, communication, reading fill-in-the-blank, reading comprehension, and writing skills \\ 
			\hline
			Physics & mechanical oscillations, mechanical waves, alternating current, electromagnetic oscillations and waves, light waves, quantum light, atomic nucleus, electric charge and field, direct current, electromagnetic induction, and light refraction \\ 
			\hline
			Chemistry &  theory of metals, alkali metals, alkaline-earth metals, aluminum, iron, inorganic and organic compounds, esters, lipids, amines, amino acids, and proteins, carbohydrates, polymers, and polymer materials \\ 
			\hline
			Biology & mechanisms of inheritance and mutation, laws of genetics, population genetics, applications of genetics, human genetics, evolution, ecology, plant organismal biology, and animal organismal biology
			\\ 
			\hline
			History & World histories: Soviet Union, Eastern European, Russian Federation; Asian, African, and Latin American; United States, Western Europe, and Japan;  international relations, scientific and industrial globalization networks; new world order after World War II. Vietnamese histories: 1884-1914, 1919-1930, 1930-1945, 1945-1954, 1954-1975, and 1975-2000 periods. 
			\\ 
			\hline
			Geography & geographical skills: atlas use, data table interpretation, and chart analysis; geographical theory: natural geography, population geography, economic sector geography, economic zone geography, sea geography, and island geography.
			\\ 
			\hline
			Civic Education & legal frameworks and regulations, fundamental rights of citizens, democratic principles and concepts, as well as case studies 
			\\ 
			\hline
		\end{tabularx}
	\end{table}
	\endgroup

	A systematic assessment technique called a literature dataset is used to assess a student's reading and writing abilities. Reading comprehension is tested in Part I, while writing skills are tested in Part II. Four questions in Part I ask students to examine and interpret an essay or poem, including determining the genre and any words or phrases that have particular meanings. Their own view on the text must be expressed in the final question, or it must be evaluated. Two essay questions are included in Part II, one on how to write a social argumentative essay and the other on how to write a literary argumentative essay. The essay questions test a student's ability to create a coherent and concise argument, back it with evidence, and analyze and interpret literary materials in order to develop a well-supported argument. The literature dataset offers a thorough assessment of a student's writing and reading comprehension abilities.
	
	The score distribution is an indicator to show how candidates scored in exams. Every year, VMET publishes the score distribution, which is shown as a chart for each subject. The distribution of scores is used to evaluate the competency of candidates and to assess exams according to their degree of difficulty, so assessing the level of competency of the applicants. Score distributions from 2019 to 2022 were gathered. We can assess the capability of LLMs by contrasting their outcomes with those of Vietnamese students (see Appendix section~\ref{sec:Score Spectrum} for a detailed breakdown of the score distribution and a comparison of LLMs' performance). The average score (AVS) and most reached score (MVS) of the Vietnamese students are presented in Table~\ref{tabl:AMVS} for a simpler comparison of the LLMs' performance. For instance, in 2019 the AVS and MVS for mathematics are 5.64 and 6.4, respectively. 
	
	\begingroup
	\renewcommand{\arraystretch}{1.5} 
	\begin{table}[ht!]
		\caption{Average score and Most reached score of Vietnamese students}
		\label{tabl:AMVS}
		\resizebox{\textwidth}{!}{%
			\begin{tabular}{c|cc|cc|cc|cc|cc|cc|cc|cc|cc|}
				\cline{2-19}
				\multicolumn{1}{l|}{}               & \multicolumn{2}{c|}{\textbf{Math}} & \multicolumn{2}{c|}{\textbf{Lit}} & \multicolumn{2}{c|}{\textbf{Eng}} & \multicolumn{2}{c|}{\textbf{Phy}} & \multicolumn{2}{c|}{\textbf{Che}} & \multicolumn{2}{c|}{\textbf{Bio}} & \multicolumn{2}{c|}{\textbf{His}} & \multicolumn{2}{c|}{\textbf{Geo}} & \multicolumn{2}{c|}{\textbf{Civ}} \\ \hline
				\multicolumn{1}{|c|}{\textbf{2019}} & \multicolumn{1}{c|}{5.64}      & 6.4      & \multicolumn{1}{c|}{5.49}       & 6      & \multicolumn{1}{c|}{4.36}    & 3.2    & \multicolumn{1}{c|}{5.57}    & 6.25   & \multicolumn{1}{c|}{5.35}     & 6       & \multicolumn{1}{c|}{4.68}    & 4.5    & \multicolumn{1}{c|}{4.3}     & 3.75   & \multicolumn{1}{c|}{6}        & 6       & \multicolumn{1}{c|}{7.37}        & 7.75       \\ \hline
				\multicolumn{1}{|c|}{\textbf{2020}} & \multicolumn{1}{c|}{6.67}      & 7.8      & \multicolumn{1}{c|}{6.61}       & 7      & \multicolumn{1}{c|}{4.58}    & 3.4    & \multicolumn{1}{c|}{6.72}    & 7.75   & \multicolumn{1}{c|}{6.71}     & 7.75    & \multicolumn{1}{c|}{5.6}     & 5.25   & \multicolumn{1}{c|}{5.19}    & 4.5    & \multicolumn{1}{c|}{6.78}     & 7.25    & \multicolumn{1}{c|}{8.14}        & 8.75       \\ \hline
				\multicolumn{1}{|c|}{\textbf{2021}} & \multicolumn{1}{c|}{6.61}      & 7.8      & \multicolumn{1}{c|}{6.47}       & 7      & \multicolumn{1}{c|}{5.84}    & 4      & \multicolumn{1}{c|}{6.56}    & 7.5    & \multicolumn{1}{c|}{6.63}     & 7.75    & \multicolumn{1}{c|}{5.51}    & 5.25   & \multicolumn{1}{c|}{4.97}    & 4      & \multicolumn{1}{c|}{6.96}     & 7       & \multicolumn{1}{c|}{8.37}        & 9.25       \\ \hline
				\multicolumn{1}{|c|}{\textbf{2022}} & \multicolumn{1}{c|}{6.47}      & 7.8      & \multicolumn{1}{c|}{6.51}       & 7      & \multicolumn{1}{c|}{5.15}    & 3.8    & \multicolumn{1}{c|}{6.72}    & 7.25   & \multicolumn{1}{c|}{6.7}      & 8       & \multicolumn{1}{c|}{5.02}    & 4.5    & \multicolumn{1}{c|}{6.34}    & 7      & \multicolumn{1}{c|}{6.68}     & 7       & \multicolumn{1}{c|}{8.03}        & 8.5        \\ \hline
			\end{tabular}
		}
	\end{table}
	\endgroup

	\section{Collection Methods}
	\label{sec:collection}
	
	Any research project must start with the gathering of raw data, and for this study, we obtained our data from free public websites in Vietnam. We painstakingly selected and arranged the gathered information into a brand-new dataset of questions from VNHSGE and similar exams. We specifically used the illustrated exam questions that VMET publishes every year. To give students and teachers a general idea of the content and structure of the official exam, these exam questions are made available to them. We gathered the official exam questions from VMET in addition to the illustrated exam questions. VMET produced a brief answer key following the exam, and the teachers then supplied more thorough responses. Additionally, we have included similar exam questions that are created by instructors and high schools around Vietnam in our data collection. This strategy guarantees that our dataset has a wide variety of questions that cover a wide range of subjects and degrees of difficulty. Our dataset contains exam questions, answers, and thorough step-by-step explanations (see Appendix section~\ref{sec:Raw data} for a raw data example) that have all been meticulously examined and validated by our team of subject matter experts. Instead of employing \href{https://www.mturk.com/}{Amazon Mechanical Turk}, as some earlier datasets did, detailed explanations are given by qualified teachers.
	
	The extensive dataset gathered for this study offers a great chance to assess how well LLMs complete Vietnamese national tests. Our dataset's vast variety of themes and levels of difficulty provide a thorough assessment of the LLMs' accuracy and deductive reasoning abilities when responding to various questions. We may learn important lessons about the benefits and drawbacks of LLMs in handling actual tests by utilizing this dataset, which can guide further study and advancement in this area.
	
	\section{Dataset}
	\label{sec:dataset}
	
	The dataset is available in Word format and JSON format. In addition, we provide the dataset in Vietnamese and English (VNHGE-V and VNHSGE-E). The dataset was originally written in Vietnamese. Using GPT-4/ChatGPT, the dataset is translated into English, similar to how OpenAI tests the capability of GPT-4~\cite{OpenAI_gpt_4_report} in other languages by using Azure Translate to translate the MMLU benchmark~\cite{hendrycks2020measuring} into another language. Language models can handle several languages, as is well recognized. However, if LLMs do not support multilingualism they can use the English VNHSE version. We may also employ comparable strategies for additional languages by using GPT-4/ChatGPT, BingChat/Azure Translate, and Google Translate.
	
	\subsection{Format}
	\label{sec:Format}
	
	In the VNHSGE dataset, we convert formulas, equations, tables, images, and charts from raw text formats like Word, Pdf, and HTML into a text-only format and an image folder including steps: (1) collecting raw data and convert them into Word format, (2) transforming symbols, formulas, and equations into Latex format, (3) converting Word format to JSON format (see Appendix section~\ref{convert_data_format} for more details of a step-by-step conversion).  
	
	\subsubsection{Word format}
	
	We transform the symbols, equations, and formulas into text using the Latex format so that it is compatible with LLMs transformed BERT or GPT. For those who lack programming skills, we also offer a text format in the form of a Word file for evaluating the performance of LLMs. In this situation, the VNHSGE dataset can be thought of as a question bank for assessing LLMs over a range of subjects. However, full language models like ChatGPT and BingChat are typically more appropriate in this situation. It is vital to keep in mind that symbols, formulas, and equations were converted to text format while utilizing a text format in a Word file; we only ask questions of LLMs and receive responses.
	
	\begin{multicols}{2}
		\textbf{Question}: Let \$y=f(x)\$ be a cubic function with the graph shown in the picture.\\
		\begin{tikzpicture}[scale=0.75]
			\begin{axis}[
				axis lines=middle,
				xlabel={$x$},
				ylabel={$y$},
				xmin=-2.5, xmax=5,
				ymin=-2, ymax=4.5,
				xtick={-2,2},
				ytick={-1,2},
				legend style={at={(1,1)},anchor=north east},
				]
				\addplot[domain=-2.5:4.5, samples=200, ultra thick, blue] {0.0208*x^5-0.1417*x^4 + 0.3375*x^3 -0.1833*x^2 -1.6833*x+ 3.0000};
			\end{axis}
		\end{tikzpicture} \\
		The number of real solutions of the equation \$|f(x\textasciicircum3-3 x)|=\textbackslash{}frac\{2\}\{3\}\$ is:\\
		\begin{oneparchoices}
			\choice 6
			\choice 10
			\choice 3
			\choice 9
		\end{oneparchoices} \\
		\textbf{Solution}: From the graph of the function \$y=f(x)\$, we deduce that the graph of the function \$y=|f(x)|\$ is:\\
		\begin{tikzpicture}[scale=0.75]
			\begin{axis}[
				axis lines=middle,
				xlabel={$x$},
				ylabel={$y$},
				xmin=-2.5, xmax=5,
				ymin=-2, ymax=4.5,
				xtick={-2,2},
				ytick={-1,1,2},
				legend style={at={(1,1)},anchor=north east},
				]
				\addplot[domain=-3:-2, samples=200, ultra thick, blue] {-(0.0208*x^5-0.1417*x^4 + 0.3375*x^3 -0.1833*x^2 -1.6833*x+ 3.0000)};
				\addplot[domain=-2:2, samples=200, ultra thick, blue] {0.0208*x^5-0.1417*x^4 + 0.3375*x^3 -0.1833*x^2 -1.6833*x+ 3.0000};
				\addplot[domain=2:4, samples=200, ultra thick, blue] {-(0.0208*x^5-0.1417*x^4 + 0.3375*x^3 -0.1833*x^2 -1.6833*x+ 3.0000)};
				\addplot[domain=4:4.5, samples=200, ultra thick, blue] {0.0208*x^5-0.1417*x^4 + 0.3375*x^3 -0.1833*x^2 -1.6833*x+ 3.0000};
			\end{axis}
		\end{tikzpicture}\\
		Setting \$t=x\textasciicircum3-3x\$, we have \$|f(x\textasciicircum3-3x)|=\textbackslash{}frac\{2\}\{3\} \textbackslash{}Leftrightarrow |f(t)|=\textbackslash{}frac\{2\}\{3\}\$.
		From the above graph, we conclude that the equation \$|f(t)|=\textbackslash{}frac\{2\}\{3\}\$ has six distinct solutions \$t=t\_\{i\}\$ (with \$i=\textbackslash{}overline\{1,6\}\$ and \$(t\_\{1\}<-2;~-2<t\_\{2\},~t\_\{3\}<2;~t\_\{4\},~t\_\{5\},~t\_\{6\}>2)\$. \\
		Considering the function \$t(x)=x\textasciicircum\{3\}-3x\$, we have \$t\textasciicircum\{\textbackslash{prime}\}(x)=3 x\textasciicircum\{2\}-3 ; t\textasciicircum\{\textbackslash{}prime\}(x)=0~\textbackslash{}Leftrightarrow x= \textbackslash{}pm1\$.
		The sign variation table of \$t(x)\$ is:\\
		\begin{tikzpicture}[scale=0.65]
			\newcommand*{ \E}{ \ensuremath{ \mathrm{e}}}.
			\tkzTabInit[lgt=3,espcl=2.5]
			{$x$ /1,$f'(x)$ /1,$f(x)$ /2}
			{$-\infty$,$-1$,$1$,$+\infty$}
			\tkzTabLine{,+,z,-,z,+}
			\tkzTabVar{-/$-\infty$,
				+/$2$,
				-/$-2$,
				+/$+\infty$}
			\tkzTabIma{2}{4}{3}{$0$}            
		\end{tikzpicture}
		Based on the table of variations, we have: 
		\begin{itemize}
			\item The equation \$x\textasciicircum\{3\}-3x=t\_\{1\}\$ has one solution (since \$(t\_\{1\}<-2)\$.
			\item Each equation~\$x\textasciicircum\{3\}-3x=t\_\{2\}, x\textasciicircum\{3\}-3x=t\_\{3\}\$ has three distinct solutions (since \$-2<t\_\{2\}, t\_\{3\}<2\$). 
			\item Each equation~\$x\textasciicircum\{3\}-3x=t\_\{4\},~x\textasciicircum\{3\}-3x=t\_\{5\}, x\textasciicircum\{3\}-3x=t\_\{6\}\$ has one solution (since \$t\_\{4\}, t\_\{5\}, t\_\{6\}>2\$). 
		\end{itemize}
		The equation \$|f(x\textasciicircum\{3\}-3x)|=\textbackslash{}frac\{2\}\{3\}\$ has 10 solutions. Therefore, the answer is \textcolor{red}{\textbf{B}}. 10. 		
	\end{multicols}	
	
	\subsubsection{JSON format}
	
	We adopt the JSON format for the VNHSEG dataset because it is ideal for LLMs training, testing, and evaluation. Because it makes both accessing and processing textual information linked to syntactic structure and content-related information simple, the JSON format is especially well suited for LLM inputs. A variety of text data, including formulas, equations, tables, and images, can be stored and represented in a flexible and expandable manner using the JSON format. In general, the usage of JSON format makes the VNHSEG dataset compatible with a variety of LLMs and makes it easier to train, test, and evaluate LLMs. 
	
	
	\begin{multicols}{2}	
		\{
		
		"ID": "Q1", \\
		"IQ": "Math/Q1\_1.png", \\
		"Q": "Let \$y=f(x)\$ be a cubic function with the graph shown in the picture. \textbackslash{}n\textbackslash{}nThe number of real solutions of the equation \$|f(x\textasciicircum3-3 x)|=\textbackslash{}frac\{2\}\{3\}\$ is:\textbackslash{}nA. 6.    \textbackslash{}nB. 10. \textbackslash{}nC. 3. \textbackslash{}nD. 9.", \\
		"C": "\textcolor{red}{\textbf{B}}", \\
		"IE": "Math/Q1\_2.png, Q1\_3.png", \\
		"E": "From the graph of the function \$y=f(x)\$, we deduce that the graph of the function \$y=|f(x)|\$ is:\textbackslash{}nSetting \$t=x\textasciicircum3-3x\$, we have \$|f(x\textasciicircum3-3x)|=\textbackslash{}frac\{2\}\{3\} \textbackslash{}Leftrightarrow |f(t)|=\textbackslash{}frac\{2\}\{3\}\$. \textbackslash{}nFrom the above graph, we conclude that the equation \$|f(t)|=\textbackslash{}frac\{2\}\{3\}\$ has six distinct solutions \$t=t\_\{i\}\$ (with \$i=\textbackslash{}overline\{1,6\}\$ and \$(t\_\{1\}<-2;~-2<t\_\{2\},~t\_\{3\}<2;~t\_\{4\},~t\_\{5\},~t\_\{6\}>2)\$. \textbackslash{}nConsidering the function \$t(x)=x\textasciicircum\{3\}-3x\$, we have \$t\textasciicircum\{\textbackslash{prime}\}(x)=3x\textasciicircum\{2\}-3;~t\textasciicircum\{\textbackslash{}prime\}(x)=0~\textbackslash{}Leftrightarrow x= \textbackslash{}pm1\$.The sign variation table of \$t(x)\$ is:\textbackslash{}nBased on the table of variations, we have:\textbackslash{}n\textbackslash{}begin\{itemize\} \textbackslash{}n\textbackslash{}item The equation \$x\textasciicircum\{3\}-3x=t\_\{1\}\$ has one solution (since \$(t\_\{1\}<-2)\$. \textbackslash{}n\textbackslash{}item Each equation~\$x\textasciicircum\{3\}-3x=t\_\{2\}, x\textasciicircum\{3\}-3x=t\_\{3\}\$ has three distinct solutions (since \$-2<t\_\{2\}, t\_\{3\}<2\$). \textbackslash{}n\textbackslash{}item Each equation~\$x\textasciicircum\{3\}-3x=t\_\{4\},~x\textasciicircum\{3\}-3x=t\_\{5\}, x\textasciicircum\{3\}-3x=t\_\{6\}\$ has one solution (since \$t\_\{4\}, t\_\{5\}, t\_\{6\}>2\$). \textbackslash{}n\textbackslash{}end\{itemize\} \textbackslash{}nThe equation \$|f(x\textasciicircum\{3\}-3x)|=\textbackslash{}frac\{2\}\{3\}\$ has 10 solutions. Therefore, the answer is \textcolor{red}{\textbf{B}}. 10", \\
		\}
		
	\end{multicols}	
	
	\emph{ ID refers to the ID of the question; IQ refers to the images of the question; Q refers to the question content; C refers to the choice options; IE refers to the images of the explanation; and E refers to the explanation content.}	
	
	\subsection{Language}
	
	Vietnamese and English were used in the construction of the VNHSGE dataset. VNHSGE-V is in Vietnamese and VNHSGE-E is in English. GPT-4/ChatGPT was used to translate VNHSGE-V into VNHSGE-E. According to earlier research \cite{OpenAI_gpt_4_report},
	\cite{jiao2023chatgpt}, and
	\cite{bang2023multitask}, GPT-4/ChatGPT can successfully serve as the appropriate translation engine in this circumstance. It should be noted that ChatGPT or BingChat were used to translate the illustrative examples for the dataset presented in this work from Vietnamese to English.
	
	\subsection{Subdataset}
	
	Table~\ref{tabl:VNHSGE} shows the VNHSGE dataset structure. The dataset for mathematics and English consists of 2500 multiple-choice questions per subject, while the other multiple-choice subjects have 2000 questions. Literature has 50 exams with 300 essay questions. The dataset contains a large number of questions spanning various topics, ranging from recall-level knowledge to complex multi-step reasoning requirements (see Appendix section~\ref{Dataset_response} for more details of examples).    
	
	\begingroup
	\renewcommand{\arraystretch}{1.5} 
	\begin{table}[ht!]
		\caption{VNHSGE dataset structure}
		\label{tabl:VNHSGE}
		\begin{center}
			\begin{tabular}{|c|c|c|c|c|} 
				\hline
				{\textbf{Subject}} & \textbf{Exam Type} & \textbf{Number of questions per exam} & \textbf{Number of exams} & \textbf{Question Total}  \\ 
				\hline
				{Mathematics} & Multiple choice                       & 50 & 50 & 2500 \\ 
				{Literature} & Essay                       & 6 & 50 & 300  \\ 
				{English} & Multiple choice                       & 50 & 50 & 2500  \\
				{Physics} & Multiple choice                       & 40 & 50 & 2000  \\
				{Chemistry} & Multiple choice                       & 40 & 50 & 2000  \\
				{Biology} & Multiple choice                       & 40 & 50 & 2000  \\
				{History} & Multiple choice                       & 40 & 50 & 2000  \\
				{Geography} & Multiple choice                       & 40 & 50 & 2000  \\
				{Civic Education} & Multiple choice                       & 40 & 50 & 2000  \\
				\hline
				{\textbf{Total}} &\multicolumn{4}{c|}{\textbf{19000 multiple-choice questions and 300 essay questions}}    \\
				\hline
			\end{tabular}
		\end{center}
	\end{table}
	\endgroup 
	
	\subsubsection{Mathematics}
	
	In contrast to a number of earlier mathematics datasets, including the Mathematics dataset~\cite{saxton2019analysing}, MATH dataset~\cite{hendrycks2021measuring}, GSM8K dataset~\cite{cobbe2021training}, and ScienceQA dataset~\cite{lu2022learn}, the VNHSGE mathematics dataset covers a wide range of topics, including spatial geometry, number series (arithmetic progression, geometric progression), combinations and probability, derivatives and applications, exponential and logarithmic functions, primitives and integrals, complex numbers, polyhedrons, rotating blocks, and Oxyz spatial. To help models learn how to provide answer derivations and explanations, the dataset includes questions and related solutions, which are supplied in a complete step-by-step solution (\ref{VNHSGE_Math}). The VNHSGE mathematics dataset also includes straightforward to complicated questions, necessitating strong mathematical reasoning skills from LLMs in both question answering and visual question answering tasks. 
	
	First, the knowledge level question (\ref{VNHSGE_Math_K}) has been created such that LLMs can quickly and simply solve it using their fundamental understanding. We need 1-2 steps to solve this kind of question. The mathematical calculation skills of LLMs may be put to the test by questions like ($+~-\times~\div~\int~\frac{\mathrm{d} }{\mathrm{d} x}$). In order to answer the comprehension level questions (\ref{VNHSGE_Math_C}), LLMs must then infer a few steps to arrive at the appropriate answer. LLMs' capacity for reasoning is put to the test by this kind of question at the level of an average student. Further complicating matters for LLMs is the fact that these kinds of application level problems (\ref{VNHSGE_Math_A}) mix several different mathematical ideas and need multiple complicated reasoning steps. These inquiries may assess a model's capacity for rational thinking and mathematical knowledge synthesis. Last but not least, the high application level questions (\ref{VNHSGE_Math_H}) frequently feature unique solutions based on advanced mathematical reasoning and practical problem-solving techniques. LLMs need to have very strong deductive reasoning skills and expertise in solving difficult mathematical problems in order to answer these kinds of inquiries.
	
	The VNHSGE mathematics dataset is a thorough collection that addresses a variety of mathematical topics. The dataset was created to evaluate LLMs' capacity for mathematical reasoning on a range of levels, including knowledge, comprehension, application, and high application. The questions in the dataset range in complexity from simple to complicated, therefore the models must have strong inference and reasoning skills. The dataset includes questions that may be answered in one or two steps using fundamental information, as well as problems that call for several steps and knowledge synthesis. The VNHSGE mathematics dataset is an excellent resource for developing and assessing LLMs' mathematical reasoning and inference skills since it presents a strong challenge to their mathematical aptitude in both breadth and depth.
	
	\subsubsection{Literature}
	
	The literary exam, a structured assessment tool used to evaluate a student's reading comprehension and writing abilities, serves as the foundation for the VNHSGE literature dataset. This dataset can be deployed for the training and evaluation of LLMs for a variety of language understanding tasks, including essay writing, writing proficiency, and reading comprehension. The dataset is divided into two parts: the question and the answer~(\ref{VNHSGE_Lit}). The question section~(\ref{VNHSGE_Lit_Q}) is divided into two parts. Four questions in Part I's reading comprehension assessment ask students to analyze and understand a paragraph or poetry. The questions ask one to identify the genre and any words or phrases with unique meanings before you analyze their significance. Students must give their own personal opinion of the text or assess another person's personal view of the text for the final question. Writing abilities are the main topic of Part II, which also contains two essay challenges, one on how to write an arguing social essay and the other on how to write an argumentative literary essay. The essay questions test a student's ability to formulate a coherent and succinct argument, back it with evidence, and analyze and interpret literary materials in order to develop a well-supported argument. The answer suggestions and grading guidelines are included in the solution~(\ref{VNHSGE_Lit_Q}). The scoring criteria are written down in great depth in the grading instructions~(\ref{VNHSGE_Lit_S}). The suggested answers are given in accordance with the evaluation criteria.
	
	The dataset created based on the answer key with grading guidelines and answer recommendations can assist LLMs in strengthening their capacity to respond to inquiries and offer pertinent justifications based on certain rating metrics. Language models can become more accurate and efficient at answering queries by being trained on this dataset to better grasp and adhere to grading requirements. This dataset offers a thorough assessment of a student's reading comprehension and writing abilities in high school literature, thereby providing a valuable tool for developing and testing LLMs for a variety of language understanding tasks, including sentiment analysis, question answering, text generation, and text summarization. Moreover, the VNHSGE literature dataset is built in Vietnamese, which challenges the ability of LLMs in NLP as Vietnamese is one of the languages with many layers of meaning. Additionally, because Vietnamese is one of the languages with multiple layers of implications, the VNHSGE literature dataset challenges LLMs' proficiency in NLP.

	\subsubsection{English}

	For datasets involving question-answering, there are plenty of options. For instance, the DREAM dataset~\cite{sun2019dream} focuses on reading comprehension for dialogue while the RACE dataset~\cite{Lai2017} exclusively considers paragraph reading comprehension. Another dataset that covers eight tasks is SuperGLUE~\cite{wang2019superglue}. These datasets have performed admirably for the intended purposes, but they do not provide a comprehensive examination of the LLMs' general language processing abilities.
	
	The VNHSGE English dataset contains an assortment of exam questions from high school exams that cover a variety of topics and demand a variety of linguistic abilities (\ref{VNHSGE_Eng}). In the dataset's pronunciation and stress questions (\ref{VNHSGE_Eng_PS}), LLMs are asked to choose the word whose underlined portion is pronounced differently from the other three. LLMs are also required to select the proper response from a list of alternatives for questions on vocabulary and grammar (\ref{VNHSGE_Eng_G}), identify terms with opposite or similar meanings, choose the closest-meaning sentence, and fix underlined parts. In order to pass the communication skills test (\ref{VNHSGE_Eng_C}), LLMs are required to select the appropriate response for each conversation. LLMs fill in each of the numbered blanks in the reading fill-in-the-blank questions (\ref{VNHSGE_Eng_RB}) by choosing the appropriate word or phrase. Furthermore, LLMs are required to read passages in order to respond to questions about reading comprehension (\ref{VNHSGE_Eng_RC}). At the human level, the dataset encompasses an extensive variety of topics and activities. The dataset is also made up of questions and answers, where the answers are explained in great depth in the solutions. This aids in teaching LLMs how to think critically.
	
	The VNHSGE English dataset is a useful tool for LLMs to enhance their proficiency in a range of topics and abilities connected to English language comprehension at the human-level performance. These models can perform better in a variety of language-related tasks, including question answering, language modeling, text generation, reading comprehension, text summarization, etc. by being trained on this dataset, which may assist these models comprehend and process natural language effectively.

	\subsubsection{Physics}
	
	In the previous physics datasets, the TQA dataset~\cite{kembhavi2017you} concentrates on life, earth, and physical sciences and includes both text and pictures for machine comprehension and visual question answering. Although the TQA dataset is intended for middle school students, it appears to be simple enough for LLMs in use today. The PIQA dataset~\cite{bisk2020piqa} tests the LLMs' capacity for physical reasoning, it is suited for honing their capacity for inference and leaves out the computationally demanding physics problems that they must be able to answer. Physics-related topics such as materials, magnets, velocity, and forces, force and motion, particle motion and energy, heat, and thermal energy, states of matter, kinetic and potential energy, and mixtures are covered in the ScienceQA dataset~\cite{lu2022learn}. Although ScienceQA covers a wide range of topics this is merely elementary physics. On the other hand, the VNHSGE physics dataset is geared toward high school students. The VNHSGE physics dataset also focuses on more complicated topics like electromagnetic oscillations and waves, light waves, quantum light, atomic nuclei, direct current, electromagnetic induction, and light refraction (\ref{VNHSGE_Phy}). The prior datasets can be difficult for LLMs since they demand one to comprehend and make connections between a wide range of scientific principles and notions. The VNHSGE physics dataset, however, may present a bigger challenge for language models because it deals with more complex and specialized physics topics and necessitates a higher level of scientific understanding and reasoning abilities to accurately respond to the questions.
	
	$50\%$ of the questions in the VNHSGE physics dataset are theoretical, and $50\%$ are practical and applied. Most theoretical problems fall under the knowledge level (\ref{VNHSGE_Phy_K}), which calls for both inference and a firm comprehension of theoretical knowledge. For questions at the comprehension level (\ref{VNHSGE_Phy_C}), there is a higher degree of inference about knowledge and mathematical abilities. The application level questions (\ref{VNHSGE_Phy_A}) come next, which have a high categorization and draw on complex physics concepts like understanding of practice and application. The high application level questions (\ref{VNHSGE_Phy_H}) are the last type. These include experimental questions as well as questions that make use of graphs related to mechanical oscillations and alternating currents. These inquiries demand a very high degree of inference, and the unique solutions call for in-depth knowledge of high school physics challenges.
	
	Physical concepts like mechanical oscillations, waves, quantum mechanics, and atomic nuclei might be difficult for LLMs to understand and rationalize when presented with physical information from the VNHSGE. In addition to demanding the ability to retain information, the datasets additionally inquire about the ability to draw conclusions, apply ideas to concrete circumstances, and even solve challenging issues. It is a difficult undertaking for any LLMs because the high application-level questions in the dataset demand specialized knowledge and experience in addressing physics issues at the high school level.
	
	\subsubsection{Chemistry}
	
	There aren't many datasets in the field of chemistry that are specifically focused on tackling questions. The SciQ dataset~\cite{welbl2017crowdsourcing} tests LLMs on their knowledge of chemistry with multiple-choice questions. It rates the model's comprehension and deductive reasoning skills in regard to chemistry-related scientific ideas and concepts. The chemistry dataset in \cite{hendrycks2020measuring} focuses on the LLMs' accuracy in chemistry subjects from high school, including chemical reactions, ions, acids, and bases, to college, like analytical, organic, inorganic, and physical. However, there are only a few chemistry questions. Understanding and responding to questions about chemistry subjects like solutions, physical and chemical changes, atoms and molecules, and chemical reactions are the main objectives of ScienceQA dataset~\cite{lu2022learn}. The VNHSGE chemistry dataset, on the other hand, presents difficulties for LLMs in understanding and responding to questions regarding a variety of chemistry topics, including metals, inorganic and organic molecules, polymers, and more (\ref{VNHSGE_Che}). It rates the model's comprehension and deductive reasoning skills with regard to a variety of chemistry concepts and principles.
	
	The VNHSGE chemistry dataset is made up of $30\%$ computational tasks and $70\%$ theoretical questions. Usually, theoretical problems require knowledge and comprehension. The knowledge-level questions are typically brief and demand information-retrieval-level knowledge (\ref{VNHSGE_Che_K}). Subsequently, the computations in the comprehension level (\ref{VNHSGE_Che_C}) section are rather straightforward, requiring only 1 or 2 operations for problems. Next, the high-level reasoning and the synthesis of several concepts are required to answer the application-level questions (\ref{VNHSGE_Che_A}). Finally, the high-application questions (\ref{VNHSGE_Che_H}) require in-depth knowledge, logical reasoning, and the synthesis of several chemical reaction equations.
	
	The VNHSGE chemistry dataset evaluates LLMs' high-level reasoning and problem-solving abilities as well as their comprehension of chemistry principles across a variety of topics and levels of difficulty. The dataset necessitates that the models have an adequate knowledge of chemical principles and be able to implement that understanding in challenging contexts, such as the synthesis and analysis of chemical reactions.
	
	\subsubsection{Biology}
	
	Similar to chemistry, there aren't many biology datasets created expressly for question answering tasks. BioASQ~\cite{tsatsaronis2015overview} concentrates on medical fields rather than biological ones. The SciQ~\cite{welbl2017crowdsourcing} dataset makes it difficult for LLMs to correctly respond to Biology-related multiple-choice questions on science exams. The dataset evaluates how well the model can understand and justify biological science principles and notions. The MMLU dataset~\cite{hendrycks2020measuring} assesses LLMs' accuracy in subjects from high school and college biology, including natural selection, heredity, cell cycle, and more. The ScienceQA dataset~\cite{lu2022learn}, on the other hand, focuses on understanding and responding to questions about molecular and cellular biology. Because of its extensive coverage of topics including genetic laws, population genetics, applications of genetics, human genetics, evolution, ecology, plant organismal biology, and animal organismal biology, the VNHSGE biology dataset presents a significant challenge to LLMs (\ref{VNHSGE_Bio}). 
	
	The questions in the VNHSGE biology dataset are highly challenging and complicated, and in order to accurately respond to them, one must have a thorough understanding of all aspects of biology. According to the dataset's design, there should be $75\%$ theoretical questions and $25\%$ exercises, with $70\%$ of the questions being at the knowledge and comprehension levels and $30\%$ of the questions focusing on application and higher-order thinking skills. The dataset, which includes questions of varying complexity, focuses on the capacity for calculation and inference. The knowledge level questions (\ref{VNHSGE_Bio_K}) demand a comprehensive understanding of biology to answer correctly, while the comprehension level questions (\ref{VNHSGE_Bio_C}) require one to three steps of deductive reasoning to find the answer. The application level questions (\ref{VNHSGE_Bio_A}) focus on areas including rules of genetics, human genetics, population genetics, and mechanisms of inheritance and mutation and call for the capacity to synthesize knowledge. The high application level questions (\ref{VNHSGE_Bio_H}) require sophisticated analysis and problem-solving skills.
	
	The VNHSGE biology dataset is a substantial challenge for LLMs since it calls for a mix of in-depth knowledge and sophisticated reasoning abilities in order to correctly understand and respond to questions about a wide range of biology topics.
	
	\subsubsection{History}
	
	Both the MMLU dataset~\cite{hendrycks2020measuring} and ScienceQA dataset~\cite{lu2022learn} evaluate how well LLMs perform when answering questions about historical events. While the MMLU dataset~\cite{hendrycks2020measuring} assesses LLMs' accuracy in high school histories concepts like High School US History, High School European History, and High School World History, the ScienceQA dataset~\cite{lu2022learn} focuses on understanding and responding to questions about American and global history.
	
	The purpose of the VNHSGE history dataset is to assess LLMs' knowledge of historical events and milestones as well as to give correct analysis of historical events (\ref{VNHSGE_His}). The dataset contains $80\%$ questions at the knowledge and comprehension levels covering a wide range of topics including Vietnamese and global histories (\ref{VNHSGE_His_K} and \ref{VNHSGE_His_C}) . To answer these kinds of inquiries, one must not only accurately record the facts but also use historical reasoning. Across topics in Vietnamese history from 1919 to 1975, the dataset contains $20\%$ of questions that require application and high application levels (\ref{VNHSGE_His_A} and \ref{VNHSGE_His_H}). The majority of the questions concern comparison essays, connections between topics, links between Vietnamese history and world history, or commentary and summaries of historical periods to identify key characteristics or the substance of historical events. The capacity to analyze, contrast, and comment on historical events is necessary for these kinds of issues.
	
	The VNHSGE history dataset is utilized for evaluating how well LLMs can recall and comprehend historical events as well as their timeframes. The questions in the dataset range from simple to complex, requiring varying degrees of deductive reasoning and inference skills. To correctly respond to the questions in the dataset, LLMs must be able to interpret and analyze complicated historical events, appreciate the relationships between them, and draw inferences from them.
	
	\subsubsection{Geography}
	
	Few specialized datasets are available for geography question-answering tasks. The MMLU dataset~\cite{hendrycks2020measuring} includes a few inquiries about high school geography concepts including population movement, rural land use, and urban processes. While the ScienceQA dataset~\cite{lu2022learn} focuses on questions about state capitals, geography, maps, and more. Additionally, the geography dataset in \cite{hardalov2019beyond} includes 612 Bulgarian multiple-choice questions for the matriculation exam for the 12th grade. The GeoTSQA dataset~\cite{li2021tsqa}, which was compiled from high school exams in China, has 1,000 actual questions in the geography domain that are contextualized by tabular scenarios. The VNHSGE geography dataset is intended to assess LLMs' knowledge of geographical concepts such as natural geography, population geography, economic sector geography, economic zone geography, sea geography, and island geography as well as geographical skills such as atlas use, data table interpretation, and chart analysis.
	
	The questions in the VNHSGE geography dataset are ordered in order of increasing complexity, with $80\%$  of the questions falling into the basic category (knowledge and understanding) and $20\%$  falling into the advanced category ($10\%$ application and $10\%$  high-level application) (\ref{VNHSGE_Geo}). $50\%$  of the exam's questions, such as chart analysis (\ref{VNHSGE_Geo_Chart}), data table interpretation (\ref{VNHSGE_Geo_Table}), and atlas use (\ref{VNHSGE_Geo_Atlas}), involve geographic knowledge. LLMs must be able to solve problems in order to master these skills. Additionally, LLMs must be able to think logically, have a broad understanding of society, be adept at solving problems, and have a high degree of critical thinking to complete the diversified questions (\ref{VNHSGE_Geo_Complex}). 
	
	Questions in the VNHSGE geography dataset call for a variety of abilities, such as data analysis, chart interpretation, and atlas use, which can assist in training LLMs to comprehend and process complicated material in these fields. The dataset also contains questions that call for reasoning, problem-solving, and critical thinking, which can aid in the development of more sophisticated language skills in language models.
	
	\subsubsection{Civic Education}
	
	There have been numerous attempts to construct datasets connected to the legal profession and ethics, which has recently received special attention. While the JEC-QA dataset~\cite{zhong2020jec}  contains questions connected to the national judicial examination in China, the CJRC dataset~\cite{duan2019cjrc} comprises documents and questions relating to legal knowledge in China. The CaseHOLD dataset~\cite{zheng2021does}, which focuses on finding the critical components in a legal case, is a novel and difficult dataset in the subject of law. While the PolicyQA dataset~\cite{ravichander2019question} focuses on comprehending the privacy policies of websites, the PrivacyQA dataset~\cite{ahmad2020policyqa} focuses on queries regarding the privacy policies of mobile applications. To guarantee the accuracy of the replies, both databases offer questions that have been reviewed by experts. The Vietnamese transportation law dataset~\cite{bach2017question} and the Vietnamese law dataset~\cite{kien2020answering} both concentrate on questions pertaining to law, but the Vietnamese transportation law dataset is more concerned with traffic law and the Law dataset is more concerned with broad legal issues. Additionally, MMLU dataset~\cite{hendrycks2020measuring} has a few questions about professional law as well as questions about international law including torts, criminal law, contracts, etc. Focused on questions about civics subjects like social skills, governance, and the constitution is the ScienceQA dataset~\cite{lu2022learn}. While the VNHSGE civic education dataset is intended to provide LLMs with civic education and legal training, it also focuses on case studies and multiple-choice questions on topics such as legal frameworks and regulations, fundamental civil rights, democratic principles, and case studies.
	
	The purpose of VNHSGE civic education dataset is to evaluate LLMs' understanding of and ability to apply legal concepts (\ref{VNHSGE_Civ}). $70\%$ of the exam's questions are knowledge and comprehension level questions (\ref{VNHSGE_Civ_K} and \ref{VNHSGE_Civ_C}). $30\%$ of the questions are application and high application levels, focused on topics like Citizens' fundamental rights; types of legal infractions; and equal rights in certain areas of social life. There is a lot of confusion in the answer choices for questions at the application level (\ref{VNHSGE_Civ_A}), making it difficult to accurately assess and choose the right response. Complex case studies with several plotlines and characters are offered for questions at the high level (\ref{VNHSGE_Civ_H}), and it needs a thorough comprehension of legal theory to properly examine the nature of the characters' violations.
	
	For LLMs to assess their understanding of and ability to apply legal information, particularly in the context of civic education and legal training, the VNHSGE civic education dataset is employed. The dataset includes case studies together with multiple-choice questions on topics like legal frameworks and regulations, fundamental citizen rights, democratic principles, and notions. LLMs can gain a better understanding of legal ideas and how to apply them in practical scenarios by training on this dataset, which can be helpful for a range of applications like legal research, automated legal document analysis, and legal chatbots.
	
	\section{Experiments}
	\label{sec:experiments}
	
	\subsection{ChatGPT and BingChat responses}

	\textbf{Response format}: When posing questions to LLMs, we can receive answers in various formats. To standardize response formats and simplify result processing, we request that LLMs provide replies in a specific structure. Figure~\ref{fig:chatbot_response} demonstrates an example of the required structure for LLM responses. To achieve this, we used the Explanation and Choice approach and include a "pre-question" prompt before the actual question. This prompt combines the content of the original question with instructions for the desired response format. Standardizing the format of LLM answers is crucial for several reasons. Firstly, it enables quicker and more accurate processing of model responses. Secondly, it facilitates impartial comparison and evaluation of the performance of different LLMs. Additionally, it ensures that the solutions provided by LLMs are easy to understand and applicable for further applications. By giving LLM responses a clear and consistent structure, we can effectively harness their abilities to enhance various NLP tasks.
	
	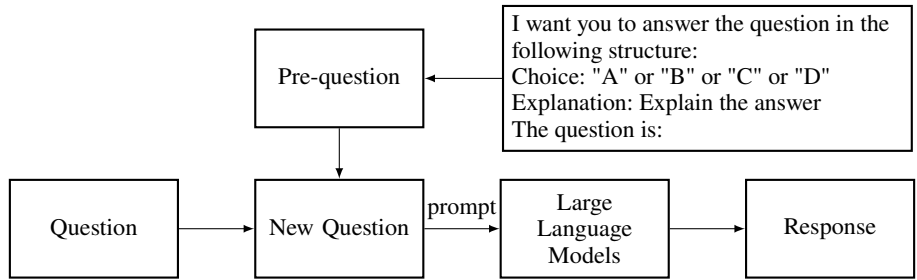
\begin{figure}[ht!]
		\begin{center}
			\begin{tikzpicture}  
				\node[block] (a) {Question};  
				\node[block,right=of a] (b) {New Question};   
				\node[block,right=of b] (c) {Large Language Models};  
				\node[block,right=of c] (d) {Response}; 
				\node[blocktext,align=flush left] (f) at ([yshift=2cm]$(c)!0.5!(d)$) {I want you to answer the question in the following structure:  \\ 
					Choice: "A" or "B" or "C" or "D"  \\
					Explanation: Explain the answer \\
					The question is: {}
				};   
				\node[block] (h) at ([yshift=2cm]$(b)!1.0!(b)$) {Pre-question}; 
				\draw[line] (a)-- (b);  
				\draw[line] (b)-- (c) node [midway, above, sloped] (TextNode) {prompt};  
				\draw[line] (c)-- (d);  
				\draw[line] (h)-- (b);  
				\draw[line] (f)-- (h);  
			\end{tikzpicture} 
		\end{center}
		\caption{Formatted question and LLMs response.}
		\label{fig:chatbot_response}
	\end{figure}
	
	\begin{mdframed}[linewidth=1pt,linecolor=red]
		\textbf{Question (Word format)}:
		
		\begingroup
		\renewcommand{\arraystretch}{1.25} 
		\begin{tabular}{|l|l|l|l|l|l|}
			\hline
			ID & IQ & Q   & C & IA & E           \\ \hline
			1  &  & \begin{tabular}[c]{@{}l@{}}1) The volume of a cube with edge 2a is:\\ A. 8a\textasciicircum3 \\ B. 2a\textasciicircum3. \\ C. a\textasciicircum3 \\ D. 6a\textasciicircum3.\end{tabular} & A      &        & \begin{tabular}[c]{@{}l@{}}The volume of a cube \\ with edge 2a is: \\ V=(2a)\textasciicircum3=8a\textasciicircum3.\end{tabular} \\ \hline
		\end{tabular}
		\endgroup
		
		\textbf{Question (JSON format)}:
		\{
		"ID": "Q1",
		"IQ": " ",
		"Q": "1) The volume of a cube with edge 2a is:\textbackslash{}nA. 8a\textasciicircum3.\textbackslash{}t\textbackslash{}nB. 2a\textasciicircum3.\textbackslash{}t\textbackslash{}nC. a\textasciicircum3.\textbackslash{}t\textbackslash{}nD. 6a\textasciicircum3.",
		"C": "A",
		"IA": " ",
		"E": "The volume of a cube with edge 2a is: V=(2a)\textasciicircum3=8a\textasciicircum3.",
		\}
		
		\textbf{Pre-question (JSON format)}: "I want you to answer the question in the following structure: " 
		"\textbackslash{}nChoice: "A" or "B" or "C" or "D" " 
		"\textbackslash{}nExplanation: Explain the answer" 
		"\textbackslash{}nThe question is: \{\}"

		\textbf{New Question (Prompt)}:
		I want you to answer the question in the following structure:    
		
		Choice: "A" or "B" or "C" or "D"
		
		Explanation: Explain the answer
		
		The question is: 1) The volume of a cube with edge 2a is:
		A. 8a\textasciicircum3. B. 2a\textasciicircum3. C. a\textasciicircum3. D. 6a\textasciicircum3."
		
		\textbf{Response (JSON format)}:
		
		\{
		"ID": "1",
		"IQ": " ",
		"Q": "1) The volume of a cube with edge 2a is:\textbackslash{}nA. 8a\textasciicircum3.\textbackslash{}t\textbackslash{}nB. 2a\textasciicircum3.\textbackslash{}t\textbackslash{}nC. a\textasciicircum3.\textbackslash{}t\textbackslash{}nD. 6a\textasciicircum3.",
		"C": "A",
		"IA": " ",
		"E": "The volume of a cube with edge 2a is: V=(2a)\textasciicircum3=8a\textasciicircum3.",
		"CC": "A",
		"CE": "The formula for the volume of a cube is V = s\textasciicircum3, where s is the length of one of its sides. Therefore, the volume of the cube with a side length of 2a is: V = (2a)\textasciicircum3 = 8a\textasciicircum3",
		\}
		
		\textbf{Response (Word format)}:
		
		\begingroup
		\renewcommand{\arraystretch}{1.25} 
		\begin{tabular}{|l|l|l|l|l|l|l|l|}
			\hline
			ID & IQ & Q                                                                                                                                       & C & IA & E                                                                                                    & CC & CE                                                                                                                                                                                                                                      \\ \hline
			1  &    & \begin{tabular}[c]{@{}l@{}}1) The volume of a cube \\ with edge 2a is:\\ A. 8a\textasciicircum3 \\ B. 2a\textasciicircum3. \\ C. a\textasciicircum3 \\ D. 6a\textasciicircum3.\end{tabular} & A &    & \begin{tabular}[c]{@{}l@{}}The volume of a cube \\ with edge 2a is: \\ V=(2a)\textasciicircum3=8a\textasciicircum3.\end{tabular} & A  & \begin{tabular}[c]{@{}l@{}}The formula for the volume of \\ a cube is V = s\textasciicircum3, where s is \\ the length of one of its sides. \\ Therefore, the volume of \\ the cube with a side length \\ of 2a is: V = (2a)\textasciicircum3 = 8a\textasciicircum3\end{tabular} \\ \hline
		\end{tabular}
		\endgroup
		
	\end{mdframed}	
	
	We conducted experiments using two state of the art language models, ChatGPT (based on GPT-3.5) and BingChat (based on GPT-4)\footnote{\href{https://blogs.bing.com/search/march_2023/Confirmed-the-new-Bing-runs-on-OpenAI\%E2\%80\%99s-GPT-4}{https://blogs.bing.com/search/march\_2023/Confirmed-the-new-Bing-runs-on-OpenAI\%E2\%80\%99s-GPT-4}}, to evaluate the performance of our dataset. We assessed each model based on accuracy and provided examples of both successful and poor responses (see Appendix section \ref{Dataset_response} for further details of examples). 
	
	In the following sections, we compared the performance of ChatGPT and BingChat using five tests for each subject, including 30 literary essays and 1700 multiple-choice questions in others. LLMs like ChatGPT and BingChat have been trained to predict the next word in a text based on the preceding words. 
	However, these models have limitations when it comes to handling complex computational problems or requiring multi-step reasoning, even though they are capable of responding to basic questions. These LLMs may also struggle to comprehend texts with intricate contexts and may encounter difficulties in certain situations, particularly when processing the Vietnamese language. They might misinterpret certain contexts and occasionally confuse words with homonyms or antonyms.

	\textbf{Mathematics}: ChatGPT and BingChat can handle knowledge and comprehension level questions (\ref{VNHSGE_Math_K}) and (\ref{VNHSGE_Math_C}). However, they struggle with complex calculations and logical reasoning that require advanced mathematical skills or multi-step deductive reasoning (\ref{VNHSGE_Math_A}). These models often provide inaccurate explanations and answers and are unable to provide appropriate solution instructions for high application level problems (\ref{VNHSGE_Math_H}).

	\textbf{Literature}: ChatGPT and BingChat are capable of responding to literary queries and generating essays due to their extensive training in various domains, including literature and journalism. They have a good grasp of natural language structure and can synthesize new responses and paragraphs based on learned knowledge and input data. However, ChatGPT and BingChat still have limitations in reasoning abilities and understanding complex language and context, particularly in languages like Vietnamese. As a result, their responses may not always be entirely accurate or suitable for the context or purpose of the question (\ref{VNHSGE_Lit_Q}). ChatGPT is more suitable for language-related topics and tends to provide more relevant and emotive responses compared to BingChat, a search engineer (\ref{VNHSGE_Lit_S}).

	\textbf{English}: ChatGPT and BingChat are unable to respond to questions on pronunciation and stress (\ref{VNHSGE_Eng_PS}) even though they both score well on other English languages topics like grammar and vocabulary (\ref{VNHSGE_Eng_G}), communication (\ref{VNHSGE_Eng_C}), reading fill-in-the-blank (\ref{VNHSGE_Eng_RB}), and reading comprehension (\ref{VNHSGE_Eng_RC}). Both ChatGPT and BingChat have been taught the rules and patterns of the English language, including grammar and vocabulary, through training on large English text data. Additionally, they receive instruction on how to comprehend and produce natural language, which involves reading fill-in-the-blank passages and reading comprehension. Though it's possible that neither BingChat nor ChatGPT received adequate training in pronunciation and stress.

	\textbf{Physics}: ChatGPT and BingChat can solve physics questions at the knowledge and comprehension levels (\ref{VNHSGE_Phy_A} and \ref{VNHSGE_Phy_C}) which are relatively simple questions about physics topics. However, they are unable to answer questions at the application and high application levels (\ref{VNHSGE_Phy_A} and \ref{VNHSGE_Phy_H}), which frequently call for substantial knowledge and skills in understanding and applying concepts to solve problems.

	\textbf{Chemistry}: ChatGPT and BingChat can respond to questions at the knowledge level (\ref{VNHSGE_Che_K}) by memorizing facts. They often fail to generate the right response to questions at the comprehension level (\ref{VNHSGE_Che_C}). Neither ChatGPT nor BingChat typically can provide accurate answers for challenging questions at the application level (\ref{VNHSGE_Che_A}) and high application level (\ref{VNHSGE_Che_H}) because these types of questions demand the capacity to infer from multiple chemical reactions and high-level synthesis knowledge.

	\textbf{Biology}: Both ChatGPT and BingChat are capable of providing responses to questions at the knowledge and comprehension levels (\ref{VNHSGE_Bio_K} and \ref{VNHSGE_Bio_C}), similar to subjects like mathematics, physics, and chemistry that require both calculation and reasoning skills. However, ChatGPT and BingChat have a very limited likelihood of correctly determining the answers to questions requiring complex thinking and information processing in diagrams at the application and high application levels (\ref{VNHSGE_Bio_C} and \ref{VNHSGE_Bio_H}). These types of questions demand a deeper understanding of biology concepts and the ability to apply them in complex scenarios.

	\textbf{History}: ChatGPT and BingChat do reasonably well when answering questions in the field of history at the knowledge and comprehension levels (\ref{VNHSGE_His_K} and \ref{VNHSGE_His_C}). However,  both ChatGPT and BingChat often struggle to provide accurate responses to the application and high application questions (\ref{VNHSGE_His_A} and \ref{VNHSGE_His_H}). These types of questions require higher-order thinking skills and a deep understanding of the historical context as well as demand the ability to compare, analyze, and express a judgment on historical events and characters.

	\textbf{Geography}: ChatGPT is able to respond to questions about charts without requesting data from the chart, BingChat does not support these questions (\ref{VNHSGE_Geo_Chart}). The result is that both they cannot answer questions related to charts or images. Both ChatGPT and BingChat can provide precise responses to questions about the information in a table (\ref{VNHSGE_Geo_Table}) and queries related to the use of the Atlas (\ref{VNHSGE_Geo_Atlas}). However, when it comes to questions that require analysis and interpretation at the application and high application levels (\ref{VNHSGE_Geo_Complex}), both ChatGPT and BingChat often struggle to give precise responses. These types of questions necessitate the ability to analyze and interpret geographical data and concepts, which the models may find challenging.

	\textbf{Civic Education}: At the knowledge and comprehension levels (\ref{VNHSGE_Civ_K} and \ref{VNHSGE_Civ_C}), ChatGPT and BingChat can provide accurate answers. However, ChatGPT often produces inaccurate responses for questions at the application level (\ref{VNHSGE_Civ_A}), while BingChat performs better. Both ChatGPT and BingChat often fail to provide precise responses when analyzing character behavior in scenario-based questions at the high application level (\ref{VNHSGE_Civ_H}).
	
	\subsection{ChatGPT and BingChat performances} 
	
	Table~\ref{tabl:llm_result} displays ChatGPT and BingChat's performance. We can see that for subjects requiring complex computation and reasoning, such as mathematics, physics, chemistry, and biology, their performance ranges from $48\%$ to $69\%$. The performance of ChatGPT and BingChat is between $56.5\%$ and $92.4\%$ for subjects that predominantly depend on languages, such as literature, English, history, geography, and civic education. LLMs such as ChatGPT and BingChat have been trained on vast amounts of text covering a wide range of fields. However, these models lack subject-matter expertise. Mathematics, physics, chemistry, and biology often demand profound knowledge and advanced computational abilities, which may not be possessed by language models like ChatGPT and BingChat for solving such challenging problems. On the other hand, subjects like literature, English, history, geography, and civic education frequently require strong language skills and the ability to comprehend complex texts, areas in which language models like ChatGPT and BingChat may have sufficient capabilities to handle.
	
	\begingroup
	\renewcommand{\arraystretch}{1.5} 
	\begin{table}[ht!]
		\caption{ChatGPT and BingChat performances on VNHSGE dataset}
		\label{tabl:llm_result}
		\resizebox{\textwidth}{!}{%
			\begin{tabular}{c|cc|cc|cc|cc|cc|cc|cc|cc|cc|}
				\cline{2-19}
				& \multicolumn{2}{c|}{\textbf{Mathematics}}               & \multicolumn{2}{c|}{\textbf{Literature}}              & \multicolumn{2}{c|}{\textbf{English}}                  & \multicolumn{2}{c|}{\textbf{Physics}}              & \multicolumn{2}{c|}{\textbf{Chemistry}}                & \multicolumn{2}{c|}{\textbf{Biology}}              & \multicolumn{2}{c|}{\textbf{History}}                  & \multicolumn{2}{c|}{\textbf{Geography}}                  & \multicolumn{2}{c|}{\textbf{Civic Education}}                  \\ \hline
				\multicolumn{1}{|c|}{\textbf{2019}} & \multicolumn{1}{c|}{52}            & 56          & \multicolumn{1}{c|}{75}          & 52.75          & \multicolumn{1}{c|}{76}            & 92            & \multicolumn{1}{c|}{60}          & 55          & \multicolumn{1}{c|}{40}          & 55            & \multicolumn{1}{c|}{60}          & 67.5        & \multicolumn{1}{c|}{42.5}          & 82.5          & \multicolumn{1}{c|}{50}            & 75            & \multicolumn{1}{c|}{60}            & 75            \\ \hline
				\multicolumn{1}{|c|}{\textbf{2020}} & \multicolumn{1}{c|}{66}            & 56          & \multicolumn{1}{c|}{68.9}        & 51.25        & \multicolumn{1}{c|}{86}            & 96            & \multicolumn{1}{c|}{62.5}        & 67.5        & \multicolumn{1}{c|}{42.5}        & 57.5          & \multicolumn{1}{c|}{60}          & 72.5        & \multicolumn{1}{c|}{47.5}          & 85            & \multicolumn{1}{c|}{52.5}          & 70            & \multicolumn{1}{c|}{70}            & 87.5          \\ \hline
				\multicolumn{1}{|c|}{\textbf{2021}} & \multicolumn{1}{c|}{60}            & 66          & \multicolumn{1}{c|}{75}          & 60.25          & \multicolumn{1}{c|}{76}            & 86            & \multicolumn{1}{c|}{60}          & 67.5        & \multicolumn{1}{c|}{62.5}        & 50            & \multicolumn{1}{c|}{52.5}        & 67.5        & \multicolumn{1}{c|}{55}            & 90            & \multicolumn{1}{c|}{75}            & 82.5          & \multicolumn{1}{c|}{62.5}          & 92.5          \\ \hline
				\multicolumn{1}{|c|}{\textbf{2022}} & \multicolumn{1}{c|}{62}            & 60          & \multicolumn{1}{c|}{56.3}        & 70        & \multicolumn{1}{c|}{80}            & 94            & \multicolumn{1}{c|}{65}          & 67.5        & \multicolumn{1}{c|}{47.5}        & 47.5          & \multicolumn{1}{c|}{57.5}        & 72.5        & \multicolumn{1}{c|}{60}            & 92.5          & \multicolumn{1}{c|}{62.5}          & 85            & \multicolumn{1}{c|}{82.5}          & 90            \\ \hline
				\multicolumn{1}{|c|}{\textbf{2023}} & \multicolumn{1}{c|}{54}            & 62          & \multicolumn{1}{c|}{64.8}        & 49.75        & \multicolumn{1}{c|}{78}            & 94            & \multicolumn{1}{c|}{57.5}        & 72.5        & \multicolumn{1}{c|}{47.5}        & 52.5          & \multicolumn{1}{c|}{60}          & 65          & \multicolumn{1}{c|}{77.5}          & 92.5          & \multicolumn{1}{c|}{67.5}          & 85            & \multicolumn{1}{c|}{77.5}          & 82.5          \\ \hline
				\multicolumn{1}{|c|}{\textbf{AVG}}  & \multicolumn{1}{c|}{\textbf{58.8}} & \textbf{60} & \multicolumn{1}{c|}{\textbf{68}} & \textbf{56.8} & \multicolumn{1}{c|}{\textbf{79.2}} & \textbf{92.4} & \multicolumn{1}{c|}{\textbf{61}} & \textbf{66} & \multicolumn{1}{c|}{\textbf{48}} & \textbf{52.8} & \multicolumn{1}{c|}{\textbf{58}} & \textbf{69} & \multicolumn{1}{c|}{\textbf{56.5}} & \textbf{88.5} & \multicolumn{1}{c|}{\textbf{61.5}} & \textbf{79.5} & \multicolumn{1}{c|}{\textbf{70.5}} & \textbf{85.5} \\ \hline
			\end{tabular}
		}
	\end{table}
	\endgroup
	
	The performance comparison between ChatGPT and BingChat is depicted in Figure~\ref{fig:ChatGPT_BingChat_exam}. BingChat performs better than ChatGPT in all categories except for literature. There is not much difference between BingChat and ChatGPT in subjects like mathematics, physics, and chemistry, which require extensive computation and reasoning. However, ChatGPT surpasses BingChat in terms of performance in the literature category. This is because BingChat is a search engine, and its results may not be suitable for the literature subject, which often involves writing extensive essays. BingChat outperforms ChatGPT in the remaining topics. It should be noted that BingChat is based on GPT-4 while ChatGPT is based on GPT-3.5. Furthermore, BingChat may find accurate answers when the questions and answers are publicly available online.

	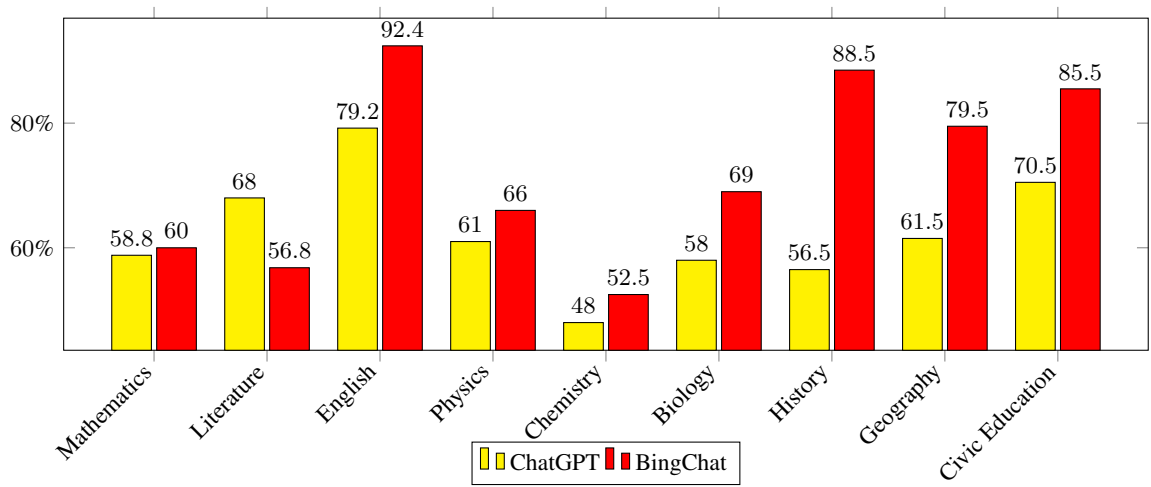
\begin{figure}[h!]
		\begin{center}
			\begin{tikzpicture}  
				
				\begin{axis}  
					[  
					ybar, 
					bar width=15pt, 
					legend style={at={(0.5,-0.275)}, 	
						anchor=north,legend columns=-1},    
					symbolic x coords={Mathematics, Literature, English, Physics, Chemistry, Biology, History, Geography, Civic Education}, 
					xtick=data,
					x tick label style={rotate=45,anchor=east},
					yticklabel={\pgfmathprintnumber{\tick}\%},  
					nodes near coords,   
					nodes near coords align={vertical}, 
					enlarge x limits,
					height=6cm, width=16cm, 
					]  
					\addplot [fill=yellow] coordinates {
						(Mathematics, 58.8)
						(Literature, 68)
						(English, 79.2)
						(Physics, 61)
						(Chemistry, 48)
						(Biology, 58)
						(History,56.5)
						(Geography, 61.5)
						(Civic Education, 70.5)
						
					}; 
					\addplot [fill=red] coordinates {
						(Mathematics, 60)
						(Literature, 56.8)
						(English, 92.4)
						(Physics, 66)
						(Chemistry, 52.5)
						(Biology, 69)
						(History,88.5)
						(Geography, 79.5)
						(Civic Education, 85.5)
						
					}; 
					\legend{
						ChatGPT,
						BingChat,}
					
				\end{axis}  
			\end{tikzpicture}
		\end{center}
		\caption{Comparison of ChatGPT and BingChat performances on VNHSGE dataset.}
		\label{fig:ChatGPT_BingChat_exam}
	\end{figure}

	\subsection{ChatGPT, BingChat, and Vietnamese Students}
	
	This section compares the effectiveness of BingChat and ChatGPT with Vietnamese students. Our aim is to determine whether LLMs possess abilities comparable to those of humans, although this comparison is challenging due to the dissimilar settings. By conducting this comparison, we can evaluate whether LLMs can serve as effective tools for Vietnamese students in various subject areas (see Appendix section~\ref{sec:Score Spectrum} for more details of spectrum comparisons).
	
	Figure~\ref{fig:core subjects} illustrates a comparison of the performance among ChatGPT, BingChat, and Vietnamese students in three core subjects: mathematics (\ref{sec:Score Spectrum_Math}), literature (\ref{sec:Score Spectrum_Lit}), and English (\ref{sec:Score Spectrum_Eng}). These subjects are integral parts of the exam and are required for all students. 
	
	\begin{figure}[ht!]
		\begin{center}
			\begin{tikzpicture}  
				\begin{axis}  
					[  
					ybar, 
					bar width=15pt,
					enlargelimits=0.15, 
					legend style={at={(0.5,-0.175)}, 	
						anchor=north,legend columns=-1},       
					ylabel={Mathematics Score}, 
					symbolic x coords={
						2019,
						2020,
						2021,
						2022,
						2023,
					},  
					xtick=data,  
					nodes near coords,  
					nodes near coords align={vertical},  
					width=\textwidth, 
					height=5cm, 
					width=16cm,
					]  
					\addplot [fill=yellow] coordinates {
						(2019,5.2)
						(2020,6.6)
						(2021,6)
						(2022,6.2)
						(2023,5.4)
					};
					\addplot [fill=red] coordinates {
						(2019,5.6)
						(2020,5.6)
						(2021,6.6)
						(2022,6)
						(2023,6.2)						
					};
					\addplot [fill=green] coordinates {
						(2019,5.64)
						(2020,6.67)
						(2021,6.61)
						(2022,6.47)
						(2023,)		
					};
					\addplot [fill=blue] coordinates {
						(2019,6.4)
						(2020,7.8)
						(2021,7.8)
						(2022,7.8)
						(2023,)
						
					};
				\end{axis}  
			\end{tikzpicture}
			\begin{tikzpicture}  
				\begin{axis}  
					[  
					ybar, 
					bar width=15pt,
					enlargelimits=0.15, 
					legend style={at={(0.5,-0.175)}, 	
						anchor=north,legend columns=-1},       
					ylabel={Literature Score}, 
					symbolic x coords={
						2019,
						2020,
						2021,
						2022,
						2023,
					},  
					xtick=data,  
					nodes near coords,  
					nodes near coords align={vertical},  
					width=\textwidth, 
					height=5cm, 
					width=16cm,
					]  
					\addplot [fill=yellow] coordinates {
						(2019,7.5)
						(2020,6.89)
						(2021,7.5)
						(2022,5.63)
						(2023,6.48)
						
					};
					\addplot [fill=red] coordinates {
						(2019,5.28)
						(2020,5.13)
						(2021,6.03)
						(2022,7)
						(2023,5)
						
					};
					\addplot [fill=green] coordinates {
						(2019,5.49)
						(2020,6.61)
						(2021,6.47)
						(2022,6.51)
						(2023,)
						
					};
					\addplot [fill=blue] coordinates {
						(2019,6)
						(2020,7)
						(2021,7)
						(2022,7)
						(2023,)
						
					};
				\end{axis}  
			\end{tikzpicture}
			\begin{tikzpicture}  
				\begin{axis}  
					[  
					ybar, 
					bar width=15pt,
					enlargelimits=0.15, 
					legend style={at={(0.5,-0.175)}, 	
						anchor=north,legend columns=-1},       
					ylabel={English Score}, 
					symbolic x coords={
						2019,
						2020,
						2021,
						2022,
						2023,
					},  
					xtick=data,  
					nodes near coords,  
					nodes near coords align={vertical},  
					width=\textwidth, 
					height=5cm, 
					width=16cm,
					]  
					\addplot [fill=yellow] coordinates {
						(2019,7.6)
						(2020,8.6)
						(2021,7.6)
						(2022,8)
						(2023,7.8)
					};
					\addplot [fill=red] coordinates {
						(2019,9.2)
						(2020,9.6)
						(2021,8.6)
						(2022,9.4)
						(2023,9.4)						
					};
					\addplot [fill=green] coordinates {
						(2019,4.36)
						(2020,4.58)
						(2021,5.84)
						(2022,5.15)
						(2023,)							
					};
					\addplot [fill=blue] coordinates {
						(2019,3.2)
						(2020,3.4)
						(2021,4)
						(2022,3.8)
						(2023,)
					};
					\legend{ChatGPT, BingChat, AVS, MVS} 
				\end{axis}  
			\end{tikzpicture}  
		\end{center}
		\caption{Comparison in core subjects.}
		\label{fig:core subjects}
	\end{figure}
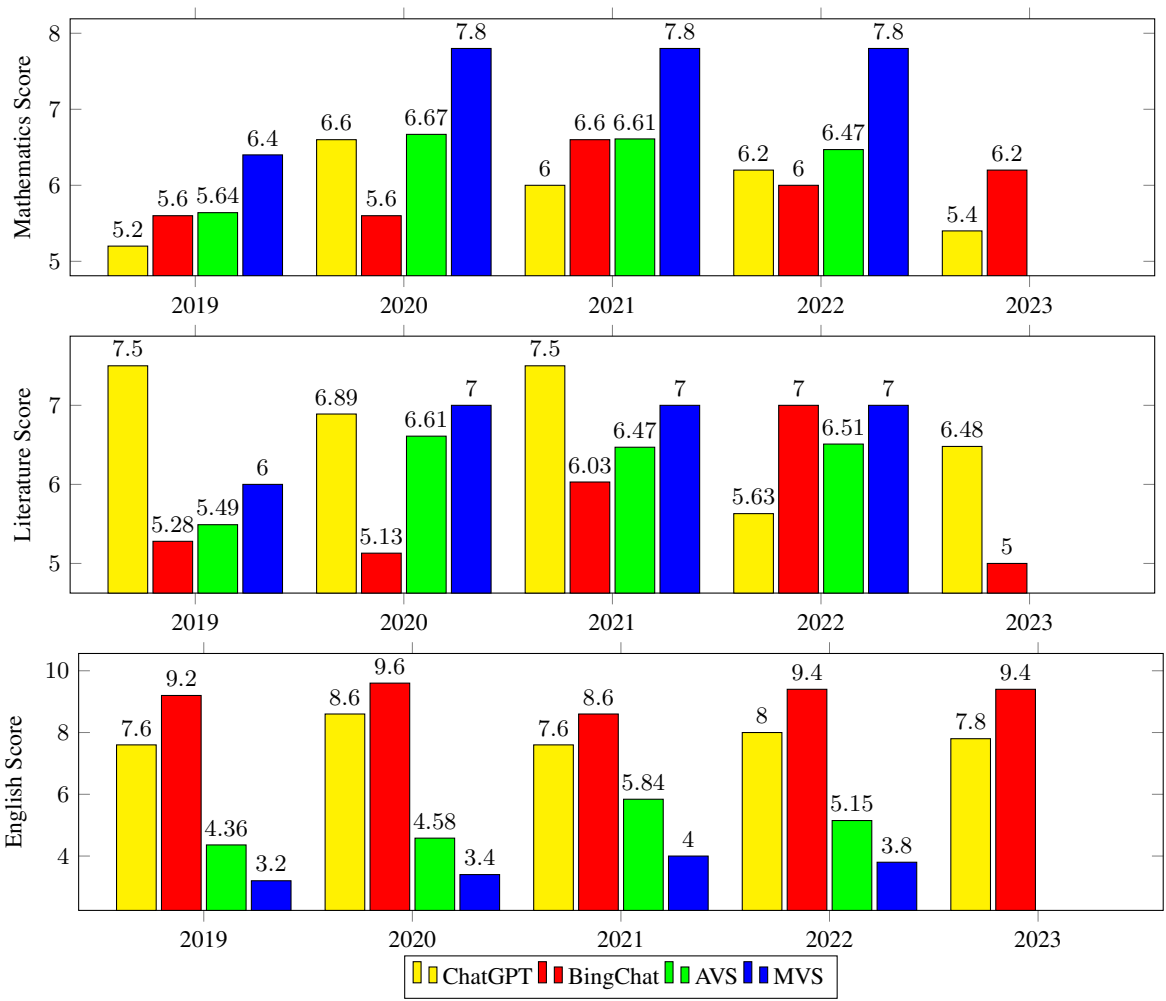

	\textbf{Mathematics}: According to the findings, ChatGPT and BingChat are unable to match the performance of human students in Vietnam's high school mathematics curriculum. Despite being trained on vast amounts of textual data from the internet, they struggle with complex mathematical problems, although they can handle simpler mathematical concepts. The high school mathematics questions require reasoning, logical thinking, analytical skills, and the ability to apply knowledge in practical situations. To achieve performance on par with humans in high school mathematics, ChatGPT and BingChat's mathematical abilities need substantial improvement.
	
	\textbf{Literature}: Both ChatGPT and BingChat have been extensively trained on large Vietnamese language datasets, enabling them to analyze and generate essays with considerable proficiency. In terms of high school literature, the performance of LLMs such as ChatGPT and BingChat is human-like level. However, it should be emphasized that ChatGPT and BingChat are unable to write emotionally rich essays or conduct in-depth literary analyses. In summary, ChatGPT can be considered a tool to support Vietnamese students in studying literature.
	
	\textbf{English}: According to the results, ChatGPT and BingChat performed better in high school English compared to Vietnamese students. It should be mentioned that Vietnamese students' English proficiency is not very high compared to the global average. ChatGPT and BingChat are effective tools that Vietnamese students can utilize to study foreign languages.
	
	Figure~\ref{fig:natural combination} depicts a comparison of the performance among ChatGPT, BingChat, and Vietnamese students in the natural combination, including physics (\ref{sec:Score Spectrum_Phy}), chemistry (\ref{sec:Score Spectrum_Che}), and biology (\ref{sec:Score Spectrum_Bio}), respectively.

	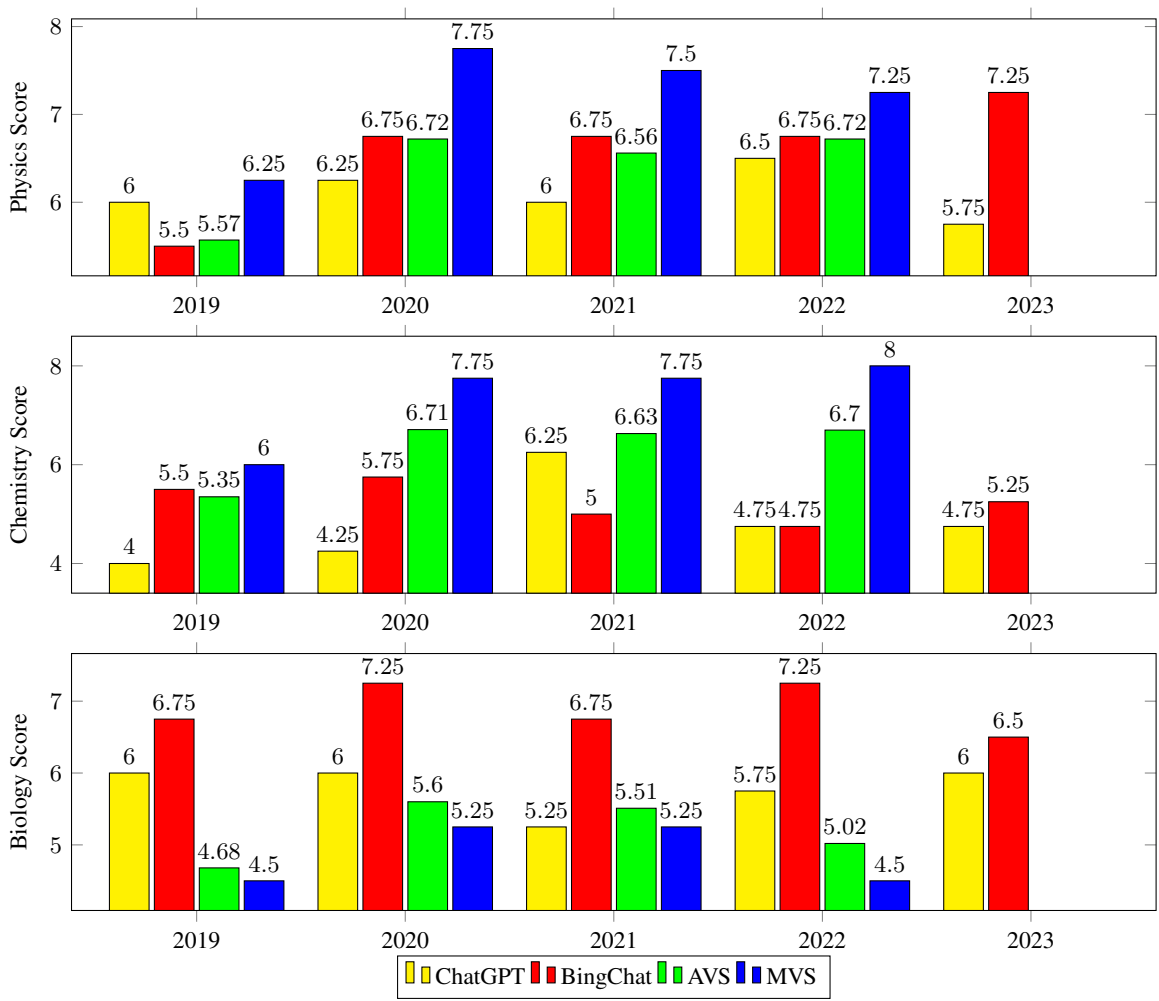
\begin{figure}[ht!]
		\begin{center}
			\begin{tikzpicture}  
				\begin{axis}  
					[  
					ybar, 
					bar width=15pt,
					enlargelimits=0.15, 
					legend style={at={(0.5,-0.1775)}, 	
						anchor=north,legend columns=-1},          
					ylabel={Physics Score}, 
					symbolic x coords={
						2019,
						2020,
						2021,
						2022,
						2023,
					},  
					xtick=data,  
					nodes near coords,  
					nodes near coords align={vertical},  
					width=\textwidth, 
					height=5cm, 
					width=16cm,
					]  
					
					\addplot [fill=yellow] coordinates {
						(2019,6.0)
						(2020,6.25)
						(2021,6.0)
						(2022,6.5)
						(2023,5.75)
					};
					\addplot [fill=red] coordinates {
						(2019,5.5)
						(2020,6.75)
						(2021,6.75)
						(2022,6.75)
						(2023,7.25)
						
					};
					\addplot [fill=green] coordinates {
						(2019,5.57)
						(2020,6.72)
						(2021,6.56)
						(2022,6.72)
						(2023,)
						
					};
					\addplot [fill=blue] coordinates {
						(2019,6.25)
						(2020,7.75)
						(2021,7.5)
						(2022,7.25)
						(2023,)
						
					};
					
				\end{axis}  
			\end{tikzpicture} 
			\begin{tikzpicture}  
				\begin{axis}  
					[  
					ybar, 
					bar width=15pt,
					enlargelimits=0.15, 
					legend style={at={(0.5,-0.175)}, 	
						anchor=north,legend columns=-1},       
					ylabel={Chemistry Score}, 
					symbolic x coords={
						2019,
						2020,
						2021,
						2022,
						2023,
					},  
					xtick=data,  
					nodes near coords,  
					nodes near coords align={vertical},  
					width=\textwidth, 
					height=5cm, 
					width=16cm,
					]  
					\addplot [fill=yellow] coordinates {
						(2019,4)
						(2020,4.25)
						(2021,6.25)
						(2022,4.75)
						(2023,4.75)
					};
					\addplot [fill=red] coordinates {
						(2019,5.5)
						(2020,5.75)
						(2021,5)
						(2022,4.75)
						(2023,5.25)						
					};
					\addplot [fill=green] coordinates {
						(2019,5.35)
						(2020,6.71)
						(2021,6.63)
						(2022,6.7)
						(2023,)
						
					};
					\addplot [fill=blue] coordinates {
						(2019,6)
						(2020,7.75)
						(2021,7.75)
						(2022,8)
						(2023,)

					};
				\end{axis}  
			\end{tikzpicture} 
			\begin{tikzpicture}  
				\begin{axis}  
					[  
					ybar, 
					bar width=15pt,
					enlargelimits=0.15, 
					legend style={at={(0.5,-0.175)}, 	
						anchor=north,legend columns=-1},       
					ylabel={Biology Score}, 
					symbolic x coords={
						2019,
						2020,
						2021,
						2022,
						2023,
					},  
					xtick=data,  
					nodes near coords,  
					nodes near coords align={vertical},  
					width=\textwidth, 
					height=5cm, 
					width=16cm,
					]  
					\addplot [fill=yellow] coordinates {
						(2019,6)
						(2020,6)
						(2021,5.25)
						(2022,5.75)
						(2023,6)
					};
					\addplot [fill=red] coordinates {
						(2019,6.75)
						(2020,7.25)
						(2021,6.75)
						(2022,7.25)
						(2023,6.5)						
					};
					\addplot [fill=green] coordinates {
						(2019,4.68)
						(2020,5.6)
						(2021,5.51)
						(2022,5.02)
						(2023,)					
					};
					\addplot [fill=blue] coordinates {
						(2019,4.5)
						(2020,5.25)
						(2021,5.25)
						(2022,4.5)
						(2023,)
					};
					\legend{ChatGPT, BingChat, AVS, MVS} 
				\end{axis}  
			\end{tikzpicture} 
		\end{center}
		\caption{Comparison in natural combination}
		\label{fig:natural combination}
	\end{figure}
	
	\textbf{Physics}: The performance of ChatGPT and BingChat is comparable to the average score of Vietnamese students in physics. However, they are still less than the score achieved by most Vietnamese students. With thorough training in the field of physics, LLMs can provide accurate answers and insightful explanations to assist students in understanding physics. The models, however, still require development, particularly for physics issues that call for intricate computations and reasoning.
	
	\textbf{Chemistry}: ChatGPT and BingChat still do not possess the same level of proficiency in chemistry as Vietnamese high school students do. While these LLMs can provide relevant knowledge and solutions in the field of chemistry, they lack the expertise required to solve complex chemistry problems that demand advanced levels of analysis and reasoning. However, in terms of delivering theoretical knowledge and information, it is certainly possible for LLMs to become useful tools for Vietnamese students in high school chemistry.
	
	\textbf{Biology}: The findings indicate that ChatGPT and BingChat outperform Vietnamese students in biology. It is important to note that biology is considered a less prioritized subject for many Vietnamese students compared to mathematics, physics, and chemistry. The biology score of Vietnamese students is less in mathematics, physics, and chemistry. LLMs are capable of addressing basic questions in biology, such as definitions, concepts, simple problem-solving, and specific examples. Therefore, LLMs can serve as helpful resources for high school students to comprehend fundamental biology concepts and problems.
	
	Figure~\ref{fig:social combination} presents a comparison of the performance among ChatGPT, BingChat, and Vietnamese students in the social combination, including history~(\ref{sec:Score Spectrum_His}), geography~(\ref{sec:Score Spectrum_Geo}), and civic education~(\ref{sec:Score Spectrum_Civ}), respectively. 
	
	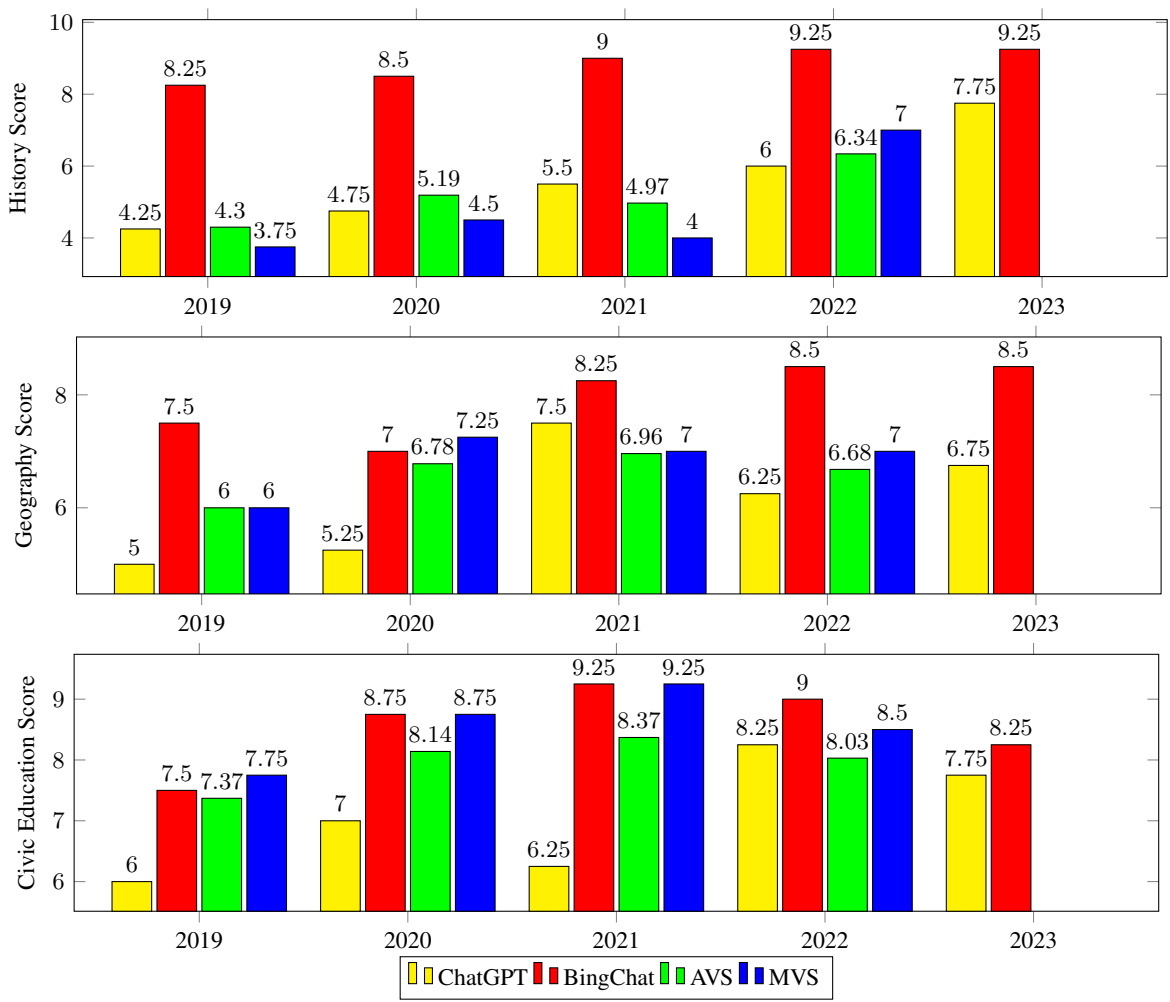
\begin{figure}[ht!]
		\begin{center}
			\begin{tikzpicture}  
				\begin{axis}  
					[  
					ybar, 
					bar width=15pt,
					enlargelimits=0.15, 
					legend style={at={(0.5,-0.175)}, 	
						anchor=north,legend columns=-1},       
					ylabel={History Score}, 
					symbolic x coords={
						2019,
						2020,
						2021,
						2022,
						2023,
					},  
					xtick=data,  
					nodes near coords,  
					nodes near coords align={vertical},  
					width=\textwidth, 
					height=5cm, 
					width=16cm,
					]  
					\addplot [fill=yellow] coordinates {
						(2019,4.25)
						(2020,4.75)
						(2021,5.5)
						(2022,6)
						(2023,7.75)	
					};
					\addplot [fill=red] coordinates {
						(2019,8.25)
						(2020,8.5)
						(2021,9)
						(2022,9.25)
						(2023,9.25)						
					};
					\addplot [fill=green] coordinates {
						(2019,4.3)
						(2020,5.19)
						(2021,4.97)
						(2022,6.34)
						(2023,)		
					};
					\addplot [fill=blue] coordinates {
						(2019,3.75)
						(2020,4.5)
						(2021,4)
						(2022,7)
						(2023,)
					};
				\end{axis}  
			\end{tikzpicture}
			\begin{tikzpicture}  
				\begin{axis}  
					[  
					ybar, 
					bar width=15pt,
					enlargelimits=0.15, 
					legend style={at={(0.5,-0.175)}, 	
						anchor=north,legend columns=-1},       
					ylabel={Geography Score}, 
					symbolic x coords={
						2019,
						2020,
						2021,
						2022,
						2023,
					},  
					xtick=data,  
					nodes near coords,  
					nodes near coords align={vertical},  
					width=\textwidth, 
					height=5cm, 
					width=16cm,
					]  
					\addplot [fill=yellow] coordinates {
						(2019,5)
						(2020,5.25)
						(2021,7.5)
						(2022,6.25)
						(2023,6.75)
						
					};
					\addplot [fill=red] coordinates {
						(2019,7.5)
						(2020,7)
						(2021,8.25)
						(2022,8.5)
						(2023,8.5)
						
					};
					\addplot [fill=green] coordinates {
						(2019,6)
						(2020,6.78)
						(2021,6.96)
						(2022,6.68)
						(2023,)
						
					};
					\addplot [fill=blue] coordinates {
						(2019,6)
						(2020,7.25)
						(2021,7)
						(2022,7)
						(2023,)

					};
				\end{axis}  
			\end{tikzpicture} 
			\begin{tikzpicture}  
				\begin{axis}  
					[  
					ybar, 
					bar width=15pt,
					enlargelimits=0.15, 
					legend style={at={(0.5,-0.175)}, 	
						anchor=north,legend columns=-1},       
					ylabel={Civic Education Score}, 
					symbolic x coords={
						2019,
						2020,
						2021,
						2022,
						2023,
					},  
					xtick=data,  
					nodes near coords,  
					nodes near coords align={vertical},  
					width=\textwidth, 
					height=5cm, 
					width=16cm,
					]  
					\addplot [fill=yellow] coordinates {
						(2019,6)
						(2020,7)
						(2021,6.25)
						(2022,8.25)
						(2023,7.75)
						
					};
					\addplot [fill=red] coordinates {
						(2019,7.5)
						(2020,8.75)
						(2021,9.25)
						(2022,9)
						(2023,8.25)
						
					};
					\addplot [fill=green] coordinates {
						(2019,7.37)
						(2020,8.14)
						(2021,8.37)
						(2022,8.03)
						(2023,)
						
					};
					\addplot [fill=blue] coordinates {
						(2019,7.75)
						(2020,8.75)
						(2021,9.25)
						(2022,8.5)
						(2023,)	
					};
					\legend{ChatGPT, BingChat, AVS, MVS} 
				\end{axis}  
			\end{tikzpicture} 
		\end{center}
		\caption{Comparison in social combination.}
		\label{fig:social combination}
	\end{figure}
	
	\textbf{History}: While BingChat performs better, ChatGPT's results are comparable to those of Vietnamese students. With extensive and diverse training datasets, ChatGPT and BingChat are able to understand and process different types of historical questions and provide logical and useful responses. Although ChatGPT and BingChat may still encounter challenges with complex questions, they can be valuable resources for high school students in history.
	
	\textbf{Geography}: While BingChat achieves higher scores, ChatGPT performs at a similar level to Vietnamese students. The results indicate that both ChatGPT and BingChat are capable of understanding and responding to high school-level geography questions. They can effectively teach geography concepts and terminology, enhancing students' learning in high school geography. However, they may still face limitations when dealing with complex and in-depth inquiries that require advanced critical thinking.
	
	\textbf{Civic Education}: BingChat and ChatGPT showcase human-like abilities in the field of civic education. With their training in civic education and law-related subjects, they possess the expertise to provide high school-level knowledge in areas such as politics, law, citizen rights and responsibilities, and other social issues. Therefore, as reference tools, ChatGPT and BingChat can be highly valuable for Vietnamese students studying civic education. 
	
	\subsection{VNHSGE dataset and other datasets}
	
	In Figure~\ref{fig:chatgpt_exam}, the performance of ChatGPT and BingChat on the VNHSGE dataset is compared to other datasets in the GPT-4 Report~\cite{OpenAI_gpt_4_report}. The results show that ChatGPT's performance on the VNHSGE dataset is comparable to that of GPT-3.5 across subjects ranging from AP Statistics to AP Psychology. BingChat improves its performance in text-based subjects such as history, geography, civic education, and English. However, BingChat's performance does not significantly outperform ChatGPT in subjects like mathematics, physics, chemistry, and biology, which require complex computation and reasoning. On the other hand, GPT-4 exhibits better performance than GPT-3.5 in tasks of similar nature. This could be due to the structure of questions in these subjects from the VNHSGE dataset, which presents challenges for BingChat, particularly at the application and high application levels.
	
	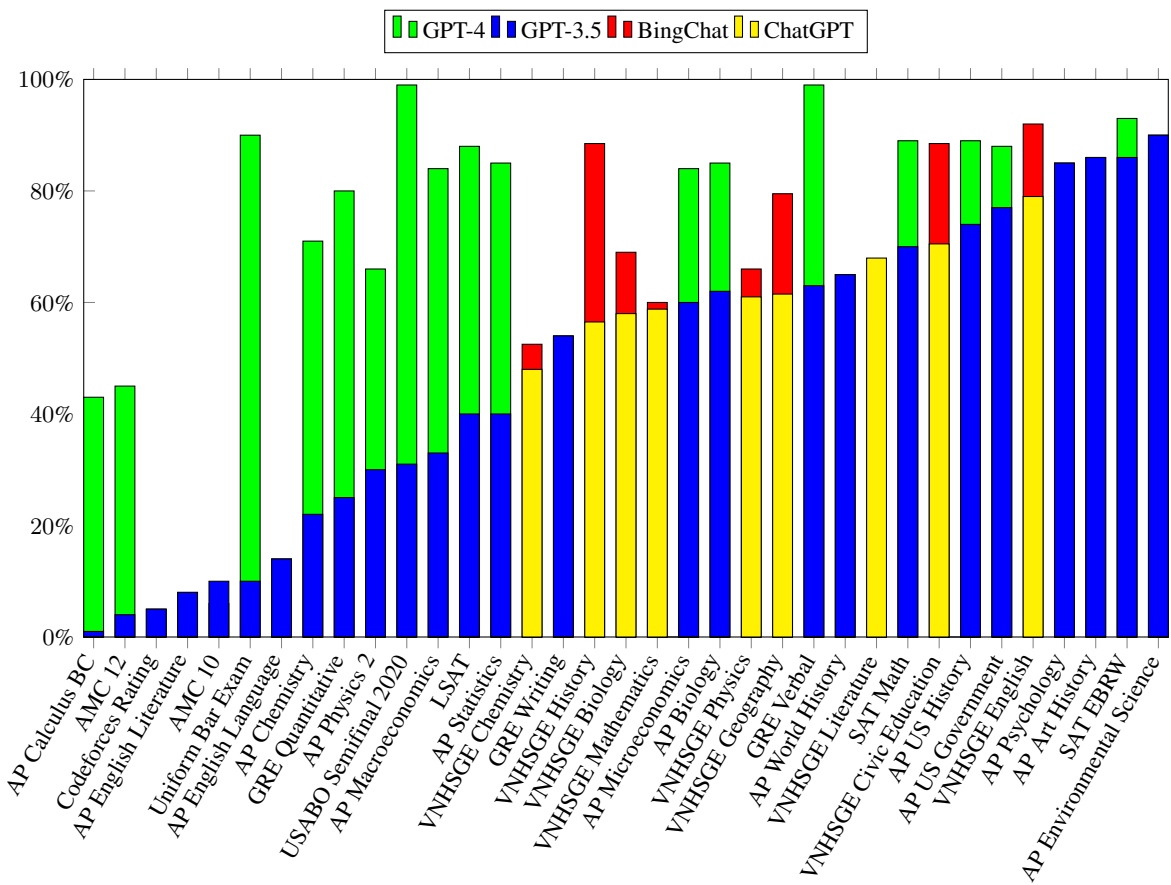
\begin{figure}[h!]
		\begin{center}
			\begin{tikzpicture}
				\begin{axis}[
					ylabel={\pgfmathprintnumber{\tick}\%},
					legend style={at={(0.5,1.125)}, 	
						anchor=north,legend columns=-1},
					symbolic x coords={
						AP Calculus BC,
						AMC 12,
						Codeforces Rating,
						AP English Literature,
						AMC 10,
						Uniform Bar Exam,
						AP English Language,
						AP Chemistry,
						GRE Quantitative,
						AP Physics 2,
						USABO Semifinal 2020,
						AP Macroeconomics,
						LSAT,
						AP Statistics,
						VNHSGE Chemistry,
						GRE Writing,
						VNHSGE History,
						VNHSGE Biology,
						VNHSGE Mathematics,
						AP Microeconomics,
						AP Biology,
						VNHSGE Physics,
						VNHSGE Geography,
						GRE Verbal,
						AP World History,
						VNHSGE Literature,
						SAT Math,
						VNHSGE Civic Education,
						AP US History,
						AP US Government,
						VNHSGE English,
						AP Psychology,
						AP Art History,
						SAT EBRW,
						AP Environmental Science,
					},
					xtick=data,
					hide axis,
					x tick label style={rotate=60,anchor=east},
					ybar,
					bar width=7.5pt,
					ymin=0,
					ymax=100,
					width=\textwidth, 
					enlarge x limits={abs=0.5*\pgfplotbarwidth},					
					height=9cm, width=16cm,
					]
					\addplot [fill=green] coordinates {
						(AP Calculus BC,0)	
					};
					\addplot [fill=blue] coordinates {
						(AP Microeconomics,0)
					};	
					\addplot [fill=red] coordinates {
						(VNHSGE Civic Education,0)	
					};	
					\addplot [fill=yellow] coordinates {
						(AP Environmental Science,0)	
					};		
					\legend{GPT-4, GPT-3.5, BingChat, ChatGPT}				
				\end{axis}
				\begin{axis}[
					ylabel={},
					symbolic x coords={
						AP Calculus BC,
						AMC 12,
						Codeforces Rating,
						AP English Literature,
						AMC 10,
						Uniform Bar Exam,
						AP English Language,
						AP Chemistry,
						GRE Quantitative,
						AP Physics 2,
						USABO Semifinal 2020,
						AP Macroeconomics,
						LSAT,
						AP Statistics,
						VNHSGE Chemistry,
						GRE Writing,
						VNHSGE History,
						VNHSGE Biology,
						VNHSGE Mathematics,
						AP Microeconomics,
						AP Biology,
						VNHSGE Physics,
						VNHSGE Geography,
						GRE Verbal,
						AP World History,
						VNHSGE Literature,
						SAT Math,
						VNHSGE Civic Education,
						AP US History,
						AP US Government,
						VNHSGE English,
						AP Psychology,
						AP Art History,
						SAT EBRW,
						AP Environmental Science,
					},
					xtick=data,
					x tick label style={rotate=60,anchor=east},
					yticklabel={\pgfmathprintnumber{\tick}\%},
					ybar,
					bar width=7.5pt,
					ymin=0,
					ymax=100,
					width=\textwidth, 
					enlarge x limits={abs=0.5*\pgfplotbarwidth},					
					height=9cm, width=16cm,
					]
					\addplot [fill=green] coordinates {
						(AP Calculus BC,43)
						(AMC 12,45)
						(Codeforces Rating,5)
						(AP English Literature,8)
						(AMC 10,6)
						(Uniform Bar Exam,90)
						(AP English Language,14)
						(AP Chemistry,71)
						(GRE Quantitative,80)
						(AP Physics 2,66)
						(USABO Semifinal 2020,99)
						(AP Macroeconomics,84)
						(LSAT,88)
						(AP Statistics,85)
						(VNHSGE Chemistry,0)
						(GRE Writing,54)
						(VNHSGE History,0)
						(VNHSGE Biology,0)
						(VNHSGE Mathematics,0)
						(AP Microeconomics,84)
						(AP Biology,85)
						(VNHSGE Physics,0)
						(VNHSGE Geography,0)
						(GRE Verbal,99)
						(AP World History,65)
						(VNHSGE Literature,0)
						(SAT Math,89)
						(VNHSGE Civic Education,0)
						(AP US History,89)
						(AP US Government,88)
						(VNHSGE English,0)
						(AP Psychology,85)
						(AP Art History,86)
						(SAT EBRW,93)
						(AP Environmental Science,90)	
					};					
				\end{axis}
				\begin{axis}[
					ylabel={Percentage},
					symbolic x coords={
						AP Calculus BC,
						AMC 12,
						Codeforces Rating,
						AP English Literature,
						AMC 10,
						Uniform Bar Exam,
						AP English Language,
						AP Chemistry,
						GRE Quantitative,
						AP Physics 2,
						USABO Semifinal 2020,
						AP Macroeconomics,
						LSAT,
						AP Statistics,
						VNHSGE Chemistry,
						GRE Writing,
						VNHSGE History,
						VNHSGE Biology,
						VNHSGE Mathematics,
						AP Microeconomics,
						AP Biology,
						VNHSGE Physics,
						VNHSGE Geography,
						GRE Verbal,
						AP World History,
						VNHSGE Literature,
						SAT Math,
						VNHSGE Civic Education,
						AP US History,
						AP US Government,
						VNHSGE English,
						AP Psychology,
						AP Art History,
						SAT EBRW,
						AP Environmental Science,
					},
					xtick=data,
					hide axis,
					x tick label style={rotate=60,anchor=east},
					ybar,
					bar width=7.5pt,
					ymin=0,
					ymax=100,
					width=\textwidth, 
					enlarge x limits={abs=0.5*\pgfplotbarwidth},					
					height=9cm, width=16cm,
					]
					\addplot [fill=blue] coordinates {
						(AP Calculus BC,1)
						(AMC 12,4)
						(Codeforces Rating,5)
						(AP English Literature,8)
						(AMC 10,10)
						(Uniform Bar Exam,10)
						(AP English Language,14)
						(AP Chemistry,22)
						(GRE Quantitative,25)
						(AP Physics 2,30)
						(USABO Semifinal 2020,31)
						(AP Macroeconomics,33)
						(LSAT,40)
						(AP Statistics,40)
						(VNHSGE Chemistry,0)
						(GRE Writing,54)
						(VNHSGE History,0)
						(VNHSGE Biology,0)
						(VNHSGE Mathematics,0)
						(AP Microeconomics,60)
						(AP Biology,62)
						(VNHSGE Physics,0)
						(VNHSGE Geography,0)
						(GRE Verbal,63)
						(AP World History,65)
						(VNHSGE Literature,0)
						(SAT Math,70)
						(VNHSGE Civic Education,0)
						(AP US History,74)
						(AP US Government,77)
						(VNHSGE English,0)
						(AP Psychology,85)
						(AP Art History,86)
						(SAT EBRW,86)
						(AP Environmental Science,90)	
					};					
				\end{axis}
				\begin{axis}[
					ylabel={Percentage},
					symbolic x coords={
						AP Calculus BC,
						AMC 12,
						Codeforces Rating,
						AP English Literature,
						AMC 10,
						Uniform Bar Exam,
						AP English Language,
						AP Chemistry,
						GRE Quantitative,
						AP Physics 2,
						USABO Semifinal 2020,
						AP Macroeconomics,
						LSAT,
						AP Statistics,
						VNHSGE Chemistry,
						GRE Writing,
						VNHSGE History,
						VNHSGE Biology,
						VNHSGE Mathematics,
						AP Microeconomics,
						AP Biology,
						VNHSGE Physics,
						VNHSGE Geography,
						GRE Verbal,
						AP World History,
						VNHSGE Literature,
						SAT Math,
						VNHSGE Civic Education,
						AP US History,
						AP US Government,
						VNHSGE English,
						AP Psychology,
						AP Art History,
						SAT EBRW,
						AP Environmental Science,
					},
					xtick=data,
					x tick label style={rotate=60,anchor=east},
					hide axis,
					ybar,
					bar width=7.5pt,
					ymin=0,
					ymax=100,
					width=\textwidth, 
					enlarge x limits={abs=0.5*\pgfplotbarwidth},					
					height=9cm, width=16cm,
					]	
					\addplot [fill=red] coordinates {
						(AP Calculus BC,0)
						(AMC 12,0)
						(Codeforces Rating,0)
						(AP English Literature,0)
						(AMC 10,0)
						(Uniform Bar Exam,0)
						(AP English Language,0)
						(AP Chemistry,0)
						(GRE Quantitative,0)
						(AP Physics 2,0)
						(USABO Semifinal 2020,0)
						(AP Macroeconomics,0)
						(LSAT,0)
						(AP Statistics,00)
						(VNHSGE Chemistry,52.5)
						(GRE Writing,0)
						(VNHSGE History,88.5)
						(VNHSGE Biology,69)
						(VNHSGE Mathematics,60)
						(AP Microeconomics,0)
						(AP Biology,0)
						(VNHSGE Physics,66)
						(VNHSGE Geography,79.5)
						(GRE Verbal,0)
						(AP World History,0)
						(VNHSGE Literature,56.8)
						(SAT Math,0)
						(VNHSGE Civic Education,88.5)
						(AP US History,0)
						(AP US Government,0)
						(VNHSGE English,92)
						(AP Psychology,0)
						(AP Art History,0)
						(SAT EBRW,0)
						(AP Environmental Science,0)	
					};
					
				\end{axis}
				\begin{axis}[
					ylabel={Percentage},
					symbolic x coords={
						AP Calculus BC,
						AMC 12,
						Codeforces Rating,
						AP English Literature,
						AMC 10,
						Uniform Bar Exam,
						AP English Language,
						AP Chemistry,
						GRE Quantitative,
						AP Physics 2,
						USABO Semifinal 2020,
						AP Macroeconomics,
						LSAT,
						AP Statistics,
						VNHSGE Chemistry,
						GRE Writing,
						VNHSGE History,
						VNHSGE Biology,
						VNHSGE Mathematics,
						AP Microeconomics,
						AP Biology,
						VNHSGE Physics,
						VNHSGE Geography,
						GRE Verbal,
						AP World History,
						VNHSGE Literature,
						SAT Math,
						VNHSGE Civic Education,
						AP US History,
						AP US Government,
						VNHSGE English,
						AP Psychology,
						AP Art History,
						SAT EBRW,
						AP Environmental Science,
					},
					x tick label style={rotate=60,anchor=east},
					hide axis,
					ybar,
					bar width=7.5pt,
					ymin=0,
					ymax=100,
					width=\textwidth, 
					enlarge x limits={abs=0.5*\pgfplotbarwidth},					
					height=9cm, width=16cm,
					]	
					\addplot [fill=yellow] coordinates {
						(AP Calculus BC,0)
						(AMC 12,0)
						(Codeforces Rating,0)
						(AP English Literature,0)
						(AMC 10,0)
						(Uniform Bar Exam,0)
						(AP English Language,0)
						(AP Chemistry,0)
						(GRE Quantitative,0)
						(AP Physics 2,0)
						(USABO Semifinal 2020,0)
						(AP Macroeconomics,0)
						(LSAT,0)
						(AP Statistics,00)
						(VNHSGE Chemistry,48)
						(GRE Writing,0)
						(VNHSGE History,56.5)
						(VNHSGE Biology,58)
						(VNHSGE Mathematics,58.8)
						(AP Microeconomics,0)
						(AP Biology,0)
						(VNHSGE Physics,61)
						(VNHSGE Geography,61.5)
						(GRE Verbal,0)
						(AP World History,0)
						(VNHSGE Literature,67.96)
						(SAT Math,0)
						(VNHSGE Civic Education,70.5)
						(AP US History,0)
						(AP US Government,0)
						(VNHSGE English,79)
						(AP Psychology,0)
						(AP Art History,0)
						(SAT EBRW,0)
						(AP Environmental Science,0)	
					};
				\end{axis}
			\end{tikzpicture}
		\end{center}
		\caption{Performance of ChatGPT, BingChat on VNHSGE dataset and GPT-3.5, GPT-4 on other datasets.}
		\label{fig:chatgpt_exam}
	\end{figure}

	\section{Conclusion}
	\label{sec:conclusion}

	In this paper, we present the VNHSGE dataset, which is intended to evaluate and train LLMs' multitask abilities such as question answering, text generation, reading comprehension, visual question answering, and more. The dataset covers nine subject areas from the Vietnamese National High School Graduation Examination, including social and language subjects such as literature, English, history, geography, and civic education, as well as calculation and inference subjects like mathematics, physics, chemistry, and biology. The dataset encompasses a wide range of question types, spanning from basic recall to complex calculation and reasoning questions. The VNHSGE dataset serves as a valuable resource for training LLMs, offering a diverse set of challenges at the human level. The dataset helps researchers identify critical flaws in models, thereby facilitating the improvement of LLMs' abilities. The VNHSGE dataset has various benefits for developing LLMs, including:

\begin{itemize}
	\item Comprehensive coverage: The dataset provides thorough coverage of a wide range of topics in nine high school subjects. This enables more thorough training of language models across diverse computing and inference domains.
	
	\item Various question types: The dataset contains a wide range of question types, from straightforward knowledge-based inquiries to intricate application-based inquiries requiring extensive investigation and evaluation. This offers a wide range of learning challenges for language models.
	
	\item Different difficulty levels: The VNHSGE dataset contains questions that range in complexity from simple to sophisticated, making it possible to train models that can handle a variety of question challenges.
	
	\item Vietnamese language: Given that the dataset is in Vietnamese, it is possible to train language models in a language other than English, enhancing their adaptability and global applicability.  
\end{itemize}

	The state of the art of LLMs, ChatGPT and BingChat, tested on the VNHGE dataset showed that the VNHSGE dataset is perfectly suited for LLMs. This outcome not only demonstrates the models' abilities but also presents chances and difficulties for LLMs deploying in the field of education.
	
	The VNHSGE dataset demonstrates the advantages and disadvantages of LLMs and offers information about possible instructional applications for these models. Additionally, it poses a challenge for LLMs to enhance their abilities to handle challenging, high-level application questions.
	
	\bibliographystyle{unsrt}  
	\bibliography{arxiv_final_1}  
			
	\appendix
	\newpage
	
	{\LARGE Appendix}
	
	\section{Available datasets}
	
	\label{paperwithcode}
	
	The authors searched Paperwithcode \footnote{\href{https://paperswithcode.com/datasets?mod=texts&task=question-answering}{https://paperswithcode .com/datasets?mod=texts\&task=question-answering}} as of April 25, 2023 for pre-existing datasets for a variety of topics, tasks, and languages in order to construct a new dataset. We searched for pertinent datasets on three levels: "General" datasets, datasets linked to "Texts", and datasets connected to "Text and Question Answering (QA)" using keywords like mathematics, literature, english, physics, chemistry, biology, history, geography, and law. Figure~\ref{fig:dataset} shows the number of datasets in subjects. The most datasets, including RACE~\cite{Lai2017}, MLQA~\cite{lewis2019mlqa}, SuperGLUE~\cite{wang2019superglue}, and DREAM~\cite{sun2019dream} were found in English, whereas Mathematics had Mathematics~\cite{saxton2019analysing}, MATH~\cite{hendrycks2021measuring} and GSM8K~\cite{cobbe2021training}. Numerous datasets were available for Physics, including TQA~\cite{kembhavi2017you},~SWAG~\cite{zellers2018swag}, PIQA~\cite{bisk2020piqa}, PROST~\cite{aroca2021prost}, and ScienceQA~\cite{lu2022learn}. The only two datasets for Literature and Law are \cite{shaham2022scrolls}, \cite{taktasheva2022tape} and  JEC-QA~\cite{zhong2020jec}, CaseHOLD~\cite{zheng2021does}, respectively, whereas the only one dataset available for Chemistry, Biology, History and Geography were ScienceQA~\cite{lu2022learn}.  We discovered that QA datasets had the highest number of datasets in the "Texts" category shown in Figure~\ref{fig:dataset in topic of text}. It is observed that only three datasets, MLQA~\cite{lewis2019mlqa}, XQuAD~\cite{artetxe2019cross}, and MKQA~\cite{longpre2021mkqa}, shown in Figure~\ref{fig:datasets in language}, supported Vietnamese.
	
	
	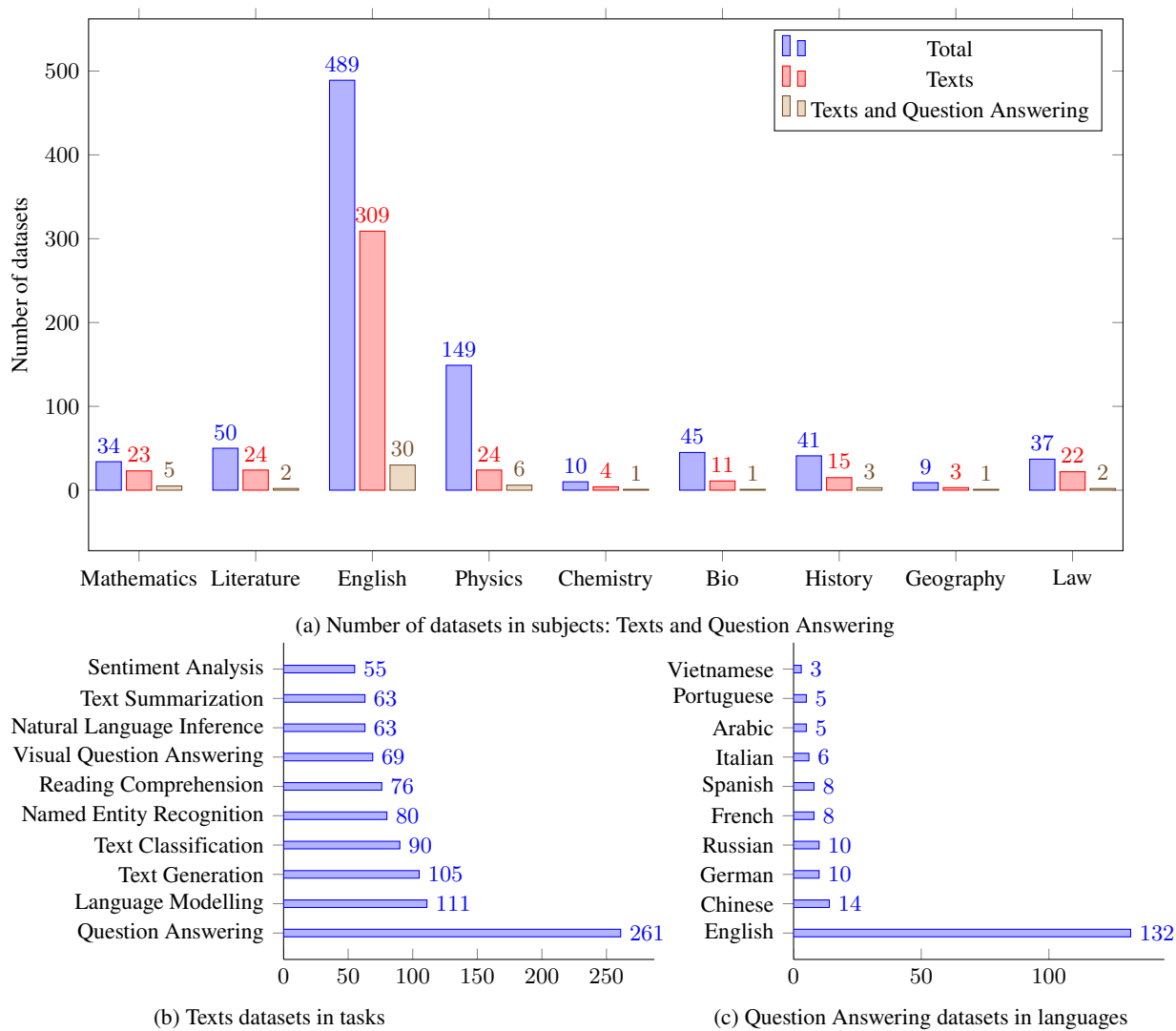
\begin{figure}[h!t]
		\centering
		\begin{subfigure}[b]{\textwidth}
			\begin{tikzpicture} 
				\begin{axis}  
					[  
					ybar, 
					enlargelimits=0.15, 
					ylabel={Number of datasets}, 
					symbolic x coords={
						Mathematics,
						Literature,
						English,
						Physics,
						Chemistry,
						Bio,
						History,
						Geography,
						Law,
					},  
					xtick=data,  
					nodes near coords,  
					nodes near coords align={vertical},  
					width=\textwidth, 
					enlarge x limits={abs=2*\pgfplotbarwidth},
					height=9cm, 
					width=16cm,
					]  
					\addplot coordinates {
						(Mathematics,34)
						(Literature,50)
						(English,489)
						(Physics,149)
						(Chemistry,10)
						(Bio,45)
						(History,41)
						(Geography,9)
						(Law,37)
					};   
					\addplot coordinates {
						(Mathematics,23)
						(Literature,24)
						(English,309)
						(Physics,24)
						(Chemistry,4)
						(Bio,11)
						(History,15)
						(Geography,3)
						(Law,22)
					};  
					\addplot coordinates {
						(Mathematics,5)
						(Literature,2)
						(English,30)
						(Physics,6)
						(Chemistry,1)
						(Bio,1)
						(History,3)
						(Geography,1)
						(Law,2)
						
					};
					
					\legend{Total, Texts, Texts and Question Answering}  
					
				\end{axis}  
			\end{tikzpicture} 
			\caption{Number of datasets in subjects: Texts and Question Answering}
			\label{fig:dataset}
		\end{subfigure}		
		\begin{subfigure}[b]{0.45\textwidth}
			\centering
			\begin{tikzpicture}
				\begin{axis}[
					symbolic y coords={
						Question Answering,
						Language Modelling,
						Text Generation,
						Text Classification,
						Named Entity Recognition,
						Reading Comprehension,
						Visual Question Answering,
						Natural Language Inference,
						Text Summarization,
						Sentiment Analysis,
					},
					ytick=data,
					xbar,
					bar width=3pt,
					xmin=0,
					enlarge y limits,
					height=6cm, 
					width=6.75cm,
					nodes near coords, nodes near coords align={horizontal},
					axis x line*=bottom, axis y line*=left
					]
					
					\addplot coordinates {
						(261,Question Answering)
						(111,Language Modelling)
						(105,Text Generation)
						(90,Text Classification)
						(80,Named Entity Recognition)
						(76,Reading Comprehension)
						(69,Visual Question Answering)
						(63,Natural Language Inference)
						(63,Text Summarization)
						(55,Sentiment Analysis)
					};
				\end{axis}
			\end{tikzpicture}
			\caption{Texts datasets in tasks}
			\label{fig:dataset in topic of text}
		\end{subfigure}
		\hfill
		\begin{subfigure}[b]{0.45\textwidth}
			\centering
			\begin{tikzpicture}
				\begin{axis}[
					symbolic y coords={
						English,
						Chinese,
						German,
						Russian,
						French,
						Spanish,
						Italian,
						Arabic,
						Portuguese,
						Vietnamese,
					},
					ytick=data,
					xbar,
					bar width=3pt,
					xmin=0,
					enlarge y limits,
					height=6cm, 
					width=6.75cm,
					nodes near coords, nodes near coords align={horizontal},
					axis x line*=bottom, axis y line*=left
					]
					
					\addplot coordinates {
						(132,English)
						(14,Chinese)
						(10,German)
						(10,Russian)
						(8,French)
						(8,Spanish)
						(6,Italian)
						(5,Arabic)
						(5,Portuguese)
						(3,Vietnamese)
					};
				\end{axis}
			\end{tikzpicture}
			\caption{Question Answering datasets in languages}
			\label{fig:datasets in language}
		\end{subfigure}
		\caption{Available datasets on Paperwithcode.}
		\label{fig:dataset topic language}
	\end{figure}
	
	\section{Dataset format}
	\label{convert_data_format}
	
	In this section, we describe how to convert formulas, equations, tables, photos, and charts from raw text formats like Word, Pdf, and HTML into a text-only format and an image folder. The exact steps of the method are shown in detail in Figure~\ref{fig:raw_json} including steps: (1) collecting raw data in Word format file, (2) translating symbols, formulas, and equations into Latex format, (3) converting Word format to JSON format.  
	
	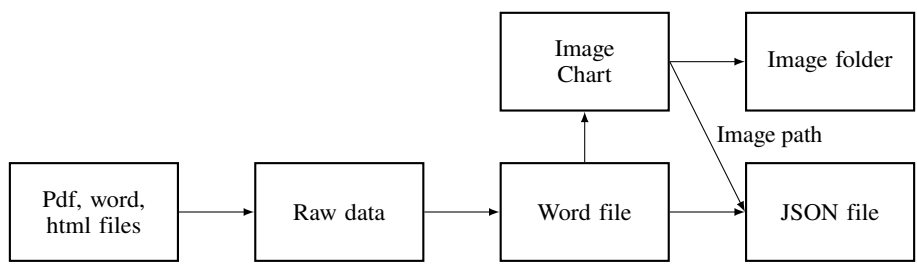
\begin{figure}[ht!]
		\begin{center}
			\begin{tikzpicture}  
				\node[block] (a) {Pdf, word, html files};  
				\node[block,right=of a] (b) {Raw data};   
				\node[block,right=of b] (c) {Word file};  
				\node[block,right=of c] (d) {JSON file}; 
				\node[block] (e) at ([yshift=2cm]$(d)!0.5!(d)$) {Image folder};   
				\node[block] (g) at ([yshift=2cm]$(c)!1.0!(c)$) {Image\\Chart};   
				\draw[line] (a)-- (b);  
				\draw[line] (b)-- (c);  
				\draw[line] (c)-- (d);  
				\draw[line] (g)-- (e);  
				\draw[line] (c)-- (g) ;   
				\draw[line] (g.east) -- (d.west) node [midway, right] (TextNode) {Image path};
			\end{tikzpicture} 
		\end{center}
		\caption{Convert raw data to json files and images.}
		\label{fig:raw_json}
	\end{figure}
	
	\begin{itemize}
		\item Step 1: Take "Raw data" in. Questions and answers are the basic data that we present. The answers are multiple-choice with in-depth explanations. Microsoft Word displays the raw data as a table. A row with six columns represents each question's counterpart. The subsequent processing of the results is made easier with the aid of this data structure.
		
		\begingroup
		\renewcommand{\arraystretch}{1.25} 
		\begin{tabular}{|l|l|l|l|l|l|}
			\hline
			ID & Image Question & Question   & Choice & Image Answer & Explanation           \\ \hline
		\end{tabular}
		\endgroup
		
		\item Step 2: Symbols, formulas, and equations are converted to text format using the LaTeX format in the "Raw data" to "Word format" conversion. In mathematics, physics, chemistry, and biology, we convert symbols, formulas, and equations using three different techniques. The first technique converts Word documents with equations and formulae to the Latex format using the built-in equation editor in Microsoft Word. If the first approach is unable to convert the raw data, the second option employs the Mathpix\footnote{https://mathpix.com/} software to convert pdf files to the Latex format. Sometimes it's not possible to utilize any of the two ways mentioned earlier, in which case we must manually input the formulas and equations.
		
		\item Step 3: Convert "Word format" into "JSON format". With the aid of Python libraries, including "docx\footnote{\href{https://pypi.org/project/python-docx/}{https://pypi.org/project/python-docx/}}" and "JSON," it is simple to convert Word files to JSON files. The procedure entails importing the necessary libraries before using their functions to parse and convert the text data to JSON format. The "docx" library offers tools for reading and writing Microsoft Word documents. Data conversion to the JSON format is made easy and effective by the "JSON" library.

	\end{itemize}

	\subsection{Raw data}
	\label{sec:Raw data}
	
	There are several phases involved in transforming raw data into a machine-readable format. Finding and removing pertinent information from the raw data is one of the crucial tasks. The raw data may be in several formats, including HTML, PDF, or Word, and may include information on a variety of disciplines, including math, literature, english, physics, chemistry, biology, history, geography, and civic education. Each of these subjects has distinct qualities that call for various extraction techniques. For instance, symbols, formulas, and equations in mathematics, physics, chemistry, and biology must be precisely retrieved and represented. These equations might be as simple as simple biological equations or as sophisticated as complex mathematical equations. On the other hand, geography frequently contains a large number of images and charts that must be accurately retrieved and provided. Symbols, formulas, and equations are often not used in literature, english, history, or civic education; instead, these subjects place a greater emphasis on textual content. In conclusion, a variety of approaches and methodologies were used to transform raw data into a machine-readable format, ensuring that all pertinent information is extracted and accurately represented. A similar strategy must be used to accurately extract and represent each subject within the raw data. This procedure is essential for ensuring that the result is precise, trustworthy, and simple to understand.
	
	Our raw data is displayed in \ref{prob:Raw_data} as an example. For ease of viewing, we don't present the example in table format. This is a query from the math dataset. As we can see, the questions and answers both include illustrations, equations, and formulas. The answers include thorough justifications that call for high-level inference skills, while the questions demand the ability to extract information from images. The information is complicated and may require specialist knowledge or training to properly understand, as suggested by the use of images and technical language.
	
	\begin{ProblemSpecBox}{"Raw data sample"}
		\label{prob:Raw_data}	
		\begin{multicols}{2}
			\textbf{Question}: 
			Let $y=f(x)$ be a cubic function with the graph shown in the picture. \\
			\begin{tikzpicture}[scale=0.75]
				\begin{axis}[
					axis lines=middle,
					xlabel={$x$},
					ylabel={$y$},
					xmin=-2.5, xmax=5,
					ymin=-2, ymax=4.5,
					xtick={-2,2},
					ytick={-1,2},
					legend style={at={(1,1)},anchor=north east},
					]
					\addplot[domain=-2.5:4.5, samples=200, ultra thick, blue] {0.0208*x^5-0.1417*x^4 + 0.3375*x^3 -0.1833*x^2 -1.6833*x+ 3.0000};
				\end{axis}
			\end{tikzpicture}
			
			The number of real solutions of the equation $|f(x^{3}-3 x)|=\frac{2}{3}$ is: \\
			\begin{oneparchoices}
				\choice 6
				\choice 10
				\choice 3
				\choice 9
			\end{oneparchoices} \\
			\textbf{Solution}: From the graph of the function $y=f(x)$, we deduce that the graph of the function $y=|f(x)|$ is: \\
			\begin{tikzpicture}[scale=0.75]
				\begin{axis}[
					axis lines=middle,
					xlabel={$x$},
					ylabel={$y$},
					xmin=-2.5, xmax=5,
					ymin=-2, ymax=4.5,
					xtick={-2,2},
					ytick={-1,1,2},
					legend style={at={(1,1)},anchor=north east},
					]
					\addplot[domain=-3:-2, samples=200, ultra thick, blue] {-(0.0208*x^5-0.1417*x^4 + 0.3375*x^3 -0.1833*x^2 -1.6833*x+ 3.0000)};
					\addplot[domain=-2:2, samples=200, ultra thick, blue] {0.0208*x^5-0.1417*x^4 + 0.3375*x^3 -0.1833*x^2 -1.6833*x+ 3.0000};
					\addplot[domain=2:4, samples=200, ultra thick, blue] {-(0.0208*x^5-0.1417*x^4 + 0.3375*x^3 -0.1833*x^2 -1.6833*x+ 3.0000)};
					\addplot[domain=4:4.5, samples=200, ultra thick, blue] {0.0208*x^5-0.1417*x^4 + 0.3375*x^3 -0.1833*x^2 -1.6833*x+ 3.0000};
				\end{axis}
			\end{tikzpicture}
			\\
			Setting $t=x^{3}-3x$, we have $\left|f\left(x^{3}-3 x\right)\right|=\frac{2}{3} \Leftrightarrow|f(t)|=\frac{2}{3}$.
			From the above graph, we conclude that the equation $|f(t)|=\frac{2}{3}$ has six distinct solutions $t=t_{i}$ (with $i=\overline{1,6}$ and $\left.t_{1}<-2 ;-2<t_{2}, t_{3}<2 ; t_{4}, t_{5}, t_{6}>2\right)$.
			Considering the function $t(x)=x^{3}-3 x$, we have $t^{\prime}(x)=3 x^{2}-3 ; t^{\prime}(x)=0 \Leftrightarrow x= \pm 1$.
			The sign variation table of $t(x)$ is:\\
			\begin{tikzpicture}[scale=0.65]
				\newcommand*{ \E}{ \ensuremath{ \mathrm{e}}}.
				\tkzTabInit[lgt=3,espcl=2.5]
				{$x$ /1,$f'(x)$ /1,$f(x)$ /2}
				{$-\infty$,$-1$,$1$,$+\infty$}
				\tkzTabLine{,+,z,-,z,+}
				\tkzTabVar{-/$-\infty$,
					+/$2$,
					-/$-2$,
					+/$+\infty$}
				\tkzTabIma{2}{4}{3}{$0$}            
			\end{tikzpicture}
			\\
			Based on the table of variations, we have: 
			\begin{itemize}
				\item The equation $x^{3}-3x=t_{1}$ has one solution (since $\left.t_{1}<-2\right)$.
				\item Each equation $x^{3}-3x=t_{2}, x^{3}-3x=t_{3}$ has three distinct solutions (since $-2<t_{2}, t_{3}<2$). 
				\item Each equation $x^{3}-3x=t_{4}, x^{3}-3x=t_{5}, x^{3}-3x=t_{6}$ has one solution (since $\left.t_{4}, t_{5}, t_{6}>2\right)$. 
			\end{itemize}
			The equation $\left|f\left(x^{3}-3 x\right)\right|=\frac{2}{3}$ has 10 solutions. Therefore, the answer is \textcolor{red}{\textbf{B}}. 10. 		
		\end{multicols}

	\end{ProblemSpecBox}
	
	\newpage	
	\section{Samples} 
	\label{Dataset_response}
	
	\subsection{Mathematics} 
	\label{VNHSGE_Math}
	
	\subsubsection{Knowledge level question}
	\label{VNHSGE_Math_K}
	
	Simple math problems can be answered correctly by both ChatGPT and BingChat, and their rationales are adequate. This is due to the fact that easy arithmetic problems typically have a particular format and structure, which enables the models to employ the established patterns and guidelines to get accurate results.  
	
	\begin{mdframed}[linewidth=1pt,linecolor=red] 
		\textbf{Question}: Given the geometric sequence $(u_n)$ with $u_1=2$ and common ratio $q=\frac{1}{2}$. What is the value of $u_3$?
		
		\begin{oneparchoices}
			\choice 3
			\choice $\frac{1}{2}$
			\choice $\frac{1}{4}$
			\choice $\frac{7}{2}$
		\end{oneparchoices}
		
		\textbf{Solution}: We have $u_3 = u_1 \cdot q^2 = 2 \cdot (\frac{1}{2})^2 = 2 \cdot \frac{1}{4} = \frac{1}{2}$. Therefore, the answer is \textcolor{red}{\textbf{B}}~$\frac{1}{2}$.
		
		\textbf{ChatGPT:} We have a geometric sequence $(u_n)$ with common ratio $q=\frac{1}{2}$ and $u_1=2$. Therefore
		$$u_2 = u_1\cdot q = 2\cdot\frac{1}{2} = 1,~
		u_3 = u_2\cdot q = 1\cdot\frac{1}{2} = \frac{1}{2}$$
		Thus, the value of $u_3$ is $\frac{1}{2}$, and the correct answer is \textcolor{red}{\textbf{B}}. $\frac{1}{2}$.
		
		\textbf{BingChat:} In a geometric sequence, the nth term is calculated using the formula $u_n = u_1 * q^{n-1}$. Therefore, the value of $u_3$ in this geometric sequence is $u_3 = u_1 * q^{3-1} = 2 * (1/2)^2 = 1/2$. The correct answer is \textcolor{red}{\textbf{B}}. 1/2.
		
	\end{mdframed}
	
	\subsubsection{Comprehension level question} 
	\label{VNHSGE_Math_C}
	
	Due to limitations in their training data and their computational abilities, LLMs may have trouble providing accurate answers as arithmetic question difficulty rises. In certain circumstances, ChatGPT might be unable to offer the right solution, whereas BingChat is able to give the correct answer and provide a rational explanation.

	\begin{mdframed}[linewidth=1pt,linecolor=red]
		\textbf{Question}: A box contains 15 balls including 6 red balls numbered from 1 to 6 and 9 blue balls numbered from 1 to 9. Two balls are randomly drawn from the box, what is the probability that they have different colors and their numbers add up to an even number?
		
		\begin{oneparchoices}
			\choice $\frac{9}{35}$
			\choice $\frac{18}{35}$
			\choice $\frac{4}{35}$
			\choice $\frac{1}{7}$
		\end{oneparchoices}
		
		\textbf{Solution}:
		The number of ways to randomly select 2 balls from the box is: $C_{15}^{2}=105$ ways.
		To have the sum of numbers on the two balls even, we have two cases:
		\begin{itemize}
			\item Case 1: Two balls of different colors with odd numbers: $C_{3}^{1} \cdot C_{5}^{1}=15$ ways.
			\item Case 2: Two balls of different colors with even numbers: $C_{3}^{1} \cdot C_{4}^{1}=12$ ways.
		\end{itemize}
		The probability is: $P=\frac{12+15}{105}=\frac{9}{35}$.
		Therefore, the answer is \textcolor{red}{\textbf{A}}. $\frac{9}{35}$.
		
		\textbf{ChatGPT}: To choose two balls of different colors and have the sum of their numbers even, we must have the following cases:
		\begin{itemize}
			\item Choose one red ball and one green ball. This case occurs when we select one ball from the group of 6 red balls and one ball from the group of 9 green balls. The sum of the numbers on these two balls is necessarily odd, so we do not need to consider this case.
			\item Choose two green balls. This case occurs when we select two balls from the group of 9 green balls. The number of ways to choose two green balls is $\binom{9}{2}=36$. There are 5 cases where the sum of the numbers on the two green balls is even:
			\item Choose two green balls numbered 1: sum is 2 (even).
			\item Choose two green balls numbered 2: sum is 4 (even).
			\item ...
			\item Choose two green balls numbered 9: sum is 18 (even). Thus, there are 5 satisfying cases. The probability we seek is:
		\end{itemize}
		$$\frac{\text{number of satisfying cases}}{\text{number of ways to choose two balls from the box
		}} = \frac{5}{\binom{15}{2}} = \frac{5}{105} = \frac{1}{21}$$ 
		Therefore, the correct answer is \textcolor{red}{\textbf{D}}. $\frac{1}{7}$
		
		\textbf{BingChat}: The correct answer is \textcolor{red}{\textbf{A}}. 9/35. The number of ways to choose two balls of different colors and the sum of the two numbers on them is even equals the number of ways to choose a red ball with an odd number on it and a blue ball with an odd number on it plus the number of ways to choose a red ball with an even number on it and a blue ball with an even number on it. Therefore, the number of ways to choose two balls of different colors and the sum of the two numbers on them is even equals 3 * 5 + 3 * 4 = 27. The number of elements in the sample space equals C(15,2) = 105. Therefore, the probability of getting two balls of different colors and the sum of the two numbers on them is even equals 27/105 = 9/35.
		
	\end{mdframed}
	
	\subsubsection{Application level question}
	\label{VNHSGE_Math_A}
	
	Both ChatGPT and BingChat are unable to resolve complex issues that need an in-depth analysis of derivatives and graphs. This is due to the fact that their training data and computational language abilities are constrained, and it's possible that they lack the mathematical knowledge required to address these issues.

	\begin{mdframed}[linewidth=1pt,linecolor=red]
		\textbf{Question}: How many integer values of the parameter $m$ make the function $y = -x^4 + 6x^2 + mx$ have three extreme points? 
		
		\begin{oneparchoices}
			\choice 17
			\choice 15
			\choice 3
			\choice 7
		\end{oneparchoices}
		
		\textbf{Solution}: We have: $y'=-4x^3+12x+m$. Consider the equation $y'=0 \Leftrightarrow -4x^3+12x+m=0~(1)$.
		For the function $y$ to have three extreme points, equation (1) must have 3 distinct roots. We have: $(1) \Leftrightarrow m=4x^3-12x$.
		Consider the function $g(x)=4x^3-12x$ with $g'(x)=12x^2-12$. We have $g'(x)=0 \Leftrightarrow 12x^2-12=0 \Leftrightarrow x=\pm 1$. The tables of signs and variations of $g(x)$:
		\begin{center}
			\begin{tikzpicture}
				\newcommand*{ \E}{ \ensuremath{ \mathrm{e}}}.
				\tkzTabInit{$x$ /1,$y'(x)$ /1,$y(x)$ /2}
				{$-\infty$,$-1$,$1$,$+\infty$}
				\tkzTabLine{,+,z,-,z,+}
				\tkzTabVar{-/$-\infty$,
					+/$8$,
					-/$-8$,
					+/$+\infty$}
				\tkzTabIma{2}{4}{3}{$0$}            
			\end{tikzpicture}
		\end{center}
		Based on the tables of signs and variations, equation (1) has 3 distinct roots when $-8<m<8$.
		Since $m \in \mathbb{Z} \Rightarrow m \in{-7,-6,-5, \ldots, 5,6,7}$.
		Thus, there are 15 integer values of the parameter $m$ that satisfy the requirement in the problem. Answer choice \textcolor{red}{\textbf{B}}.~$15$ is correct.
		
		\textbf{ChatGPT}: To have the function $y=-x^4+6x^2+mx$ have three extreme points, the graph of the function must have two local maximum points and one local minimum point. The local maximum points occur at $x=\pm\sqrt{2}$ and the local minimum point occurs at $x=0$. The derivative of the function is: $ y' = -4x^3 + 12x + m $. To find the integer values of the parameter $m$ such that the function has two local maximum points at $x=\pm\sqrt{2}$, we need to satisfy the conditions: 
		$ y'(\sqrt{2})=y'(-\sqrt{2})=0$ 
		Which is equivalent to:
		$ \Leftrightarrow 
		\left\{\begin{matrix}
			-8\sqrt{2}+12+\sqrt{2}m=0  \\
			8\sqrt{2}+12-\sqrt{2}m=0 \\
		\end{matrix}\right.
		$ 
		Solving this system of equations yields
		$ \Leftrightarrow 
		\left\{\begin{matrix}
			m = 5\sqrt{2}  \\
			m = 19\sqrt{2} \\
		\end{matrix}\right.
		$
		To have a local minimum point at $x=0$, we need $y'(0)=0$, which implies $m=0$. Therefore, the integer values of $m$ that satisfy all of the conditions are $m=5\sqrt{2}$, $m=19\sqrt{2}$, and $m=0$. Thus, the answer is \textcolor{red}{\textbf{D}}.~7.
		
		\textbf{BingChat}: The correct answer is \textcolor{red}{\textbf{D}}. 7. Here’s how to solve it:
		
		The derivative of the function $y = -x^4 + 6x^2 + mx$ is $y’ = -4x^3 + 12x + m$. To find the critical points of the function, we need to solve the equation y’ = 0.
		
		This equation can be rewritten as $-4x^3 + 12x + m = 0$ or $x(4x^2 - 12) = -m$. Since $x$ is a factor of $-m$, $x$ must be an integer divisor of $-m$.
		
		For the function to have three critical points, the quadratic equation $4x^2 - 12 = -m/x$ must have two distinct real roots for some value of $x$. The discriminant of this quadratic equation is $(-12)^2 - 4 * 4 * (-m/x) = 144 + 16m/x$.
		
		For the quadratic equation to have two distinct real roots, the discriminant must be positive. So we need to find the integer values of $m$ for which $144 + 16m/x >$ 0 for some integer value of x.
		
		Since $x$ is an integer divisor of $-m$, we can write $m = -kx$ for some integer $k$. Substituting this into the inequality above, we get $144 + 16m/x > 0$ becomes $144 - 16k >$ 0 or $k < 9$.
		
		So m can be any integer multiple of $x$ such that the absolute value of the multiple is less than 9. There are 7 such values: $m = \pm x, \pm 2x, \pm 3x, \pm 4x$.
		
	\end{mdframed}
	
	\subsubsection{High application level question}
	\label{VNHSGE_Math_H}
	
	Due to the problem's extremely challenging level and the need for complex analytical skills to solve it, ChatGPT is unable to handle so. The solution offered by ChatGPT is inaccurate and unreliable. Additionally, BingChat is unable to resolve this issue. Although it attempts to suggest a seemingly reliable solution strategy, which is wholly wrong, it lacks a solution approach.

	\begin{mdframed}[linewidth=1pt,linecolor=red]
		\textbf{Question}: Consider complex numbers $z$ that satisfy $|z^2 - 3 - 4i| = 2|z|$. Let $M$ and $m$ respectively be the largest and smallest values of $|z|$. The value of  $M^2+m^2$ is
		
		\begin{oneparchoices}
			\choice 28
			\choice $18+4\sqrt{6}$
			\choice 14
			\choice $11+4\sqrt{6}$
		\end{oneparchoices}
		
		\textbf{Solution}:
		Applying the triangle inequality, we have: 
		$2|z|=\left|z^{2}-3-4 i\right| \geq|| z^{2}|-| 3+4 i|=||z|^{2}-5$
		(since $\left.\left|z^{2}\right|=|z|^{2}\right)$. The equality occurs when $z^{2}=k(-3-4 i)$. Hence, we get $4|z|^{2} \geq(|z|-5)^{2} \Leftrightarrow|z|^{4}-14|z|^{2}+25 \leq 0 \Leftrightarrow 7-2 \sqrt{6} \leq|z|^{2} \leq 7+2 \sqrt{6}$. Therefore, we have $\sqrt{6}-1 \leq|z| \leq \sqrt{6}+1$. Thus, we get $M=1+\sqrt{6}$ and $m=\sqrt{6}-1$. Therefore, $M^{2}+m^{2}=14$. Therefore, the answer is \textcolor{red}{\textbf{C}}.~14.
		
		\textbf{ChatGPT}: We have the equation $|z^2 - 3 - 4i| = 2|z|$ which is equivalent to the system:
		$\left\{\begin{matrix}
			|z^2 - 3 - 4i|^2 = 4|z|^2 \\
			|z|^2 = x^2 + y^2 \geq 0
		\end{matrix}\right. $
		
		Thus:
		$|z^2 - 3 - 4i|^2 = |(x^2-y^2-3) + 2xyi - 4i|^2 = (x^2-y^2-3)^2 + 4x^2y^2 - 8x^2 + 16y^2 = 4(x^2+y^2)|z|^2$
		
		Which is equivalent to:
		$(x^2-y^2-3)^2 + 4x^2y^2 - 8x^2 + 12y^2 = 0$. Setting $t = x^2 - y^2$, $u = xy$ we have $t^2 + 4u^2 - 8t + 12y^2 = 0$. We have $4t^2 - 16y^2 = (2t-4y)(2t+4y) \leq 0$ so $|t| \leq 2|y|$. Thus:
		$0 \leq (x^2+y^2) = |z|^2 = \frac{1}{2}(|z|^2 + |z|^2) \leq \frac{1}{2}(M^2+m^2)$. And $(x^2 + y^2)^2 = t^2 + 4u^2 \leq 4|y|^2 (|t|+4|y|) \leq 8|y|^3$. Therefore: $0 \leq |z|^2 \leq 2\sqrt{2}|y|^{3/2}$. In general, we have $M^2+m^2 \leq 8\sqrt{2}$.

		\textbf{BingChat}: This is a complex math question. Would you like me to explain how to solve it?	To solve this question, we can rewrite the expression $|z^2 - 3 - 4i| = 2 |z|$ as $|z^2 - 3 - 4i| = |2z|$. Then, we apply the theorem on the distance between two points in the complex plane to infer that $z^2$ is the midpoint of the two complex points $z$ and $(3 + 4i)$. From there, we can find the value of $z$ and calculate the value of $M^2 + m^2$. The correct answer to this question is \textcolor{red}{\textbf{D}}. $11+4\sqrt{6}$.
		
	\end{mdframed}

	\subsection{Literature}
	\label{VNHSGE_Lit}
	
	\subsubsection{Question and Answer}
	\label{VNHSGE_Lit_Q}
	
	Both ChatGPT and BingChat are capable of correctly determining the poetry genre from an excerpt and offering pertinent explanations. The characteristics that distinguish poetry apart from various kinds of writing, including meter, rhyme, and imagery, can be recognized by ChatGPT and BingChat. Employing this knowledge, they can next categorize the sample into the proper poetry genre and explain their categorization.

	\begin{mdframed}[linewidth=1pt,linecolor=red] 
		I. Reading Comprehension (3.0 points)
		
		Đọc đoạn trích: 
		\begin{multicols}{2}
			
			chính mẹ đẻ anh hùng và truyền thuyết 
			
			tù túp lều lơp lá lơp tranh 
			
			cắt cuống nhau bằng lưỡi liềm 
			
			bàn chân thô quanh năm bùn lấm 
			
			chưa một lần ướm qua sử sách
			tập con bước vịn vào ca dao tục ngữ 
			
			dù uống nước đâu lòng vẫn nhớ nguồn 
			
			thương từ cái kiến con ong 
			
			tím ruột bầm gan thù bọn ác 
			dân tộc tôi khi đứng dậy làm người 
			
			là đứng theo dáng mẹ 
			
			"đòn gánh tre chín dạn hai vai"
			
			mùa hạ gió Lào quăng quật
			
			mùa đông sắt se gió bấc 
			
			dân tộc tôi khi đứng dậy làm người 
			
			mồ hôi vã một trời sao trên đất 
			
			trời sao lặn hóa thành muôn mạch nước 
			
			chảy âm thầm chảy dọc thời gian	 
			
		\end{multicols}
		Thực hiện các yêu cầu sau: 
		
		\textbf{Câu 1}: Xác định thể thơ của đoạn trích 
		
		\textbf{Đáp án}: Thể thơ tự do 
		
		\textbf{ChatGPT}: Đoạn trích không phải là một bài thơ đầy đủ, tuy nhiên, nó được viết theo thể thơ tự do (free verse), không có quy luật về độ dài câu thơ, số lượng câu thơ, hoặc âm vần. 
		
		\textbf{BingChat}: Đoạn trích trên được viết theo thể thơ tự do. Thể thơ tự do không tuân theo những quy tắc về số lượng chữ trong một câu thơ hay số lượng câu thơ trong một khổ. Thay vào đó, nó tập trung vào cách sử dụng ngôn ngữ và hình ảnh để truyền tải cảm xúc và ý nghĩa của tác giả.
		
	\end{mdframed}
	
	Although ChatGPT and BingChat are capable of understanding the inquiries, the intended language might not always be possible. LLMs have reduced abilities in both word positioning and language comprehension. This drawback results from the fact that LLMs rely on statistical correlations and patterns revealed in training data for their operation. They may choose the wrong words or use the wrong wording because they may not fully comprehend the underlying meanings and implications of words and phrases. LLMs find it difficult to precisely capture the intended meaning of a given text since language is complex and can have various meanings and interpretations.

	\begin{mdframed}[linewidth=1pt,linecolor=red] 	
		\textbf{Câu 2}: Chỉ ra những từ diễn tả đời sống nghèo khó vất vả của người mẹ trong đoạn thơ: 
		
		chính mẹ đẻ anh hùng và truyền thuyết 
		
		từ túp lều lợp lá lợp tranh 
		
		cắt cuống nhau bằng lưỡi liềm 
		
		bàn chân thô quanh năm bùn lấm 
		
		chưa một lần ướm qua sử sách

		\textbf{Đáp án}: Từ ngữ diễn tả đời sống nghèo khó của mẹ: Túp lều, lợp lá, lợp tranh, lưỡi liềm, bàn chân thô, bùn lấm. 
		
		\textbf{ChatGPT}: Những từ ngữ diễn tả đời sống nghèo khó, vất vả của người mẹ trong đoạn thơ bao gồm:

		tù túp (tựa như đang chen chúc)
		
		lều lơp lá lơp tranh (lều nhỏ, vách dựng bằng lá)
		
		cắt cuống nhau bằng lưỡi liềm (công việc khó khăn, vất vả)
		
		bàn chân thô quanh năm bùn lấm (cuộc sống vất vả)
		
		chưa một lần ướm qua sử sách (không có cơ hội học hỏi).

		\textbf{BingChat}: 
		
		Trong đoạn thơ trên, có một số từ ngữ diễn tả đời sống nghèo khó và vất vả của người mẹ. Đó là:

		“tù túp lều lơp lá lơp tranh”: diễn tả ngôi nhà nghèo khó, đơn sơ của người mẹ.
		
		“cắt cuống nhau bằng lưỡi liềm”: diễn tả công việc vất vả của người mẹ trong việc cắt cỏ để nuôi gia súc.
		
		“bàn chân thô quanh năm bùn lấm”: diễn tả sự vất vả và khó khăn trong cuộc sống của người mẹ khi phải làm việc chân trần trong bùn đất.
		
		“chưa một lần ướm qua sử sách”: diễn tả sự thiếu học vấn và kiến thức của người mẹ do hoàn cảnh khó khăn.
		
		Những từ ngữ này giúp người đọc hình dung được cuộc sống nghèo khó và vất vả của người mẹ trong bài thơ.
		
	\end{mdframed}	
	
	The meanings of the aforementioned poems can be presented on ChatGPT and BingChat, respectively. In this setting, LLMs' reading comprehension skills are quite strong. ChatGPT and BingChat can employ their language comprehension skills to analyze the meanings of poetry by determining the tone, mood, and overall message of the poem. Additionally, they are able to identify literary techniques like metaphors, similes, and symbolism, which can serve to further explain the meaning of the poetry.

	\begin{mdframed}[linewidth=1pt,linecolor=red] 	
		\textbf{Câu 3}: Nêu nội dung của hai dòng thơ:

		tập con bước vịn vào ca dao tục ngữ 
		
		dù uống nước đâu lòng vẫn nhớ nguồn

		\textbf{Đáp án}: Nội dung câu thơ như một lời nhắc nhở thế hệ sau phải biết giữ gìn truyền thống của dân tộc. Biết ghi nhớ và biết ơn giá trị của dân tộc mình.
		
		\textbf{ChatGPT}: Hai dòng thơ này nói về việc tập con người học tập và lấy nét từ các ca dao, tục ngữ, đồng thời nhắc nhở người ta rằng dù ở đâu, họ vẫn không nên quên nguồn gốc của mình. Nó thể hiện tình cảm gắn bó với nơi sinh ra, nơi mình trưởng thành, nơi mà các giá trị truyền thống và văn hóa dân tộc được tạo hình và truyền lại từ thế hệ này sang thế hệ khác
		
		\textbf{BingChat}: Hai dòng thơ “tập con bước vịn vào ca dao tục ngữ / dù uống nước đâu lòng vẫn nhớ nguồn” nói về việc người mẹ dạy con bằng những câu ca dao tục ngữ và giáo dục con nhớ về nguồn cội, gốc rễ của mình. Ca dao tục ngữ là những câu nói, bài hát dân gian truyền miệng từ đời này sang đời khác, phản ánh đời sống, tư tưởng và tâm hồn của người dân. Người mẹ dùng những câu ca dao tục ngữ để dạy con những bài học quý giá về cuộc sống và giáo dục con biết ơn đời, biết nhớ về nguồn cội của mình. Đây là một thông điệp quan trọng trong đoạn thơ này.
		
	\end{mdframed}
	
	The excerpt's content is commentable on both ChatGPT and BingChat. Both ChatGPT and BingChat have good Vietnamese language comprehension. Based on their understanding of the language and the text's context, both models can analyze the material and offer useful commentary. They are able to offer insightful comments on the material by recognizing and deciphering the themes, imagery, and symbols included in the passage.

	\begin{mdframed}[linewidth=1pt,linecolor=red] 		
		\textbf{Câu 4}: Nhận xét về hình ảnh dân tộc Việt Nam được thể hiện trong đoạn trích 
		
		\textbf{Đáp án}: Nhận xét về hình ảnh dân tộc Việt Nam:
		
		- Con người Việt Nam phải sống trong muôn vàn khó khăn vất vả nhưng họ vẫn ngời sáng phẩm chất tinh thần tốt đẹp.
		
		- Lòng biết ơn, sự kiên cường.
		
		- Lòng yêu nước căm thù giặc sâu sắc
		
		\textbf{ChatGPT}: Đoạn trích thể hiện hình ảnh một dân tộc Việt Nam chịu đựng và vượt qua khó khăn, gắn bó mạnh mẽ với quê hương và truyền thống văn hóa. Những từ ngữ miêu tả cuộc sống nghèo khó, vất vả, như "tù túp", "lều lơp lá lơp tranh", "cắt cuống nhau bằng lưỡi liềm", "bàn chân thô quanh năm bùn lấm" đã thể hiện một cuộc sống khắc nghiệt và cực kỳ khó khăn của người dân tộc Việt Nam.
		
		Tuy nhiên, bài thơ cũng thể hiện tình yêu thương sâu sắc của người dân tộc Việt Nam dành cho đất nước và truyền thống văn hóa của mình. Những câu thơ "dù uống nước đâu lòng vẫn nhớ nguồn" hay "tập con bước vịn vào ca dao tục ngữ" đã đưa ra một thông điệp rõ ràng về sự quan tâm và tôn trọng giá trị của truyền thống dân tộc.
		
		\textbf{BingChat}: Trong đoạn trích trên, hình ảnh dân tộc Việt Nam được thể hiện một cách sinh động và đầy ý nghĩa. Dân tộc Việt Nam được miêu tả là một dân tộc kiên cường, bền bỉ và không ngừng vươn lên trong cuộc sống. Điều này được thể hiện qua hình ảnh người mẹ vất vả nuôi con bằng những công việc nặng nhọc và dạy con bằng những câu ca dao tục ngữ quý giá. Dân tộc Việt Nam cũng được miêu tả là một dân tộc biết ơn đời, biết nhớ về nguồn cội và gốc rễ của mình. Điều này được thể hiện qua câu thơ “dù uống nước đâu lòng vẫn nhớ nguồn”. Ngoài ra, hình ảnh dân tộc Việt Nam còn được thể hiện qua sự đoàn kết và sức mạnh khi đứng dậy làm người, vượt qua mọi khó khăn và thử thách. Đây là một hình ảnh đẹp và đầy tự hào của dân tộc Việt Nam.
	\end{mdframed}
	
	BingChat and ChatGPT may both write a 200-word review on a certain subject. LLMs are capable of writing meaningful paragraph structures, with ChatGPT outperforming BingChat in this regard. This is due to the fact that BingChat and ChatGPT are both language models created to produce human-like language. They are able to comprehend the meaning and context of the assigned issue and provide a well-organized response.

	\begin{mdframed}[linewidth=1pt,linecolor=red] 
		II. Writing (7.0 points)
		
		\textbf{Câu 1}: Từ nội dung đoạn trích ở phần đọc hiểu, hãy viết một đoạn văn khoảng 200 chữ trình bày suy nghĩ của anh chị về sức mạnh của tinh thần vượt khó trong cuộc sống.
		
		\textbf{Đáp án}:
		
		Yêu cầu chung:
		
		- Xác định đúng vấn đề nghị luận: sức mạnh của tinh thần vượt khó trong cuộc sống
		
		- Viết đúng hình thức đoạn văn, số lượng chữ, không mắc lỗi dùng từ, đặt câu
		
		Yêu cầu cụ thể: bài viết cần đảm bảo các ý sau:
		
		- Giới thiệu vấn đề cần nghị luận: sức mạnh của tinh thần vượt khó trong cuộc sống
		
		- Giải thích: tinh thần vượt khó có thể hiểu là sẵn sàng đối mặt với khó khăn thử thách trong cuộc sống, không chùn bước khi gặp thất bại. Những người có tinh thần vượt khó ắt sẽ vươn đến thành công
		
		- Bàn luận: người có tinh thần vượt khó sẽ không dễ dàng buông xuôi khi gặp khó khăn. Họ sẽ tìm cách giải quyết vấn đề, biến nguy thành cơ, biến khó khăn thành động lực
		
		- Ý nghĩa của tinh thần vượt khó: trong cuộc sống người có tinh thần vượt khó sẽ đạt được nhiều thành tựu trong cuộc sống, người có tinh thần vượt khó sẽ không dễ dàng bỏ qua các cơ hội. Người có tinh thần vượt khó cũng sẽ là tấm gương sáng truyền cảm hứng đến những người xung quanh.
		
		- Chứng minh: Học sinh này dẫn chứng phù hợp với các tấm gương trong cuộc sống
		
		- Phản đề: Bên cạnh những người không ngừng nỗ lực cố gắng khi gặp khó khăn thì vẫn còn nhiều người lười biếng khi gặp khó khăn thì chùn bước bỏ cuộc. Những người như vậy sẽ bỏ qua nhiều cơ hội tốt đẹp và sẽ thất bại
		
		- Liên hệ và tổng kết vấn đề
		
		\textbf{ChatGPT}:
		Cuộc sống đầy khó khăn và thử thách, tuy nhiên, không phải ai cũng có đủ sức mạnh và can đảm để vượt qua chúng. Sức mạnh của tinh thần vượt khó là điều cần thiết để đối mặt và vượt qua những khó khăn trong cuộc sống.
		
		Trong đoạn trích, bài thơ đã mô tả cuộc sống vất vả của người dân tộc Việt Nam, từ tù túp, lều lơp lá lơp tranh, đến bàn chân thô quanh năm bùn lấm. Tuy nhiên, những điều đó không hạnh phúc, nhưng đó là chân lý mà mọi người phải đối mặt. Có thể thấy, tinh thần vượt khó của dân tộc Việt Nam là điều đã giúp họ vượt qua những khó khăn đó.
		
		Trong cuộc sống, chúng ta sẽ gặp phải rất nhiều khó khăn và thử thách. Đôi khi chúng ta sẽ cảm thấy mệt mỏi, bất lực và không biết phải làm gì để vượt qua. Tuy nhiên, sức mạnh của tinh thần vượt khó là điều giúp chúng ta có thể đứng lên và tiếp tục đi tiếp trên con đường mà mình đã chọn.
		
		Tinh thần vượt khó giúp chúng ta tạo ra những kế hoạch, đề ra mục tiêu, tập trung vào những điều tích cực và đối mặt với những khó khăn trong cuộc sống. Nếu không có sự mạnh mẽ của tinh thần vượt khó, chúng ta có thể không thể đạt được những thành công lớn trong cuộc sống.
		
		Tuy nhiên, để có được sức mạnh của tinh thần vượt khó, chúng ta cần phải có niềm tin vào bản thân, tìm kiếm nguồn động lực bên trong và xây dựng một tâm trạng tích cực. Chúng ta cũng cần nhận ra rằng, sức mạnh của tinh thần vượt khó không phải là điều mà ta có thể có được từ ngày qua ngày, mà đó là một quá trình dài và cần phải được bồi đắp từng ngày.
		
		Với sức mạnh tinh thần và sự nỗ lực, chúng ta có thể vượt qua mọi khó khăn trong cuộc sống và đạt được những thành công lớn. Hãy học hỏi từ những hình mẫu sống động như mẹ của nhân vật trong đoạn trích để có thể trở thành một người mạnh mẽ và thành đạt.
		
		\textbf{BingChat}: 
		Sức mạnh của tinh thần vượt khó là một trong những yếu tố quan trọng giúp con người vượt qua những khó khăn và thử thách trong cuộc sống. Đoạn trích trên đã thể hiện rõ sức mạnh này qua hình ảnh người mẹ vất vả nuôi con trong hoàn cảnh nghèo khó và dạy con bằng những câu ca dao tục ngữ quý giá. Dù cuộc sống có nhiều khó khăn và thử thách, người mẹ vẫn kiên cường và không ngừng vươn lên để nuôi con khôn lớn và dạy con những bài học quý giá về cuộc sống.
		
		Sức mạnh của tinh thần vượt khó giúp con người có động lực để tiếp tục phấn đấu và không bao giờ từ bỏ ước mơ và hoài bão của mình. Nó giúp chúng ta có thể đối mặt với những khó khăn và thử thách một cách lạc quan và tích cực, tìm ra những giải pháp để vượt qua chúng. Sức mạnh này còn giúp chúng ta có thể học hỏi được nhiều điều từ những trải nghiệm khó khăn và trưởng thành hơn trong cuộc sống.
		Trong cuộc sống, chúng ta sẽ gặp phải nhiều khó khăn và thử thách. Tuy nhiên, nếu chúng ta có sức mạnh của tinh thần vượt khó, chúng ta sẽ có thể đối mặt với chúng một cách dũng cảm và vượt qua chúng để tiếp tục bước đi trên con đường của mình. Đó là sức mạnh tuyệt vời của tinh thần vượt khó trong cuộc sống.
		
	\end{mdframed}
	
	The hidden meaning, utilized to examine the main idea of literary works, cannot yet be understood by LLMs, despite their ability to comprehend and present the explicit meaning. Due to their inability to comprehend the specificity of literature—the use of the particular to allude to the general—LLMs still have significant limitations. It is observed that ChatGPT outperforms BingChat in an essay that needs the capacity to analyze and evaluate.

	\begin{mdframed}[linewidth=1pt,linecolor=red]	
		\textbf{Câu 2}: Trong Việt Bắc nhà thơ Tố Hữu viết:
		\begin{multicols}{2}
			\emph{
				Ta về, mình có nhớ ta \\
				Ta về, ta nhớ những hoa cùng người. \\
				Rừng xanh hoa chuối đỏ tươi \\
				Đèo cao nắng ánh dao gài thắt lưng. \\
				Ngày xuân mơ nở trắng rừng \\
				Nhớ người đan nón chuốt từng sợi giang. \\
				Ve kêu rừng phách đổ vàng \\
				Nhớ cô em gái hái măng một mình. \\
				Rừng thu trăng rọi hòa bình \\
				Nhớ ai tiếng hát ân tình thủy chung. \\
			}
		\end{multicols}
		Anh chị hãy phân tích đoạn trích trên, từ đó nhận xét về lẽ sống ân tình được thể hiện trong đoạn trích.
		*Yêu cầu hình thức:
		
		- Thí sinh viết kết hợp kiến thức và kỹ năng làm nghị luận văn học để tạo lập văn bản bài viết
		
		- Phải có bố cục đầy đủ rõ ràng, văn viết có cảm xúc, diễn đạt trôi chảy, đảm bảo tính liên kết, không mắc lỗi chính tả, từ ngữ, ngữ pháp
		
		*Về yêu cầu nội dung:
		
		- Giới thiệu chung: Tố Hữu là lá cờ đầu của nền văn nghệ cách mạng Việt Nam hiện đại. Nội dung thơ Tố Hữu hướng tới những sự kiện cách mạng của dân tộc trong thế kỷ XX. Bài thơ Việt Bắc ra đời sau kháng chiến chống Pháp thắng Lợi năm 1954. Được coi là thi phẩm xuất sắc tiêu biểu trong phong cách nghệ thuật thơ Tố Hữu. Đoạn thơ là lời người đi để khẳng định lòng thủy chung với Việt Bắc.
		
		- Khái quát vấn đề phân tích đoạn thơ: từ đó nhận xét về lẽ sống ân nghĩa được thể hiện trong đoạn trích phân tích
		
		*Phân tích đoạn thơ bức tranh tứ bình
		
		+ Mùa đông: Rừng xanh hoa chuối đỏ tươi/ Đèo cao nắng ánh dao gài thắt lưng:
		
		Cảnh: với sắc xanh ngút ngàn của núi rừng điểm những bông hoa chuối đỏ tươi như những bó đuốc sáng rực, xua đi sự lạnh lẽo hiu hắt của núi rừng. Thắp lên ngọn lửa ấm áp, mang lại ánh sáng nơi hơi ấm cho nơi đây.
		Con người: trước thiên nhiên bao la của núi rừng trở nên kỳ bí, hùng tráng hơn với hoạt động phát nương làm rẫy.
		+ Mùa xuân: Ngày xuân mơ nở trắng rừng/ Nhớ người ta nón chuốt từng sợi giang
		
		Cảnh: hoa mơ rừng nở trắng khiến bừng sáng cả khu rừng, làm dịu mát tâm hồn con người
		Con người: đan nón, chuốt từng sợi giang. Một vẻ đẹp tình nghĩa được biểu hiện qua bàn tay khéo léo, tài hoa, nhanh nhẹn, chăm chút cẩn thận đối với từng sản phẩm lao động
		+ Mùa hè: Ve kêu rừng phách đổ vàng/ Nhớ cô em gái hái măng một mình
		
		Cảnh: rừng khách đổ vào màu vàng rực của thiên nhiên dường như chuyển đột ngột qua "tự đổ" . Với tiếng ve kêu khiến cảnh sinh động có hồn và tưng bừng hơn.
		Con người: hình ảnh cô gái hái măng một mình đã lột tả được vẻ đẹp của sự dịu thương chịu khó của con người nơi đây
		+ Mùa thu: Rừng thu trăng gọi hòa bình/ Nhớ ai tiếng hát ân tình thủy chung
		
		Cảnh: ánh trăng dịu nhẹ, huyền ảo gợi không khí thanh bình yên ả.
		Con người: hiện lên với tiếng hát ân tình thủy chung, với bao tình cảm ân tình sâu sắc với cách mạng
		- Nhận xét về lẽ sống ân nghĩa được thể hiện trong đoạn trích: lẽ sống ân nghĩa được thể hiện thông qua nỗi nhớ, sự lưu luyến, bị lụy trong giây phút chia tay giữa đồng bào miền núi với các cán bộ khi trở về Hà Nội. Lẽ sống ân nghĩa còn được thể hiện qua sự mường tượng của tác giả về những kỉ niệm, sự gắn bó giúp đỡ của những đồng bào trong suốt quá trình sống và chiến đấu của các chiến sĩ ở khu vực miền Bắc. Lẽ sống ân nghĩa chính là một nét đẹp truyền thống của nhân dân ta, nó tạo nên sự liên kết giữa con người với con người. Góp phần xây dựng nên sức mạnh đại đoàn kết toàn dân
		
		Kết luận: Khái quát lại vấn đề, giá trị nội dung nghệ thuật
		
		\textbf{ChatGPT}:
		
		Đoạn thơ của Tố Hữu đưa người đọc vào một không gian của quê hương, nơi mà những hình ảnh thiên nhiên và con người được kết nối một cách tự nhiên và tình cảm. Tác giả thể hiện sự nhớ những hoa cùng người khi trở về quê hương, một tình cảm thiêng liêng, đậm đà lẫn trong sự vui mừng và tràn đầy kỷ niệm.
		
		Bên cạnh đó, tác giả cũng miêu tả rất sống động những hình ảnh thiên nhiên và con người của quê hương Việt Nam. Những rừng xanh hoa chuối đỏ tươi, đèo cao nắng ánh dao gài thắt lưng, ngày xuân mơ nở trắng rừng, ve kêu rừng phách đổ vàng, rừng thu trăng rọi hòa bình. Tất cả tạo nên một bức tranh quê hương đẹp đẽ, hoang sơ và vẫn giữ được sự trong sạch, tự nhiên.
		
		Điểm nổi bật trong đoạn thơ này là thông điệp về lẽ sống ân nghĩa. Tác giả nhắc nhở về sự quan tâm, chia sẻ và tình người trong cuộc sống. Ý tưởng về sự ân nghĩa xuất hiện khi tác giả nhớ lại những người thân yêu, những người đã bên cạnh anh trong những khoảnh khắc khó khăn và cùng nhau chia sẻ niềm vui, nỗi buồn. Bức tranh quê hương đẹp đẽ, với những hình ảnh thiên nhiên và con người, như một thông điệp về sự sống động và sự ấm áp, sự đoàn kết giữa mọi người.
		
		Như vậy, đoạn thơ của Tố Hữu không chỉ thể hiện tình cảm thiêng liêng đối với quê hương, mà còn khẳng định giá trị của lẽ sống ân nghĩa. Nói đến quê hương, không thể thiếu những hình ảnh thiên nhiên và con người, cũng như tình cảm giữa con người. Sự đoàn kết, sự chia sẻ và tình người trong cuộc sống là những giá trị cốt lõi của lẽ sống của con người, nhất là trong nền văn hoá truy
		
		Ngoài ra, trong đoạn thơ này, Tố Hữu còn thể hiện sự tình cảm và sự gắn bó mạnh mẽ với quê hương và những người dân nơi đó. Những hình ảnh rừng xanh, hoa chuối đỏ tươi, đèo cao nắng ánh dao gài, ngày xuân trắng rừng và ve kêu rừng phách đổ vàng đều gợi lên cho người đọc một cảm giác yên bình, trong lành, gần gũi với thiên nhiên và con người của nơi đó. Các hình ảnh này cũng thể hiện sự bền chặt, vững vàng của những nét văn hóa truyền thống của quê hương Tố Hữu.
		
		Từ đó, ta có thể thấy rõ lẽ sống ân nghĩa, tình cảm, sự gắn bó mạnh mẽ với quê hương, gia đình và đồng bào được thể hiện trong đoạn trích trên. Tố Hữu đã truyền tải thông điệp về tình yêu quê hương, tình yêu đồng bào và những giá trị truyền thống của quê hương mình. Điều này thể hiện rõ sự cần thiết và ý nghĩa của việc giữ gìn và phát huy những giá trị văn hóa, truyền thống của dân tộc, đồng thời xây dựng một xã hội đoàn kết, tình nghĩa trong cuộc sống.
		
		\textbf{BingChat}: Đoạn trích trên là một đoạn thơ của nhà thơ Tố Hữu, được viết trong bối cảnh ông đang ở Việt Bắc. Trong đoạn thơ này, Tố Hữu đã miêu tả một cảnh sắc thiên nhiên tươi đẹp và sinh động với những hình ảnh như “rừng xanh hoa chuối đỏ tươi”, “đèo cao nắng ánh dao gài thắt lưng” hay “ve kêu rừng phách đổ vàng”. Những hình ảnh này giúp người đọc hình dung được vẻ đẹp của thiên nhiên Việt Bắc.
		
		Tuy nhiên, điểm nhấn của đoạn thơ này là sự nhớ nhung và ân nghĩa mà Tố Hữu dành cho những người mà ông đã gặp gỡ và quen biết trong cuộc sống. Mỗi câu thơ đều bắt đầu bằng từ “nhớ”, thể hiện sự nhớ nhung và tình cảm mà Tố Hữu dành cho những người mà ông đã gặp gỡ. Ông nhớ về người đan nón chuốt từng sợi giang, cô em gái hái măng một mình hay ai đó có tiếng hát ân tình thủy chung. Những hình ảnh này thể hiện sự ân nghĩa và lòng biết ơn của Tố Hữu đối với những người đã giúp đỡ và chia sẻ với ông trong cuộc sống.
		
		Nhìn chung, đoạn trích trên thể hiện lẽ sống ân nghĩa của con người. Đó là sự biết ơn và trân trọng những người đã giúp đỡ và chia sẻ với mình trong cuộc sống. Đó cũng là thông điệp mà Tố Hữu muốn gửi gắm qua bài thơ này.

	\end{mdframed}
	
	\subsubsection{Scoring}
	\label{VNHSGE_Lit_S}
	
	\textit{\textbf{Scoring}}: Every year, the Vietnam Ministry of Education and Training releases the results of the high school and university entrance tests. The answers are often published on the Ministry of Education and Training of Vietnam's website and widely disseminated via official media channels.

	\begingroup
	\renewcommand{\arraystretch}{1.25} 
	
	\begin{tabular}{|c|c|l|c|}
		\hline
		\textbf{Part}              & \textbf{Question}           & \multicolumn{1}{c|}{\textbf{Score criteria}}                                                                                                                                              & \textbf{Score}            \\ \hline
		\multirow{4}{*}{\textbf{I}}   & \textbf{1}                  & Thể thơ: tự do                                                                                                                                                                     & \textbf{0.75}             \\ \cline{2-4} 
		& \textbf{2}                  & Từ ngữ diễn tả đời sống nghèo khó của mẹ: Túp lều, lợp lá.. ...                                                                                                                   & \textbf{0.75}             \\ \cline{2-4} 
		& \textbf{3}                  & Nội dung câu thơ như một lời nhắc nhở thế hệ sau phải biết giữ gìn truyền thống ...                                                                                                                                                & \textbf{1.0}              \\ \cline{2-4} 
		& \textbf{4}                  & Nhận xét về hình ảnh dân tộc Việt Nam được thể hiện trong đoạn trích: ..                                                                                                         & \textbf{0.5}              \\ \hline
		\multirow{12}{*}{\textbf{II}} & \multirow{6}{*}{\textbf{1}} & Tổng điểm câu 1 & \textbf{2}                \\ \cline{3-4} 
		&                             & a. Đảm bảo yêu cầu về hình thức đoạn văn                                                                                                                                           & 0.25 \\ \cline{3-4} 
		&                             & b. Xác định đúng vấn đề nghị luận                                                                                                                                                  & 0.25 \\ \cline{3-4} 
		&                             & c. Triển khai vấn đề nghị luận                                                                                                                                                     & 0.25 \\ \cline{3-4} 
		&                             & d. Chính tả, từ ngữ, ngữ pháp                                                                                                                                                      & 0.25 \\ \cline{3-4} 
		&                             & e. Sáng tạo                                                                                                                                                                        & 0.25 \\ \cline{2-4} 
		& \multirow{6}{*}{\textbf{2}} & Tổng điểm câu 2                                                                                                                                                                      & \textbf{5.0}              \\ \cline{3-4} 
		&                             & a. Đảm bảo cấu trúc của văn nghị luận                                                                                                                                              & 0.25                      \\ \cline{3-4} 
		&                             & b. Xác định đúng vấn đề nghị luận                                                                                                                                                  & 0.5                       \\ \cline{3-4} 
		&                             & c. Triển khai vấn đề nghị luận                                                                                                                                                     & 3.5                       \\ \cline{3-4} 
		&                             & d. Chính tả, từ ngữ, ngữ pháp                                                                                                                                                      & 0.25                      \\ \cline{3-4} 
		&                             & e. Sáng tạo                                                                                                                                                                        & 0.25                      \\ \hline
	\end{tabular}
	
	\endgroup
	
	It's possible that BingChat's search engine functionality—which is inappropriate for producing a critical essay—is the reason ChatGPT performs better than it does in terms of literature. While ChatGPT is a huge language model capable of producing coherent and contextually suitable responses to natural language input, BingChat is a search engine mechanism meant to deliver answers to inquiries based on keyword matching and retrieval of pertinent resources. Therefore, ChatGPT would be better suited to creating an essay expressing perspectives and evaluations on a topic like literature, whereas BingChat's capabilities would be more constrained.

	\begingroup
	\renewcommand{\arraystretch}{1.25} 
	
	\begin{table}[ht!]
		\resizebox{\textwidth}{!}{%
			\begin{tabular}{c|cccc|cccccccccc|c|}
				\cline{2-16}
				\multicolumn{1}{l|}{}                   & \multicolumn{4}{c|}{\textbf{I}}                                                          & \multicolumn{10}{c|}{\textbf{II}}                                                                                                                                                                                                                                & \multicolumn{1}{l|}{\multirow{2}{*}{\textbf{Total}}} \\ \cline{2-15}
				& \multicolumn{1}{c|}{1}    & \multicolumn{1}{c|}{2}    & \multicolumn{1}{c|}{3}    & 4    & \multicolumn{1}{c|}{1.a}  & \multicolumn{1}{c|}{1.b}  & \multicolumn{1}{c|}{1.c}  & \multicolumn{1}{c|}{1.d}  & \multicolumn{1}{c|}{1.e}  & \multicolumn{1}{c|}{2.a}  & \multicolumn{1}{c|}{2.b}  & \multicolumn{1}{c|}{2.c}  & \multicolumn{1}{c|}{2.d}  & 2.e  & \multicolumn{1}{l|}{}                                \\ \hline
				\multicolumn{1}{|c|}{\textbf{Score}}    & \multicolumn{1}{c|}{0.75} & \multicolumn{1}{c|}{0.75} & \multicolumn{1}{c|}{1.00} & 0.50 & \multicolumn{1}{c|}{0.25} & \multicolumn{1}{c|}{0.25} & \multicolumn{1}{c|}{1.00} & \multicolumn{1}{c|}{0.25} & \multicolumn{1}{c|}{0.25} & \multicolumn{1}{c|}{0.25} & \multicolumn{1}{c|}{0.50} & \multicolumn{1}{c|}{3.50} & \multicolumn{1}{c|}{0.25} & 0.50 & \textbf{10.00}                                       \\ \hline
				\multicolumn{1}{|c|}{\textbf{ChatGPT}}  & \multicolumn{1}{c|}{0.75} & \multicolumn{1}{c|}{0.37} & \multicolumn{1}{c|}{0.7}  & 0.4  & \multicolumn{1}{c|}{0.23} & \multicolumn{1}{c|}{0.23} & \multicolumn{1}{c|}{0.58} & \multicolumn{1}{c|}{0.2}  & \multicolumn{1}{c|}{0.07} & \multicolumn{1}{c|}{0.2}  & \multicolumn{1}{c|}{0.5}  & \multicolumn{1}{c|}{1.83} & \multicolumn{1}{c|}{0.2}  & 0.2  & \textbf{6.48}                                        \\ \hline
				\multicolumn{1}{|c|}{\textbf{BingChat}} & \multicolumn{1}{c|}{0.75} & \multicolumn{1}{c|}{0}    & \multicolumn{1}{c|}{0.3}  & 0.3  & \multicolumn{1}{c|}{0.25} & \multicolumn{1}{c|}{0.25} & \multicolumn{1}{c|}{1}    & \multicolumn{1}{c|}{0.25} & \multicolumn{1}{c|}{0}    & \multicolumn{1}{c|}{0.13} & \multicolumn{1}{c|}{0.25} & \multicolumn{1}{c|}{1.25} & \multicolumn{1}{c|}{0.25} & 0    & \textbf{4.975}                                       \\ \hline
			\end{tabular}
		}
	\end{table}
	
	\endgroup
	
	\subsection{English}
	\label{VNHSGE_Eng}

	Both ChatGPT and BingChat are unable to respond to inquiries requiring the identification of the word whose underlined portion is pronounced differently from the other three. The amount of phonetic knowledge and analysis required for this work is higher than that LLMs currently possess or are equipped to accomplish. Both ChatGPT and BingChat perform well on the remaining portions of the exam, proving that LLMs have received adequate training in English vocabulary, grammar, and reading comprehension.

	\subsubsection{Pronunciation and stress question}
	\label{VNHSGE_Eng_PS}
	
	\begin{mdframed}[linewidth=1pt,linecolor=red] 
		\textbf{Question}: Mark the letter A, B, C, or D to indicate the word whose underlined part differs from the other three in pronunciation in each of the following questions.
		\begin{oneparchoices}
			\choice m\underline{i}ss
			\choice ch\underline{i}ld
			\choice s\underline{i}gn		
			\choice m\underline{i}nd
		\end{oneparchoices}
		
		\textbf{Solution}: 
		Pronunciation: Pronounce single and double consonant sounds. Miss is pronounced as /i/ and the rest is pronounced as /ai/. Choose \textcolor{red}{\textbf{A}}. 
		
		\textbf{ChatGPT}:\textcolor{red}{\textbf{B}}. ch\underline{i}ld (pronounced with a "ch" sound)
		
		\textbf{BingChat}: Here are the correct answers to your questions: \textcolor{red}{\textbf{C}}. sign. Is there anything else you would like to know?
		
	\end{mdframed}
	
	\subsubsection{Grammar and vocabulary questions}
	\label{VNHSGE_Eng_G}
	\begin{mdframed}[linewidth=1pt,linecolor=red] 
		\textbf{Question}: Mark the letter A, B, C, or D on your answer sheet to indicate the correct answer to each of the following questions. 
		
		Question: Life in the city is ...... than life in the countryside.
		
		\begin{oneparchoices}
			\choice exciting
			\choice most exciting
			\choice the most exciting
			\choice more exciting
		\end{oneparchoices}
		
		\textbf{Solution}: Comparative structure: Comparative more + adj (with three syllables) + than 
		Eliminate option A because it is not a comparative structure. Eliminate options B and C because they use the superlative structure. Choose \textcolor{red}{\textbf{D}}. 
		
		\textbf{ChatGPT}: \textcolor{red}{\textbf{D}}. more exciting 
		
		\textbf{BingChat}: \textcolor{red}{\textbf{D}}. more exciting
		
	\end{mdframed}
	
	\begin{mdframed}[linewidth=1pt,linecolor=red] 
		\textbf{Question}: Mark the letter A, B, C, or D on your answer sheet to indicate the underlined part that needs correction in each of the following questions.
		
		Question: My mother \underline{takes} vitamins \underline{regularly} as she believes that \underline{it} benefit her \underline{mind} and skin.
		
		\begin{oneparchoices}
			\choice takes
			\choice regularly
			\choice it
			\choice mind
		\end{oneparchoices}
		
		\textbf{Solution}: Pronoun: Referencing language. The word 'it' is referring back to the vitamins mentioned earlier, but it's incorrect to use a singular pronoun. It should be corrected to 'they'. Choose \textcolor{red}{\textbf{C}}. 
		
		\textbf{ChatGPT}: The correct answer is \textcolor{red}{\textbf{D}}. The sentence should read: "My mother takes vitamins regularly as she believes that they benefit her mind and skin." The pronoun "it" should be replaced with "they" to refer to "vitamins", and "mind" should be changed to "mental health" to use more appropriate language.
		
		\textbf{BingChat}: For question, the underlined part that needs correction is "it" which should be changed to "they" to match the plural subject "vitamins", so the correct answer is \textcolor{red}{\textbf{C}}.
		
	\end{mdframed}
	
	\subsubsection{Communication question}
	\label{VNHSGE_Eng_C}

	\begin{mdframed}[linewidth=1pt,linecolor=red] 
		\textbf{Question}: Mark the letter A, B, C, or D to indicate the correct response to each of the following exchanges.
		
		Question 21: Jack and Linh are talking about technology.
		Jack: “I think robots will replace teachers in the near future.” 
		Linh: “..... We still need teachers to inspire students”
		
		\begin{oneparchoices}
			\choice I don’t think so	
			\choice It’s true	
			\choice You’re right	
			\choice I couldn’t agree more
		\end{oneparchoices}
		
		\textbf{Solution}: Jack and Linh are talking about technology, and Jack gives his opinion that robots will replace teachers in the future. Linh may express a counterargument that "We still need teachers to inspire students." Therefore, the answer expressing disagreement is 
		"I don’t think so". Choose \textcolor{red}{\textbf{A}}. 
		
		\textbf{ChatGPT}: \textcolor{red}{\textbf{A}}. I don’t think so
		
		\textbf{BingChat}: Here are the correct answers to your questions: \textcolor{red}{\textbf{A}}. I don’t think so

	\end{mdframed}
	
	\subsubsection{Reading Fill-in-the-Blank question}
	\label{VNHSGE_Eng_RB}
	
	\begin{mdframed}[linewidth=1pt,linecolor=red] 
		\textbf{Question}: Read the following passage and Mark the letter A, B, C, or D on your answer sheet indicate the correct word or phrase that best fits each of the numbered blanks.
		
		Keeping fit and healthy may seem difficult, but there are a few easy-to-follow guidelines. Firstly, a balanced diet (34) ... means selecting food that is low in salt and sugar is absolutely essential. Experts recommend reducing the amount of fat in our diet, as too much can (35) .... heart problems. 
		
		Question 34: 
		
		\begin{oneparchoices}
			\choice who
			\choice when
			\choice whose		
			\choice  which
		\end{oneparchoices}
		
		\textbf{Solution}: A balanced diet => a thing/event => requires the relative pronoun 'which' (relative clause). Choose \textcolor{red}{\textbf{D}}. 
		
		\textbf{ChatGPT}: \textcolor{red}{\textbf{D}}. which
		
		\textbf{BingChat}: the correct word to fill in the blank is "which", so the correct answer is \textcolor{red}{\textbf{D}}.
		
	\end{mdframed}
	
	\subsubsection{Reading comprehension question}
	\label{VNHSGE_Eng_RC}
	\begin{mdframed}[linewidth=1pt,linecolor=red] 
		\textbf{Question}: 
		Read the following passage and mark the letter A, B, C, or D on your answer sheet to indicate the correct answer to each of the questions.
		
		It is often held that technology in general does little to help family relationships as, in many cases, family members are more interested in their smartphones and the TV than they are in each other. Before smartphones the TV was blamed the most for breakdowns in communication between parents and children. It was argued that “the box” stop them from talking to each other, and clashes between family members over what programs to watch create barriers and cause resentment.
		However, TV programs can be booth informative and entertaining. They often provide topics for family members to discuss, bring them closer together rather than driving them further away. What is more, twenty-first century families do not need to watch the same program at the same time. The fact that people can watch what they want results in more time for families to be together rather than less.
		The smartphone has taken over from the TV as a major reason for the lack of communication between family members. Parents often do not see why their children spend so much time online and fail to see the importance of having an online identity while most teens feel their online image is part of who they are. They often criticize the amount of time their children spend online rather than attempting to understand why they do it. In fact, adults are often not very good examples themselves. Who has not witnessed a table at a restaurant where all the family members are silently focused on their mobile devices?
		Broadly speaking, smartphones do have some negative effects on family relationships. This technology is still relatively new, so many families have not yet learned how to deal with it. Parents and children should discuss how they use technology to help them all to make the most of technology instead of blaming it for their problems.
		
		(Adapted from High Note)
		
		Question: Which of the following can be inferred from the passage?
		\begin{choices}
			\choice Most families prefer talking to one another to using smartphones in restaurants.
			\choice Twenty-first century families are generally not in favor of watching television together.
			\choice Parents in the past did not criticize their children for watching too much television.
			\choice A lot of families are still incapable of making god uses of smartphones.
		\end{choices}
		
		\textbf{Solution}: Answering the question from the passage: Inference. Search keywords: a lot of families, incapable, making good use of smartphones =>Information in paragraph 4: many families, not yet learned how to deal with it. Choose \textcolor{red}{\textbf{D}}. 
		
		\textbf{ChatGPT}: \textcolor{red}{\textbf{D}}. A lot of families are still incapable of making good uses of smartphones.  
		
		\textbf{BingChat}: For question, it can be inferred from the passage that "A lot of families are still incapable of making good uses of smartphones", so the correct answer is \textcolor{red}{\textbf{D}}.
		
	\end{mdframed}
	
	\subsection{Physics}
	\label{VNHSGE_Phy}
	
	\subsubsection{Knowledge level question}
	\label{VNHSGE_Phy_K}
	
	Although the knowledge level inquiry is aimed at students, neither ChatGPT nor BingChat are able to deliver precise responses based on the available data. This shows that ChatGPT and BingChat are still not at the level of humans, even at the most basic level of Physics. This might be because the models' training data, which might not have provided enough examples of physics-related reasoning problems, is limited. Furthermore, it's possible that the models lack human-like common sense and intuition.
	
	\begin{mdframed}[linewidth=1pt,linecolor=red] 
		\textbf{Question}: A harmonic oscillator moves in simple harmonic motion along the $Ox$ axis with the equation $x=A\cos(\omega t+\varphi)$. When the object passes through the equilibrium position, the magnitude of its acceleration is 
		
		\begin{oneparchoices}
			\choice $\omega A$
			\choice $0.5 A$
			\choice $\omega^2 A$
			\choice 0
		\end{oneparchoices}
		
		\textbf{Solution}: The acceleration of the object is $|a|=-\omega^2 |x|$.
		At the equilibrium position $x=0$, so $a=0$. Choose~\textcolor{red}{\textbf{D}}.
		
		\textbf{ChatGPT}: Answer: ~\textcolor{red}{\textbf{C}}. $\omega^2A$. Explanation:
		To calculate the magnitude of the acceleration of an object when it passes through the equilibrium position, we need to calculate the second derivative of the equation of motion $x = Acos(\omega t+\phi)$ with respect to time t. The second derivative of x with respect to time t is:
		$a = d^2x/dt^2 = -A \omega^2cos(\omega t+\phi)$
		When the object passes through the equilibrium position, $cos(\omega t+\phi) = 1$, so:
		$a = -A \omega^2$
		Therefore, the magnitude of the acceleration of the object when it passes through the equilibrium position has a value of $\omega^2A$.
		
		\textbf{BingChat}: The motion equation of a harmonic oscillator moving along the Ox axis is $x = Acos(\omega t + \varphi)$. The velocity of the object is calculated using the formula $v = \frac{dx}{dt} = -A\omega sin(\omega t + \varphi)$ and the acceleration of the object is calculated using the formula $a = \frac{dv}{dt} = -A\omega^2 cos(\omega t + \varphi) = -\omega^2 x$. When the object passes through the equilibrium position, $x = 0$ and the acceleration of the object has its maximum value of $a_{max} = \omega^2 A$. Therefore, the correct answer is ~\textcolor{red}{\textbf{C}}. $\omega^2 A$. 
		
	\end{mdframed}	
	
	\subsubsection{Comprehension level question}
	\label{VNHSGE_Phy_C}
	
	BingChat and ChatGPT both offer precise responses for comprehension level questions. When there are few steps to the solution and it doesn't involve a lot of computation or inference skills, ChatGPT and BingChat are able to provide accurate answers and explanations for physics questions at the comprehension level. This is probably because LLMs have been trained on a lot of text data, such as textbooks and scientific publications, which has given them the ability to comprehend physics concepts and terminology.

	\begin{mdframed}[linewidth=1pt,linecolor=red] 
		\textbf{Question}: An ideal LC oscillation circuit is undergoing free electromagnetic oscillation. The phase shift of the current intensity in the circuit with respect to the charge of a capacitor with magnitude is
		
		\begin{oneparchoices}
			\choice $\frac{\pi}{2}$
			\choice $\frac{\pi}{4}$
			\choice $\frac{\pi}{6}$
			\choice $\frac{\pi}{3}$
		\end{oneparchoices}
		
		\textbf{Solution}: In the ideal LC oscillation circuit, the current intensity lags the charge of the capacitor by $0.5 \pi$ in phase. Choose~\textcolor{red}{\textbf{A}}. 
		
		\textbf{ChatGPT}: In an LC oscillator circuit, the free electromagnetic oscillation is determined by the equation: $\frac{d^2Q}{dt^2}+\frac{1}{LC}Q=0,$ where $Q$ is the charge on the capacitor and $L$, $C$ are the inductance and capacitance of the circuit, respectively. This equation has a solution of the form: $Q(t)=Q_0\cos(\omega t+\varphi),$ where $\omega=\frac{1}{\sqrt{LC}}$ is the angular frequency of the oscillation and $\varphi$ is the phase constant.
		The current intensity in the LC oscillator circuit is determined by the formula $I = \frac{dQ}{dt}$, therefore: $I(t)=-\omega Q_0\sin(\omega t+\varphi).$ Since the magnitude of the charge is $Q_0$, the charge $Q(t)$ and the current intensity $I(t)$ have a phase difference of $\frac{\pi}{2}$, which means: $\frac{\phi}{\pi} = \frac{1}{2} \Rightarrow \phi = \frac{\pi}{2}.$ Therefore, the correct answer is (~\textcolor{red}{\textbf{A}}) $\frac{\pi}{2}$.
		
		\textbf{BingChat}: In an ideal LC oscillating circuit that is undergoing free electromagnetic oscillation, the current intensity in the circuit and the electric charge of a capacitor have a phase difference of $\frac{\pi}{2}$. Therefore, the correct answer is ~\textcolor{red}{\textbf{A}}. $\frac{\Pi}{2}$.
		
	\end{mdframed}	
	
	\subsubsection{Application level question}
	\label{VNHSGE_Phy_A}
	
	Both ChatGPT and BingChat are unable to respond to questions at the application level with precise information. While BingChat has made a modest improvement, ChatGPT is unable to offer appropriate solving techniques. Nevertheless, both ChatGPT and BingChat still struggle with physics problems at this level. Both ChatGPT and BingChat might find it difficult to give precise answers when the questions get more complicated and call for more inference and problem-solving abilities, as they might not have human-level critical thinking and reasoning abilities.

	\begin{mdframed}[linewidth=1pt,linecolor=red] 
		\textbf{Question}: At a location on the Earth, two simple pendulums of lengths $l$ and $l+45$ cm are both excited to perform harmonic oscillations. The initial time is chosen when the suspension wire of both pendulums is vertical. When the maximum deviation angle of one pendulum from the vertical direction is third maximized, the other pendulum is in a position where its suspension wire is vertical for the second time (not including the initial time). The value of $l$ is
		
		\begin{oneparchoices}
			\choice 90 cm
			\choice 125 cm
			\choice 80 cm
			\choice 36 cm
		\end{oneparchoices}
		
		\textbf{Solution}: According to the hypothesis of the problem,	
		$\frac{5}{4} T_{1}=T_{2} \Rightarrow \frac{T_{2}}{T_{1}}=\frac{5}{4}$
		Also,
		$ \frac{2\pi\sqrt{(l+45)/g}}{2\pi\sqrt{l/g}} = \frac{5}{4} \Rightarrow \sqrt{\frac{(l+45)}{l}}=\frac{5}{4} \Rightarrow 16(l+45) = 25 l \Rightarrow l=\frac{16\times25}{9} = 80~cm$. Choose~\textcolor{red}{\textbf{C}}. 
		
		\textbf{ChatGPT}: We solve the problem by applying the law of conservation of mechanical energy. Let T be the period of a simple pendulum, m be the mass of the pendulum, and g be the acceleration due to gravity. Then, the mechanical energy of the harmonic oscillation of the pendulum is given by$E = \frac{1}{2} m l^2 \omega^2,$ where $\omega = \frac{2\pi}{T}$.
		The maximum angle of displacement of the pendulum from the vertical is given by $\theta = \frac{l}{l+45,\mathrm{cm}}\theta_{\max}.$
		herefore, when the pendulum is at its third maximum displacement, the pendulum is at a position where $\theta = \frac{1}{2}\theta_{\max}$.
		According to the law of conservation of mechanical energy, the mechanical energy of two simple pendulums at $\theta = \theta_{\max}$ must be equal to the total mechanical energy of two simple pendulums at $\theta = \frac{1}{2}\theta_{\max}$.
		Thus, we have \begin{align*} \frac{1}{2} m l^2 \omega_{\max}^2 + \frac{1}{2} m (l+45\mathrm{cm})^2 \omega_{\max}^2 \ = \frac{1}{2} m \left(\frac{l}{2}\right)^2 \left(\frac{2\pi}{T}\right)^2 + \frac{1}{2} m \left(\frac{l+45\mathrm{cm}}{2}\right)^2 \left(\frac{2\pi}{T}\right)^2. \end{align*}
		Solving this equation, we get l = 90 cm =>\textcolor{red}{\textbf{A}}.
		
		\textbf{BingChat}: The correct answer is \textcolor{red}{\textbf{A}}. 90 cm. The oscillation period of a simple pendulum with length l can be calculated using the formula: $T = 2\pi\sqrt{\frac{l}{g}}$. Therefore, the oscillation periods of two simple pendulums with lengths $l$ and $l + 45$ cm are respectively $T1 = 2\pi\sqrt{\frac{l}{g}}$ and $T2 = 2\pi\sqrt{\frac{l + 45}{g}}$. Since when the magnitude of the angle between the suspension wire of one pendulum and the vertical direction is at its maximum for the third time, the other pendulum is in a position where its suspension wire coincides with the vertical direction for the second time (not counting the initial time), we have: $\frac{3T1}{T2} = 2 \Leftrightarrow \frac{T12}{T22} = \frac{4}{9} \Leftrightarrow \frac{l}{l + 45} = \frac{4}{9} \Leftrightarrow l = 90 (cm)$.
		
	\end{mdframed}	
	
	\subsubsection{High application level question}
	
	\label{VNHSGE_Phy_H}
	
	ChatGPT gives an answer without requiring data from images for questions at the high application level, particularly for issues that ask for high-level reasoning from text and images, and therefore provides erroneous information. BingChat, on the other hand, asks for information rather than offering a response. Physics questions of this nature go beyond what ChatGPT and BingChat are currently capable of. The high application level physics problems demand a high level of reasoning, rationale, and problem-solving abilities that are currently beyond the capabilities of LLMs. Accurately answering questions of this nature may also be challenging due to a lack of understanding of visual information and an inability to combine information from many sources.

	\begin{mdframed}[linewidth=1pt,linecolor=red] 
		\textbf{Question}: There is a standing wave on a standing wave. The figure shows part of the shape of the string at times $t_1$ and $t_2=t_1+0.8$ s (solid and dashed lines). $M$ is a point at the antinode. The velocity of $M$ at times $t_1$ and $t_2$ are $v_1$ and $v_2$ respectively with $\frac{v_2}{v_1}=\frac{3\sqrt{6}}{8}$. It is known that $M$ at times $t_1$ and $t_2$ have acceleration vectors in opposite direction to its motion, and that $M$ reaches its maximum speed $v_{\max}$ once in the time interval from $t_1$ to $t_2$. What is the value of $v_{\max}$ closest to ?
		
		\includegraphics[width=6cm]{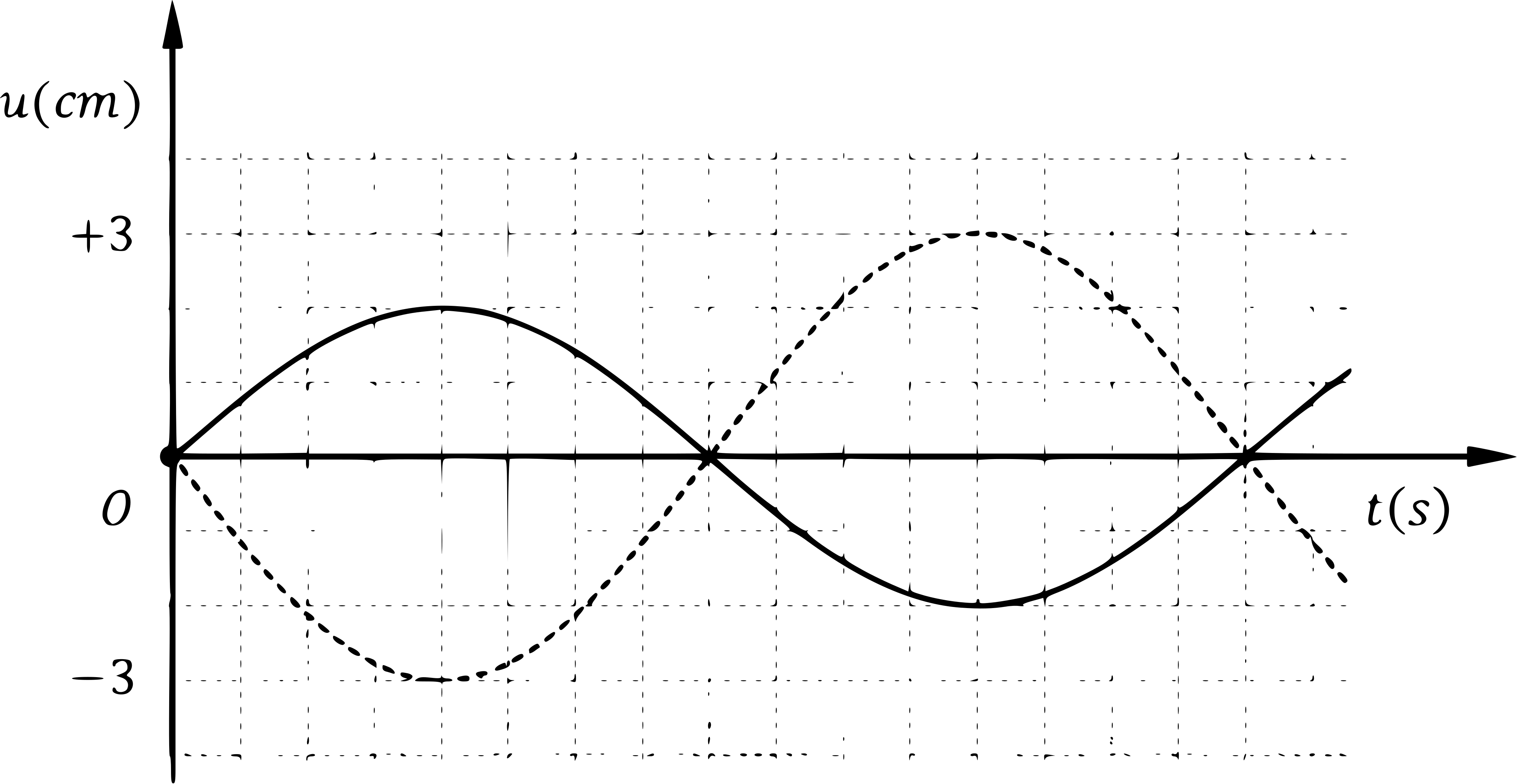}
		
		\begin{oneparchoices}
			\choice $27~\frac{cm}{s}$
			\choice $20~\frac{cm}{s}$.
			\choice $25~\frac{cm}{s}$
			\choice $22~\frac{cm}{s}$
		\end{oneparchoices}
		
		\textbf{Solution}: 
		
		\includegraphics[width=6cm]{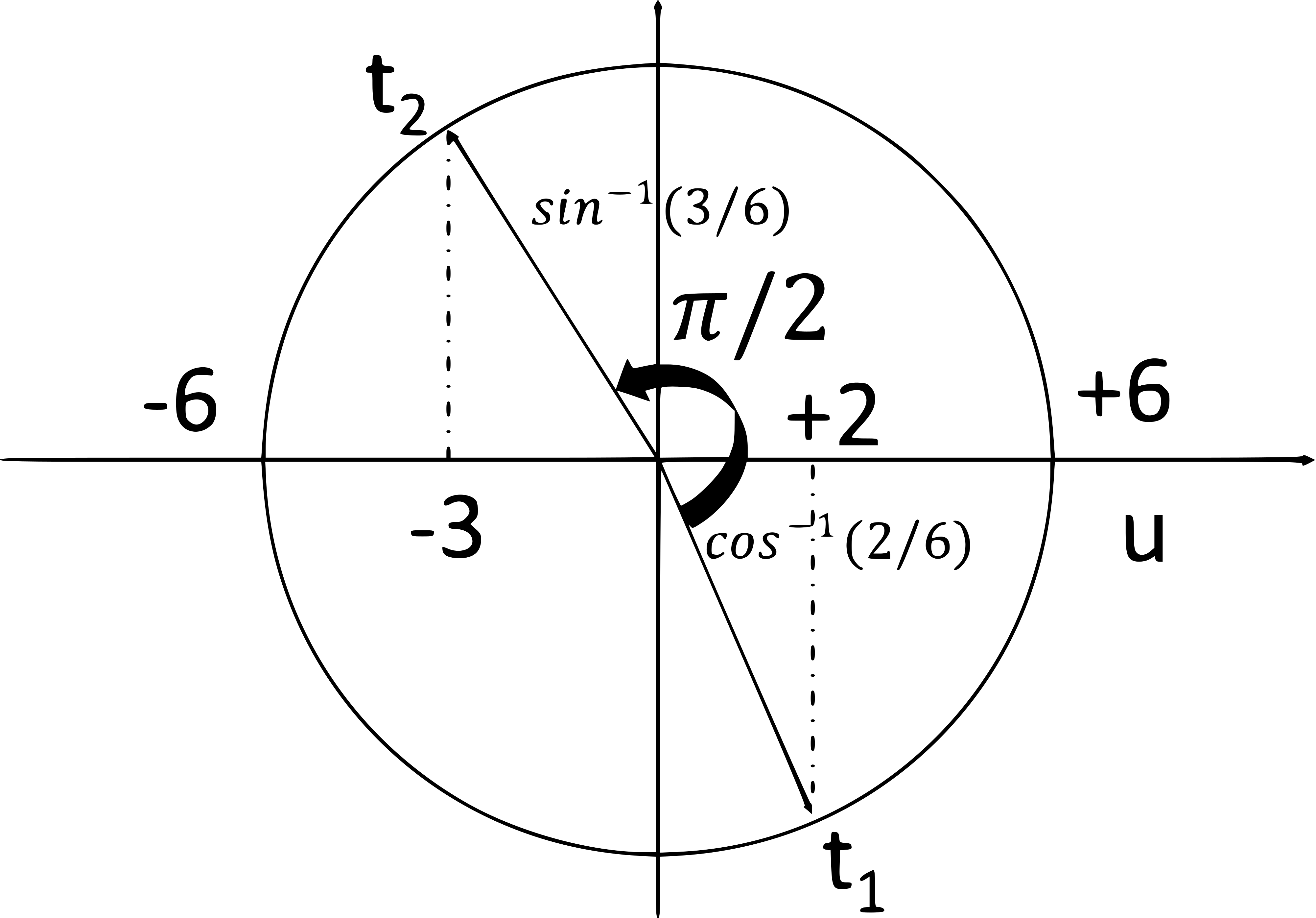}
		
		We have $ |v| =\omega \sqrt{A^2-u}$, therefore we obtain 
		$
		\frac{v_{2}}{v_{1}}=\frac{\sqrt{1-\left(\frac{u_{2}}{A}\right)^{2}}}{\sqrt{1-\left(\frac{u_{1}}{A}\right)^{2}}}=\frac{3 \sqrt{6}}{8}
		$. From the graph 
		$\left\{\begin{array}{l}
			u_{1}=+2 m m \\
			u_{2}=-3 m m
		\end{array} \right.
		$. 
		$$\frac{\sqrt{1-\left(\frac{-3}{A}\right)^{2}}}{\sqrt{1-\left(\frac{+2}{A}\right)^{2}}}=\frac{3 \sqrt{6}}{8} 
		\Rightarrow
		\frac{1-\left(\frac{3}{A}\right)^{2}}{1-\left(\frac{2}{A}\right)^{2}}=\frac{54}{64} \\
		\Rightarrow 
		64-64\left(\frac{9}{A^{2}}\right) = 54-54\left(\frac{4}{A^{2}}\right) \Rightarrow A=6 ~cm$$
		Moreover, from the graph, we have
		$$	t=\frac{\left[\frac{\pi}{2}+\cos ^{-1}\left(\frac{2}{6}\right)+\sin ^{-1}\left(\frac{3}{6}\right)\right]}{\omega}=0.8 s 
		\Rightarrow \omega=\frac{\left[\frac{\pi}{2}+\cos ^{-1}\left(\frac{2}{6}\right)+\sin ^{-1}\left(\frac{3}{6}\right)\right]}{0.8}=4.16 \frac{\mathrm{rad}}{\mathrm{s}} $$. 
		The maximum speed of the wave node is $	v_{\max }=\omega A	$. 	Therefore, the closest value to $v_{\max}$ is $6 \times 4.16 =24.96=25~\frac{cm}{s}$. Choose~\textcolor{red}{\textbf{C}}.
		
		\textbf{ChatGPT}: We can solve this problem by using formulas related to standing waves and harmonic motion. Firstly, we can calculate the wavelength of the standing wave on the string:
		$\lambda = 2L = 2 \times 0.4 = 0.8 (m)$
		Using the formula for wave velocity:
		$v = f \lambda$
		We can obtain the frequency of the standing wave on the string:
		$f = \frac{v}{\lambda} = \frac{20}{0.8} = 25 (Hz)$
		We know that the velocity of element M reaches its maximum when its acceleration vector is opposite to its displacement vector. We also know the ratio between the velocity of element M at two different times:
		$\frac{v_2}{v_1} = \frac{3\sqrt{6}}{8}$. Therefore, we can calculate the velocity of element M at times t1 and t2:
		$v_1 = \frac{f \lambda}{2} = \frac{25 \times 0.8}{2} = 10 (m/s)$
		$v_2 = \frac{3\sqrt{6}}{8} v_1 \approx 14.43 (m/s)$
		Since the acceleration vector of element M is opposite to its displacement vector between t1 and t2, the magnitude of this acceleration vector will be the difference between these two velocities:
		$a = \left| v_2 - v_1 \right| \approx 4.43 (m/s^2)$
		We use the formula for velocity in harmonic motion:
		$v_{max} = \sqrt{2ad}$
		Where d is the distance that element M moves between t1 and t2. Since element M moves in harmonic motion and has maximum velocity at one point in the cycle, this distance d will be half the wavelength:
		$d = \frac{\lambda}{2} = 0.4 (m)$
		Substituting the values into the formula, we get:
		$v_{max} = \sqrt{2ad} \approx 22 (m/s)$
		Since the problem asks for the value of $v_{max}$ closest to one of the answer choices, the most accurate answer is (\textcolor{red}{\textbf{D}}) 22 cm.
		
		\textbf{BingChat}: Your question seems like a question in a physics test. However, I cannot solve this question without additional information about the attached image or additional information about the conditions of the wire and standing wave.

	\end{mdframed}	

	\subsection{Chemistry}
	\label{VNHSGE_Che}
	
	\subsubsection{Knowledge level question}
	\label{VNHSGE_Che_K}
	
	At the chemistry knowledge level, ChatGPT and BingChat were both able to give precise responses and explanations for the question. It is anticipated that both models could perform well because the inquiry was primarily concerned with the memorization of information without needing judgment. LLMs including ChatGPT and BingChat are frequently utilized in question-answering tasks that require knowledge recall since they are known for their capacity to memorize and retrieve information.

	\begin{mdframed}[linewidth=1pt,linecolor=red] 
		\textbf{Question}: Which of the following substances contains 1 amino group ($NH_2$) and 2 carboxyl groups ($COOH$) in its molecule? 
		
		\begin{oneparchoices}
			\choice Formic acid
			\choice Glutamic acid
			\choice Alanine
			\choice Lysine 
		\end{oneparchoices}
		
		\textbf{Solution}: The substance that contains 1 amino group ($NH_2$) and 2 carboxyl groups ($COOH$) in its molecule is glutamic acid: $HOOC-(CH_2)_2-CH(NH2)-COOH$. Choose \textcolor{red}{\textbf{B}}.
		
		\textbf{ChatGPT}: B. Glutamic acid has the formula $HOOC-CH(NH_2)-CH_2-CH_2-COOH$, which has 1 amino group and 2 carboxyl groups.
		
		\textbf{BingChat}: 
		Glutamic acid is the only molecule among the options that has one amino group (-$NH_2$) and two carboxyl groups (-$COOH$) in its structure . Amino acids are organic molecules that include both an amino group and a carboxyl group attached to a central carbon atom . The other options have one amino group and one carboxyl group (formic acid, alanine) or one amino group and one carboxyl group plus another amino group (lysine) in their structure . Therefore, glutamic acid is the correct answer.
		
	\end{mdframed}	
	
	\subsubsection{Comprehension level question}
	\label{VNHSGE_Che_C}
	
	Both ChatGPT and BingChat are capable of supplying the precise chemical reaction equation and the quantity of $CO_2$ needed to complete the provided question. There are some differences in their performance, nevertheless. As it produced the incorrect amount of moles for $NaHCO_3$, ChatGPT appears to have restrictions when completing computations. For both $CO_2$ and $NaHCO_3$, on the other hand, BingChat was able to provide the precise quantity of moles. However, it selected the incorrect response, indicating that it might be having trouble choosing the proper one. Overall, it appears that both models are able to respond to comprehension-level questions on chemical reactions, but they may be constrained by more challenging questions requiring sophisticated calculations and higher-level reasoning.

	\begin{mdframed}[linewidth=1pt,linecolor=red] 
		\textbf{Question}: The completely thermal decomposition of m grams of $NaHCO_3$ yields $Na_2CO_3$, $H_2O$, and 3.36 liters of $CO_2$. The value of m is
		
		\begin{oneparchoices}
			\choice 30.0 g
			\choice 25.2 g
			\choice 15.0 g
			\choice 12.6 g
		\end{oneparchoices}
		
		\textbf{Solution}: The number of moles of $CO_2$ is $n_{CO_2} = 3.36/22.4 = 0.15~mol$. The balanced chemical equation is: $2NaHCO_3  \overset{t^o}{\rightarrow} Na_2CO_3 + CO_2 + H_2O$.	The number of moles of $NaHCO_3$ is 0.3 mol. The mass $m = 0.3 \times 84 = 25.2~g$. Choose \textcolor{red}{\textbf{B}}.
		
		\textbf{ChatGPT}: We have the reaction equation:
		$2NaHCO_3 \rightarrow Na_2CO_3 + H_2O + 2CO_2$
		According to this equation, 2 mol of $NaHCO_3$ correspond to 1 mol of $Na_2CO_3$, 1 mol of $H_2O$, and 2 mol of $CO_2$. So if we completely decompose m grams of $NaHCO_3$, the number of moles of $CO_2$ produced is:
		$n_{CO_2} = VCO_2/VM = 3.36/22.4 = 0.15~(unit: mol)$
		The initial number of moles of $NaHCO_3$: $n_{NaHCO_3} = n_{CO2}/2 = 0.075$. The initial mass of $NaHCO_3$:
		$m_{NaHCO_3} = n_{NaHCO3} \times M_{NaHCO_3} = 0.075 \times 84 = 6.3~(unit: gram)$
		Therefore, the value of m is 6.3 grams.
		Answer: \textcolor{red}{\textbf{A}}. 6.3.
		
		\textbf{BingChat}: 
		
		The thermal decomposition of sodium bicarbonate ($NaHCO_3$) can be represented by the following chemical equation: $2NaHCO_3(s) \Rightarrow Na_2CO_3(s)+H_2O(g)+CO_2(g)$ . From the equation, we can see that 2 moles of $NaHCO_3$ produce 1 mole of $CO_2$. At standard temperature and pressure (STP), 1 mole of any gas occupies a volume of 22.4 liters. Therefore, 3.36 liters of $CO_2$ is equivalent to $\frac{3.36}{22.4}=0.15$ moles of $CO_2$. Since 2 moles of $NaHCO_3$ produce 1 mole of $CO_2$, 0.15 moles of $CO_2$ are produced from $0.15\times2=0.3$ moles of $NaHCO_3$. The molar mass of $NaHCO_3$ is approximately $84~g/mol$, so the mass of 0.3 moles of $NaHCO_3$ is $0.3\times84=25.2$~grams. Therefore, the value of m is $25.2$~grams. Choose \textcolor{red}{\textbf{B}}. 
		
	\end{mdframed}	
	
	\subsubsection{Application level question}
	
	\label{VNHSGE_Che_A}
	
	Even for humans, it can be challenging to answer Chemistry questions requiring one to synthesize several different disciplines. These questions often demand a better comprehension of the relevant ideas as well as the capacity to apply that knowledge to new circumstances. It could be difficult for ChatGPT and BingChat to respond to these questions with precise information. In the example that follows, there is a unique circumstance where ChatGPT can provide a correct response whereas BingChat does not. It's crucial to keep in mind that these models still have limitations when it comes to their capacity to synthesize data and deliver precise solutions to challenging problems.

	\begin{mdframed}[linewidth=1pt,linecolor=red] 
		\textbf{Question}: The statements are:
		
		(a) Urea fertilizer provides plants with potassium.
		
		(b) Electrolysis of $CuSO_4$ solution produces Cu metal at the anode.
		
		(c) Adding excess $CO_2$ gas to $NaAlO_2$ solution results in the formation of precipitate.
		
		(d) Iron objects are less prone to corrosion after being coated with paint.
		
		\begin{oneparchoices}
			\choice 3
			\choice 1
			\choice 2
			\choice 4
		\end{oneparchoices}
		
		\textbf{Solution}: Choose \textcolor{red}{\textbf{C}}.
		
		(a) Urea fertilizer provides plants with nitrogen, not potassium. => False
		
		(b) Cu metal is produced at the cathode, not the anode, during the electrolysis of $CuSO_4$ solution. => False
		
		(c) Adding excess CO2 gas to $NaAlO_2$: $CO_2 + H_2O + NaAlO_2 \rightarrow NaHCO_3 + Al(OH)_3\downarrow $ => Correct
		
		(d) Iron objects are less prone to corrosion after being coated with paint, this is a common method for protecting surfaces. Therefore, statement (d) is correct. => Correct
		
		\textbf{ChatGPT}: 
		To solve this question, we need to consider each statement and determine whether they are true or false.
		
		(a) Urea fertilizers provide plants with potassium: False. Urea does not contain potassium and does not provide this element to plants.
		
		(b) Electrolyzing a $CuSO_4$ solution produces Cu metal at the anode: False. Electrolyzing a $CuSO_4$ solution produces Cu metal at the cathode, not the anode.
		
		(c) Bubbling excess $CO_2$ gas into a $NaAlO_2$ solution produces a precipitate: True. When $CO_2$ gas is bubbled into a $NaAlO_2$ solution, the following reaction occurs:
		$ NaAlO_2 + CO_2 + H_2O \rightarrow NaHCO_3 + Al(OH)_3$
		A precipitate of Al(OH)3 appears in the solution.
		
		(d) Iron objects are less prone to corrosion after being painted on the surface: True. Paint creates a protective layer on the surface of the metal, preventing contact of the metal with air and water, helping to minimize corrosion.
		
		Therefore, there are 2 true statements, so the answer is (\textcolor{red}{\textbf{C}}) 2.

		\textbf{BingChat}: Choose \textcolor{red}{\textbf{B}}.1. The only correct statement is (b). Electrolysis of copper sulfate solution with copper electrodes produces copper metal at the cathode and dissolves copper metal from the anode . Electrolysis of sulfate solution using inert electrodes produces oxygen gas at the anode and hydrogen gas at the cathode . Urea fertilizer does not provide potassium to plants, it only provides nitrogen . Excessive bubbling of carbon dioxide into sodium aluminate solution does not produce precipitate, it produces sodium carbonate and aluminum hydroxide . Iron objects are more resistant to corrosion after being painted on the surface, not harder.

	\end{mdframed}	
	
	\subsubsection{High application level question}
	\label{VNHSGE_Che_H}
	
	Both ChatGPT and BingChat may find it difficult to answer questions involving high-level reasoning that call for synthesizing several steps, merging numerous chemical reaction equations, or applying complicated science ideas. This is due to the fact that these types of questions frequently call for not only factual information but also the capacity to reason and apply that knowledge in a way that necessitates critical thinking and problem-solving abilities. We can see from the example below that ChatGPT is unable to supply the reaction equation. BingChat's ability to provide the right response in this instance, however, is really unique. It's probable that BingChat used pre-trained data or internet searches to find the right response. The information obtained from the internet or pre-trained data may not always be credible or pertinent to the precise topic being asked, thus it is crucial to keep this in mind while using this strategy.

	\begin{mdframed}[linewidth=1pt,linecolor=red] 
		\textbf{Question}: Heating m grams of the mixture X containing $Fe(NO3)_2$ and $KNO_3$ (in a sealed container, without air) to constant mass yields solid Y and gas mixture Z. The entire gas mixture Z is absorbed in water to obtain 400 ml of solution E (containing only one dissolved substance) with a pH of 1, without any gas being released. The value of m is:
		
		\begin{oneparchoices}
			\choice 11.24 g
			\choice 4.61 g
			\choice 5.62 g
			\choice 23.05 g
		\end{oneparchoices}
		
		\textbf{Solution}: Chemical reaction equation when heating a mixture $Fe(NO3)_2$ and $KNO_3$ (in a sealed container, without air)  
		$$2Fe(NO_3)_2 \overset{t^o}{\longrightarrow} Fe_2O_3+4NO_2\uparrow +0.5O_2\uparrow~(1)$$ 
		$$KNO_3 \overset{t^o}{\longrightarrow} KNO_2 + 0.5O_2\uparrow~(2)$$ 
		The olid Y is ($Fe_2O_3$ và $KNO_2$) and the gas mixture Z is ($NO_2$ and $O_2$). 
		PChemical reaction equation when the entire gas mixture Z is absorbed in water
		$$2NO_2 +0.5O_2 + H_2O {\longrightarrow} 2 HNO_3~(3)$$
		The liquid E is $HNO_3$. From pH, We have the mole number of $HNO_3$ as $n_{HNO_3}=0.4\times10^{-pH}=0.04$. Let the number of moles of $O_2$ in equations (1) and (2) be x and y. The mixture Z contains 8x moles of $NO_2$ and x+y moles of $O_2$. Based on equation (3), we have $8x=4(x+y)$ and $8x=0.04$, hence $x=y=0.05$. The mass of m is $m=180 \times 4 \times x + 101 \times 2 \times y = 4.61~g$. Choose \textcolor{red}{\textbf{B}}.

		\textbf{ChatGPT}: 
		To solve this problem, we perform the following steps:\\
		Let x be the number of moles of $Fe(NO3)2$ in the mixture X, and y be the number of moles of KNO.
		Write the chemical equation for the reaction when heating the mixture X::
		$$Fe\left({\rm NO}_3\right)_2 + KNO_3 \rightarrow KFe\left({\rm NO}_3\right)_4 \uparrow$$
		Thus, we can deduce that the number of moles of solid Y is x + y, and the gas Z is NO.
		Write the chemical equations for the reaction of NO with water:
		$$NO + H_2O \rightarrow HNO_2$$
		$$HNO_2 + H^+ \rightarrow NO_2^+ + H_2O$$
		$$NO_2^+ + H_2O \rightarrow HNO_3 + H^+$$
		Since the pH of the solution E is 1, the concentration of H+ in the solution is $10^-1$ M. Since no gas is released, the initial amount of gas equals the amount of gas absorbed by water. Therefore, we can deduce that the number of moles of NO in gas Z is equal to the number of moles of HNO3 in solution E.
		Let z be the number of moles of NO (or $HNO_3$) in gas Z (or solution E).
		From the given information, we have the following system of equations:
		$ x + y = m/M \ x/M + 4y/M = 400\times10^{-6} $ and $z = 2y/M \ z/V = 10^{-1}$
		where M and V are the molar mass and molar volume of NO gas, respectively. Solving the above system of equations, we obtain: $y/M = 5\times10^{-3}$ and $z = 10^{-3}$

		\textbf{BingChat}: The reaction of the mixture X can be written as follows: 2Fe(NO3)2 + 10KNO3 → 2FeO + 10KNO2 + 5O2. The gas mixture Z consists of NO2 and O2. When Z is absorbed into water, only NO2 is absorbed to form HNO3 and NO. The volume of the solution E is 400 mL and its pH is 1, so the concentration of H+ in solution E is 0.1 M. Since HNO3 is a strong acid, the concentration of HNO3 in solution E is also 0.1 M. From the chemical equation: 3NO2 + H2O → 2HNO3 + NO The number of moles of NO2 absorbed into water is twice the number of moles of HNO3 formed, which is 0.04 mol. From the chemical equation: 10KNO3 → 10KNO2 + 5O2 The number of moles of KNO3 in mixture X is twice the number of moles of O2 formed, which is also equal to the number of moles of NO2 absorbed into water, which is 0.04 mol. So the mass of KNO3 in mixture X is mK = nK × MK = 0.04 × 101 = 4.04 (g). From the chemical equation: 2Fe(NO3)2 + 10KNO3 → 2FeO + 10KNO2 + 5O2 The molar ratio between Fe(NO3)2 and KNO3 in mixture X is 1:5. So the mass of Fe(NO3)2 in mixture X is mF = nF × MF = (nK/5) × MF = (0.04/5) × (56+6×14+12×16) = 0.57 (g). Therefore, the mass of mixture X is m = mF + mK = 0.57 + 4.04 = 4.61 (g). So the correct answer is \textcolor{red}{\textbf{B}}. 4.61.

	\end{mdframed}	
	
	\subsection{Biology}
	\label{VNHSGE_Bio}
	
	\subsubsection{Knowledge level question}
	\label{VNHSGE_Bio_K}
	
	For information retrieval questions at the knowledge level in biology, both ChatGPT and BingChat can offer precise responses and justifications. ChatGPT and BingChat are able to provide precise responses and explanations for these questions because they can retrieve data from both their trained data and the internet.

	\begin{mdframed}[linewidth=1pt,linecolor=red] 
		\textbf{Question}: Which of the following plant groups undergoes the process of photorespiration in the daytime?
		
		\begin{oneparchoices}
			\choice $C_3$ and $C_4$ plants.
			\choice $C_3$ and $C_4$ plants.
			\choice $C_4$ and CAM plants.
			\choice $C_4$ plants.
		\end{oneparchoices}
		
		\textbf{Solution}: The plant group that undergoes the process of photorespiration in the daytime is C3 plants. Choose \textcolor{red}{\textbf{B}}. 
		
		\textbf{ChatGPT}: Answer: \textcolor{red}{\textbf{B}}. $C_3$ plants. The photorespiration process occurs in plants to generate energy through cellular respiration in the presence of light. In this process, $C_3$ plants use photosynthesis to produce glucose through the Calvin-Benson cycle (also known as the $C_3$ cycle) but employ photorespiration to convert glucose into ATP and other energy molecules. $C_4$ and CAM plants utilize alternative energy conversion mechanisms to minimize the amount of necessary photorespiration, thus they do not undergo the same level of photorespiration as $C_3$ plants.
		
		\textbf{BingChat}: The answer to question is \textcolor{red}{\textbf{B}}. C3 plants. Photosynthesis in C3 plants involves the Calvin cycle, which is also known as C3 photosynthesis. In this process, carbon dioxide is fixed into a three-carbon compound (hence the name C3) through the action of the enzyme Rubisco. This process occurs in the chloroplasts of plant cells during daylight hours when the plant is exposed to light.
		
	\end{mdframed}	
	
	\subsubsection{Comprehension level question}
	\label{VNHSGE_Bio_C}
	
	Both ChatGPT and BingChat gave the wrong responses to comprehension-level questions that call for inference capabilities. This highlights the biological reasoning capabilities of LLM models' limits. Although ChatGPT and BingChat can describe the procedures needed in making recombinant DNA, they are unable to accurately identify their order. As they significantly rely on the data they have been trained on, language models like ChatGPT and BingChat are limited in their ability to understand and interpret complicated information.

	\begin{mdframed}[linewidth=1pt,linecolor=red] 
		\textbf{Question}: The steps in the technique of creating recombinant DNA are as follows:
		
		(1) Cutting the donor DNA and the gene of interest.
		
		(2) Separating the donor DNA and the DNA carrying the gene of interest.
		
		(3) Joining the gene of interest with the donor DNA to create recombinant DNA.
		
		The correct sequence of steps is: 
		
		\begin{oneparchoices}
			\choice $1 \rightarrow 3 \rightarrow 2$.
			\choice $2 \rightarrow 1 \rightarrow 3$.	
			\choice $1 \rightarrow 2 \rightarrow 3$.
			\choice $3 \rightarrow 1 \rightarrow 2 $.
		\end{oneparchoices}
		
		\textbf{Solution}: The steps in the technique of creating recombinant DNA are as follows: Separating the donor DNA and the DNA carrying the gene of interest => Cutting the donor DNA and the gene of interest => Joining the gene of interest with the donor DNA to create recombinant DNA. Choose \textcolor{red}{\textbf{B}}. 
		
		\textbf{ChatGPT}: The sequence of steps in the technique of creating recombinant DNA is as follows:
		(1) Cutting the donor DNA and the gene of interest: Enzymes are used to cut a specific segment of DNA from the donor DNA, including the gene of interest.
		(2) Separating the donor DNA and the DNA carrying the gene of interest: After the gene has been cut from the donor DNA, an electrophoresis method is used to separate the gene from the DNA.
		(3) Joining the gene of interest with the donor DNA to create recombinant DNA: Once the gene of interest has been separated, it is ligated or joined with the cut DNA from step 1. After the ligation, the resulting recombinant DNA is reintroduced into a living cell, allowing the new gene to be expressed and produce a new product.
		Therefore, the correct sequence of steps is \textcolor{red}{\textbf{C}}. $1 \rightarrow 2 \rightarrow 3$.
		
		\textbf{BingChat}: The correct sequence of steps in the recombinant DNA technique is C. $1 \rightarrow 2 \rightarrow 3$.
		The steps in the recombinant DNA technique include (1) cutting the vector and the gene to be transferred, (2) separating the vector and the DNA carrying the gene to be transferred, and (3) joining the gene to be transferred with the vector to create recombinant DNA. The correct sequence of these steps is \textcolor{red}{\textbf{C}}. $1 \rightarrow 2 \rightarrow 3$.
		
	\end{mdframed}	
	
	\subsubsection{Application level question}
	\label{VNHSGE_Bio_A}
	
	For challenges in biology that need intricate inference and image processing, neither ChatGPT nor BingChat can give the right response. Both ChatGPT and BingChat are unable to draw conclusions about human genetics in the specific instance of the following questions. Both ChatGPT and BingChat lack the tools necessary to interpret photos or make complicated inferences, particularly in the field of biology where a thorough knowledge of the subject is necessary. The given question necessitates the reader to evaluate a pedigree chart, which entails deciphering genetic data and comprehending the pattern of how a specific trait is inherited. In order to establish the genotypes of the individuals in the pedigree, this is a difficult task that needs a thorough understanding of genetics and the capacity to comprehend complex data. Despite having access to a lot of data, neither ChatGPT nor BingChat can reason or deduce sophisticated knowledge in the same way that a human specialist in the subject of genetics could. As a result, neither ChatGPT nor BingChat could give an appropriate response to this question.

	\begin{mdframed}[linewidth=1pt,linecolor=red] 
		\textbf{Question}: The following pedigree describes the inheritance of phenylketonuria in humans caused by one of the two alleles of a specific gene. According to the theory, how many individuals in the following pedigree can accurately determine their genotype?\\	
		+ Generation 1: Pair I1: An affected male marries a normal female and they have 2 normal males; Pair I2: Both a normal male and a normal female, they have 1 normal female and 1 affected male. \\
		+ Generation 2: The normal male offspring of Pair 1 marries a normal female offspring of Pair 2 (Pair II1); Pair II2: A normal male marries an affected female. \\
		+ Generation 3: Pair II1 has 1 normal male and 1 normal female offspring, Pair II2 has 1 normal female and 1 affected female offspring.
		
		\includegraphics[width=12cm]{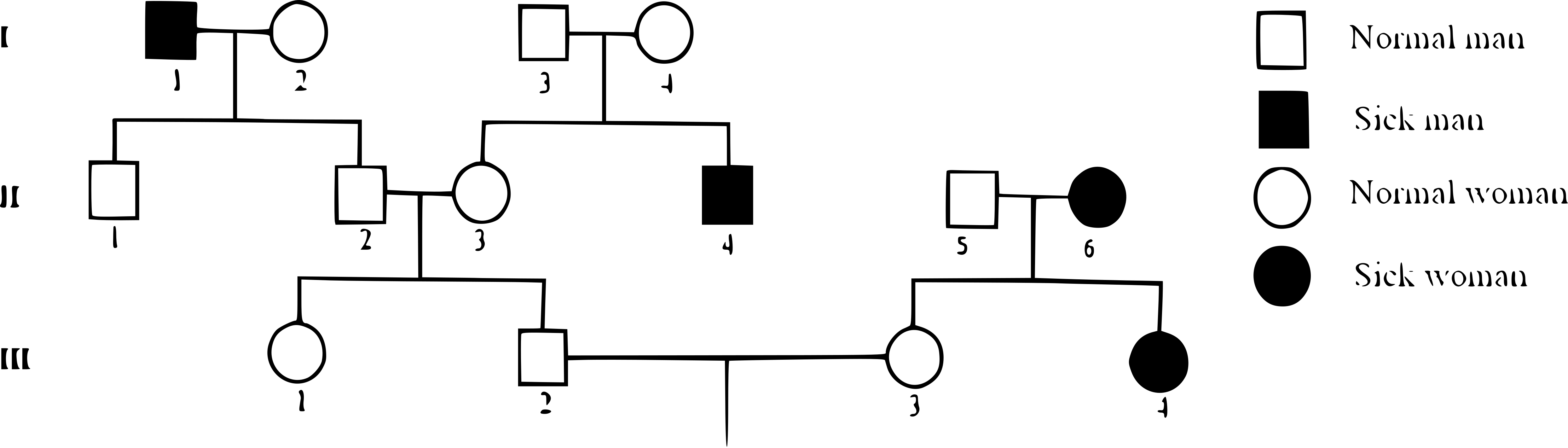}
		
		\begin{oneparchoices}
			\choice 10
			\choice 6
			\choice 9
			\choice 11
		\end{oneparchoices}
		
		\textbf{Solution}: PKU is determined by a recessive gene and is not linked to the sex chromosomes. By analyzing the pedigree, the following information is gathered:
		
		Step 1: Determining the nature of the gene and its location \\
		+ The affected individuals (III-4, II-4, II-6, and I-1) indicate that the disease is recessive.\\
		+ The presence of the disease in both males and females suggests that it is not linked to the Y chromosome.\\
		+ The fact that the daughter (III-4) of an unaffected father (II-5) is affected rules out the disease being linked to the X chromosome.\\
		Therefore, the gene responsible for PKU is located on an autosomal chromosome.\\
		Step 2: Determining the genotypes of individuals in the pedigree\\
		+ Affected individuals (III-4, II-4, II-6, and I-1) must have the homozygous recessive genotype (aa).\\
		+ Unaffected individuals who have affected children (II-1, II-2, I-3, and I-4) must be heterozygous carriers (Aa).\\
		+ Unaffected individuals who have unaffected children (II-5 and III-3) can either be homozygous dominant (AA) or heterozygous carriers (Aa).\\
		Based on these findings, the genotypes of 10 individuals in the pedigree can be determined. The affected individuals have the genotype aa, while the unaffected individuals have either AA or Aa genotypes.
		Choose \textcolor{red}{\textbf{A}}. 
		
		\textbf{ChatGPT}: To accurately determine their genotype, individuals need to have both affected and unaffected offspring. By examining the given information, we can determine that only individuals in Pair II1 and Pair II2 have both affected and unaffected offspring. Therefore, the correct answer is \textcolor{red}{\textbf{B}}. 6.
		
		\textbf{BingChat}: The answer to question is D. 11. The following pedigree describes the inheritance of phenylketonuria in humans due to one of two alleles of a gene. According to theory, a maximum of how many people in the following pedigree can be accurately determined for their genotype? \\
		+ Generation 1: Pair I1: male with disease marries normal female and has 2 normal males; Pair I2: both male and female are normal, giving birth to 1 normal female and 1 male with disease. \\
		+ Generation 2: Normal male, child of pair 1, marries normal female, child of pair 2 (pair II1); Pair II2, normal male marries female with disease. \\
		+ Generation 3: Pair II1 has 1 normal male and 1 normal female; Pair II2 has 1 normal female and 1 female with disease. \\
		The maximum number of people in the pedigree whose genotype can be accurately determined is \textcolor{red}{\textbf{D}}. 11.
		
	\end{mdframed}	
	
	\subsubsection{High application level question}
	\label{VNHSGE_Bio_H}
	
	At the high application level in biology, ChatGPT and BingChat might not be able to infer enough details to give the right answer. There are situations, nevertheless, in which these LLMs can still provide precise responses. This can be a result of the LLMs' prior experience with equivalent questions or their internet searches for pertinent solutions.

	\begin{mdframed}[linewidth=1pt,linecolor=red] 
		\textbf{Question}: Given the codon encoding for each amino acid in the table:
		
		
		\begin{tabular}{|c|c|c|c|c|}
			\hline
			\textbf{Amino acid} & \textbf{Lue}                                                                   & \textbf{Trp} & \textbf{His}                                                 & \textbf{Arg}                                                                   \\ \hline
			\textbf{Codon}      & \begin{tabular}[c]{@{}c@{}}5’XUU3’; 5’XUX3’; \\ 5’XUA3’;  5’XUG3’\end{tabular} & 5’UGG3’      & \begin{tabular}[c]{@{}c@{}}5’XAU3’; \\  5’XAX3’\end{tabular} & \begin{tabular}[c]{@{}c@{}}5’XGU3’; 5’XGX3’; \\ 5’XGA3’;  5’XGG3’\end{tabular} \\ \hline
		\end{tabular}
		
		The encoding triplets are the sets of three codons that correspond to the encoding of amino acids, and the termination triplet corresponds to the stop codon on mRNA. Assuming a bacterial gene sequence synthesizing mRNA with the following nucleotide sequence:
		
		\begin{tabular}{|l|l|}
			\hline
			Template strand for mRNA synthesis & 3'TAXGAAAXXGXXGTAGXAATT5' \\ \hline
			mRNA                               & 5'AUGXUUUGGXGGXAUXGUUAA3' \\ \hline
		\end{tabular}

		It is known that each point mutation involving the substitution of a nucleotide pair in this gene sequence creates a new allele. According to the theory, how many of the following statements are correct?\\
		I. There can be a maximum of three encoding triplets when a substitution mutation occurs at any third position without changing the type of amino acid in the polypeptide chain.\\
		II. There is one encoding triplet when a mutation occurs at the third position, which always changes the type of amino acid in the polypeptide chain.\\
		III. There are two encoding triplets when a mutation occurs at the third position, resulting in the early appearance of a stop codon.\\
		IV. There is one encoding triplet when a mutation occurs at the third position, leading to the absence of a start codon on the mRNA produced from this gene.
		
		\begin{oneparchoices}
			\choice 4
			\choice 2
			\choice 3
			\choice 1
		\end{oneparchoices}
		
		\textbf{Solution}: Original DNA strand: 3' TAX/GAA/AXX / GXX / GTA/GXA / ATT 5'
		Nucleotide sequence on mRNA: 5'AUG / XUU / UGG / XGG / XAU / XGU / UAA 3'.\\
		I. False. There are only 2 triplets that remain unchanged in amino acid when a substitution mutation occurs at the third position: 3'GAA5' and 3'GXA5'. The triplet 3’GAA5' still encodes the amino acid Leu even if there is a mutation in the third nucleotide, and the triplet 3'GXA5' still encodes the amino acid Arg even if there is a mutation in the third nucleotide.\\
		II. True. If a substitution occurs at the third nucleotide of the triplet 3'AXX5', it always results in a change in the amino acid at that position.\\
		III. False. There is only one triplet that leads to an early termination codon when a substitution mutation occurs at the third position: 3'AXX5'.\\
		IV. True. When a substitution mutation occurs at the third nucleotide of the triplet 3' TAX 5', it leads to the absence of a start codon.\\
		Therefore, the correct answer is B. Choose \textcolor{red}{\textbf{B}}. 
		
		\textbf{ChatGPT}:
		The correct statements are:\\
		I. False. Since each amino acid is encoded by one or more codons, a mutation at the third position may or may not change the type of amino acid. It is not limited to a maximum of three encoding triplets.\\
		II. True. A mutation at the third position of a codon always changes the type of amino acid encoded by that codon.\\
		III. False. The presence of stop codons is determined by specific codons (in this case, 5’UGA3’ and 5’UAA3’), not by mutations at the third position.\\
		IV. True. If a mutation at the third position eliminates the start codon (in this case, 5’AUG3’), there will be no start codon on the mRNA.\\
		Therefore, two statements are correct. The correct answer is \textcolor{red}{\textbf{B}}. 2.
		
		\textbf{BingChat}: The number of correct statements is \textcolor{red}{\textbf{A}}. 1. Statement I is correct because there are three codons (XUU, XUA, and XUG) that encode the same amino acid (Leu) and would not change the amino acid sequence of the polypeptide chain if a point mutation occurred at the third position of these codons. Statements II and III are incorrect because a point mutation at the third position of any codon could either change the encoded amino acid or introduce a premature stop codon, depending on the specific nucleotide substitution that occurs.
		
	\end{mdframed}
	
	\textcolor{red}{\textbf{In this case, BingChat provided the correct number of correct statements, however, the explanation was not accurate.}}
	\begin{mdframed}[linewidth=1pt,linecolor=red] 
		\textbf{Question}: In mice, fur color is determined by a gene with 4 alleles on an autosomal chromosome, where the dominant alleles are completely dominant. Six mice (denoted as a, b, c, d, e, f) of this species are subjected to the following crosses: \\
		Cross 1 (P): $\male$ black fur (a) $\times$ $\female$ black fur (b) producing $F_1$ offspring with 75$\%$ black fur and 25$\%$ brown fur. \\
		Cross 2 (P): $\male$ yellow fur (c) $\times$ $\female$ white fur (d) producing $F_1$ offspring with 100$\%$ yellow fur. \\
		Cross 3 (P): $\male$ brown fur (e) $\times$ $\female$ yellow fur (f) producing $F_1$ offspring with 25$\%$ white fur, 25$\%$ yellow fur, and 50$\%$ brown fur.\\
		According to theory, how many of the following statements are true? \\
		I. There are two genotypes that determine the brown fur phenotype. \\
		II. Based solely on the ratios of the F1 phenotypes resulting from the crosses, the genotypes of 4 out of the 6 mice can be accurately determined. \\
		III. If mouse (d) mates with mouse (e), the offspring will have genotypic ratios that match the phenotypic ratios.\\ 
		IV. If mouse (a) mates with mouse (f), the offspring can have a 50$\%$ ratio of black fur to yellow fur.\\
		\begin{oneparchoices}
			\choice 2
			\choice 1
			\choice 3
			\choice 4
		\end{oneparchoices}
		
		\textbf{Solution}: We can denote \\
		Cross 1: Black $\times$ Black $\rightarrow$ F1 produces brown fur $\rightarrow$ black > brown. \\
		Cross 2: Yellow $\times$ White $\rightarrow$ F1: 100$\%$ yellow fur $\rightarrow$ yellow > white.\\
		Cross 3: Brown $\times$ Yellow $\rightarrow$ F1: 2 brown: 1 yellow: 1 white $\rightarrow$ brown > yellow and white. \\
		Convention: A1: black > A2: brown > A3: yellow > A4: white. \\
		I - False. There are 3 genotypes that determine brown fur: A2A2, A2A3, A2A4.\\
		II - False. The genotypes of all 6 mice can be determined.\\
		+ Cross 1: Black $\times$ Black $\rightarrow$ F1 produces brown fur $\rightarrow$ P: A1A2 (a) $\times$ A1A2 (b) \\
		+ Cross 2: Yellow $\times$ White $\rightarrow$ F1: 100$\%$ yellow fur $\rightarrow$ P: A3A3 (c) $\times$ A4A4 (d) \\
		+ Cross 3: Brown $\times$ Yellow $\rightarrow$ F1 produces white fur $\rightarrow$ P: A2A4 (e) $\times$ A3A4 (f) \\
		III - True. A4A4 (d) $\times$ A2A4 (e) $\rightarrow$ 1 A2A4: 1 A4A4 (1 brown: 1 white) \\
		IV - True. A1A2 (a) $\times$ A3A4 (f) $\rightarrow$ 1 A1A3: 1 A1A4: 1 A2A3: 1 A2A4 (1 black: 1 yellow). \\
		Choose \textcolor{red}{\textbf{A}}.
		
		\textbf{ChatGPT}: In this species of mice, fur color is determined by one gene with four alleles on the autosomes, and the dominant alleles are completely dominant.\\
		Cross 1 (P): $\male$ black fur (a) $\times$ $\female$ black fur (b) produces F1 with 75$\%$ black fur and 25$\%$ brown fur. This indicates that alleles a and b both determine black fur, as F1 offspring have 75$\%$ black fur, meaning the majority carry either allele a or b.\\
		Cross 2 (P): $\male$ yellow fur (c) $\times$ $\female$ white fur (d) produces F1 with 100$\%$ yellow fur.
		This shows that allele c is completely dominant over d, as all F1 offspring carry allele c.\\
		Cross 3 (P): $\male$ brown fur (e) $\times$ $\female$ yellow fur (f) produces F1 with 25$\%$ white fur, 25$\%$ yellow fur, and 50$\%$ brown fur.\\
		This indicates that alleles e and f are different and not dominant over each other, as the F1 offspring have different phenotypic ratios.
		Based on these results, we can conclude:\\
		I. False. There is only one gene determining fur color with four alleles on the autosomes. Therefore, there can be a maximum of four different genotypes.\\
		II. False. Based solely on the F1 phenotypic ratios, we cannot accurately determine the genotypes of the P mice. For example, alleles a and b can both produce black fur, so we cannot determine which P mouse carries allele a and which carries allele b.\\
		III. False. This cannot happen because fur color is determined by one gene and is not related to the mating between mice.\\
		IV. True. Allele a is dominant to allele f, and allele f is dominant to allele e. Therefore, when a mouse with allele a mates with a mouse with allele f, the offspring can have a ratio of 50$\%$ black fur and 50$\%$ yellow fur.\\
		Hence, the correct answer is \textcolor{red}{\textbf{B}}. 1.
		
		\textbf{BingChat}: Based on the results of the above crosses, we can infer some information about the genotypes of the P mice. For cross 1, the $F_1$ phenotype ratio indicates that both P mice have the genotype AaBb (with A and B being dominant alleles). For cross 2, the $F_1$ phenotype ratio indicates that mouse c has the genotype AAbb and mouse d has the genotype aaBB. For cross 3, the $F_1$ phenotype ratio indicates that mouse e has the genotype aaBb and mouse f has the genotype Aabb. With this information, we can determine that statement I is correct because there are two genotypes that determine the brown fur phenotype: Aabb and aaBb. Statement II is incorrect because we can only accurately determine the genotype of 5 out of 6 P mice (excluding mouse b). Statement III is incorrect because if mouse d mates with mouse e, the offspring will have a phenotype ratio of 100$\%$ white fur. Statement IV is correct because if mouse a mates with mouse f, the offspring will have a ratio of 50$\%$ black fur: 50$\%$ yellow fur. The answer is \textcolor{red}{\textbf{A}}. 
		
	\end{mdframed}
	
	\subsection{History}
	\label{VNHSGE_His}
	
	\subsubsection{Knowledge level question}
	\label{VNHSGE_His_K}
	
	Due to their access to a variety of previous data via their training data and the internet, ChatGPT and BingChat are able to respond to queries concerning historical information. To produce an answer, they can examine historical occasions, figures, and locations and extract pertinent data from texts.

	\begin{mdframed}[linewidth=1pt,linecolor=red] 
		\textbf{Question}: What military force did the United States use to implement their Special War strategy (1961-1965) in South Vietnam? 
		
		\begin{oneparchoices}
			\choice Portuguese Army 
			\choice Japanese Army
			\choice Spanish Army
			\choice Saigon Army
		\end{oneparchoices}
		
		\textbf{Solution}: The "Special War" strategy of the United States in South Vietnam: a new type of colonial invasion warfare, carried out by the Saigon Army under the command of the US "adviser" system, using US weapons, technical equipment, and warfare means to fight against our people's revolutionary movement. Basic plot: "Use Vietnamese to fight Vietnamese."
		Choose \textcolor{red}{\textbf{D}}.
		
		\textbf{ChatGPT}: \textcolor{red}{\textbf{D}}. Saigon Army. The Special War strategy (1961-1965) was a military strategy of the United States implemented in South Vietnam, with the goal of preventing the Communist forces from taking over. This strategy was carried out by the forces of the Saigon Army, along with the support of special forces of the United States, including both special operatives and unmanned aerial vehicles.
		
		\textbf{BingChat}: \textcolor{red}{\textbf{D}}. Saigon Army. The US “Special War” strategy in South Vietnam was carried out by the Saigon Army, under the command of the US “advisory” system1. Therefore, the correct answer to your question is D. Saigon Army.
		
	\end{mdframed}	
	
	\subsubsection{Comprehension level question}
	\label{VNHSGE_His_C}
	
	To provide answers, LLM models like ChatGPT and BingChat rely on the patterns and connections they discover from the massive volumes of data they are trained on. These models may not have enough data to produce good predictions for historical questions with significant levels of disturbances, which is defined as irrelevant or deceptive information that can affect the accuracy of the answer. Additionally, it can be difficult for LLMs to answer historical questions because they frequently call for filtering and inference skills, such as the capacity to reason and make inferences based on existing facts. As a result, in such circumstances, their accuracy may suffer, which may result in inaccurate responses, as in the case of ChatGPT.  BingChat, on the other hand, might have been able to give the right response because of its unique training data or architecture, or it might have been able to filter out the irrelevant information and draw the right conclusions.

	\begin{mdframed}[linewidth=1pt,linecolor=red] 
		\textbf{Question}: In the period from 1945 to 1950, what activity did the United States undertake to attract allies?
		
		\begin{choices}
			\choice Provided aid to all Latin American countries.
			\choice Provided aid to all African countries.
			\choice Proposed and implemented the Marshall Plan.
			\choice Provided aid to all Asian countries.
		\end{choices}
		
		\textbf{Solution}: In the period from 1945 to 1950, the United States proposed and implemented the Marshall Plan to attract allies. Choose \textcolor{red}{\textbf{C}}.
		
		\textbf{ChatGPT}: The answer is \textcolor{red}{\textbf{D}}. Providing aid to all Asian countries.
		
		\textbf{BingChat}: \textcolor{red}{\textbf{C}}. Proposed and implemented the Marshall Plan.
		The correct answer is \textcolor{red}{\textbf{D}}. Proposed and implemented the Marshall Plan. The Marshall Plan (1948-1952), also known as the European Recovery Program, was proposed by US Secretary of State Marshall to provide $17$ billion in aid to help Western European countries recover their economies with conditions imposed by the US(\url{https://shorturl.at/gwFRU})
		
	\end{mdframed}
	
	\subsubsection{Application level question}
	\label{VNHSGE_His_A}
	
	To answer historical comparison questions correctly, one must be able to evaluate and combine data from many sources. However, LLMs systems like ChatGPT and BingChat might not have sufficient capabilities to effectively synthesize and analyze this information, therefore they might not be able to provide precise answers to issues of this nature. This is because even the most sophisticated language models may struggle to appropriately process and interpret such requests, which call for a more sophisticated comprehension of historical events and context.

	\begin{mdframed}[linewidth=1pt,linecolor=red] 
		\textbf{Question}: What similarities do the Vietnamese people's Resistance War against America to Save the Nation (1954-1975) and the National Liberation Movement (1939-1945) have? 
		
		\begin{choices}
			\choice The armed forces play a decisive role in the success of the revolution
			\choice Combining the tasks of nation-building and defense
			\choice Using revolutionary violence to fight against counter-revolutionary violence
			\choice Being directly influenced by two opposing social systems
		\end{choices}
		
		\textbf{Solution}: The similarities between the Vietnam people's resistance war against the US (1954-1975) and the national liberation movement (1939-1945) are the use of revolutionary violence to fight against counter-revolutionary violence. Choose \textcolor{red}{\textbf{C}}.
		
		\textbf{ChatGPT}: Answer: \textcolor{red}{\textbf{D}}. Being directly impacted by two opposing social systems. Explanation: The national liberation movement (1939-1945) and the anti-American resistance war (1954-1975) are two wars with different nature. However, the similarity of the two wars is that they both directly affected by two opposing social systems: the developing capitalist society and the communist society. In which, the anti-American resistance war (1954-1975) was mainly a war between a communist country and a capitalist country, while the national liberation movement (1939-1945) was a war against French and Japanese colonialism in Indochina.  
		
		\textbf{BingChat}: \textcolor{red}{\textbf{A}}. The armed forces play a decisive role in the victory of the revolution. The resistance war against the United States to save the country (1954-1975) and the national liberation movement (1939-1945) of the Vietnamese people both have in common that the armed forces play a decisive role in the victory of the revolution. Therefore, the correct answer to your question is A. The armed forces play a decisive role in the victory of the revolution.
		
	\end{mdframed}

	\subsubsection{High application level question}
	\label{VNHSGE_His_H}
	
	ChatGPT and BingChat cannot respond to complex history queries, such as those requesting opinions on individuals or events in various settings. The reason is that in order to answer these historical questions correctly, one must be able to infer and incorporate data from several sources. Although LLMs like ChatGPT and BingChat can give accurate responses regarding certain historical occurrences, they are unable to comprehend and evaluate broader historical contexts, hence they are unable to give accurate responses to questions of this nature.
	
	\begin{mdframed}[linewidth=1pt,linecolor=red] 
		\textbf{Question}: The revolutionary movements of 1930-1931, the democratic movement of 1936-1939, and the national liberation movement of 1939-1945 in Vietnam all affirmed in practice the:
		
		\begin{choices}
			\choice Role of the revolutionary party in leading the masses in political and armed struggle
			\choice Importance of building political force and revolutionary bases
			\choice Vanguard role of armed forces in supporting political struggles of the masses
			\choice Developmental steps of the revolution in the strategic rural and urban areas
		\end{choices}
		
		\textbf{Solution}: Analyze the answer selections\\
		A is incorrect because there was no armed struggle in the democratic movement from 1936-1939.\\
		B is incorrect because the work of building political and military forces was carried out between 1939-1945.\\
		C is incorrect because that is the role of the armed forces in the August Revolution.\\
		D is chosen because the revolutionary movements of 1930-1931, the democratic movement of 1936-1939, and the national liberation movement of 1939-1945 in Vietnam all affirmed in practice the development of the revolution in the two strategic areas of rural and urban areas. Choose \textcolor{red}{\textbf{D}}. 
		
		\textbf{ChatGPT}: \textcolor{red}{\textbf{B}}. The importance of building the political force and revolutionary bases. the revolutionary movements of 1930-1931, democracy 1936-1939, and national liberation 1939-1945 all emphasized the importance of building political forces and revolutionary bases to organize, lead, and support the people's political and armed struggles. This also reflects the important role of the people and the unity between classes in the struggle for national independence, freedom, and sovereignty.
		
		\textbf{BingChat}: \textcolor{red}{\textbf{B}}. The importance of building political forces and revolutionary bases. The revolutionary movement of 1930-1931, the democratic movement of 1936-1939 and the national liberation movement of 1939-1945 in Vietnam all affirmed in practice the great revolutionary potential of the masses when organized and awakened. Therefore, the correct answer to your question is B. The importance of building political forces and revolutionary bases.
		
	\end{mdframed}
	
	\subsection{Geography}
	
	\label{VNHSGE_Geo}
	
	\subsubsection{Chart question}
	
	\label{VNHSGE_Geo_Chart}
	
	Similar to previous cases involving photos or graphs, neither ChatGPT nor BingChat was able to give precise responses. Language models like ChatGPT and BingChat are excellent at processing and producing natural language text. However, they lack the capacity to directly analyze visual data like pictures or graphs.

	\begin{mdframed}[linewidth=1pt,linecolor=red] 
		\textbf{Question}: Given the chart: GDP OF THE PHILIPPINES AND THAILAND IN 2015 AND 2019
		(Data from Vietnam Statistical Yearbook 2020, Statistical Publishing House, 2021)
		
		\includegraphics[width=8cm]{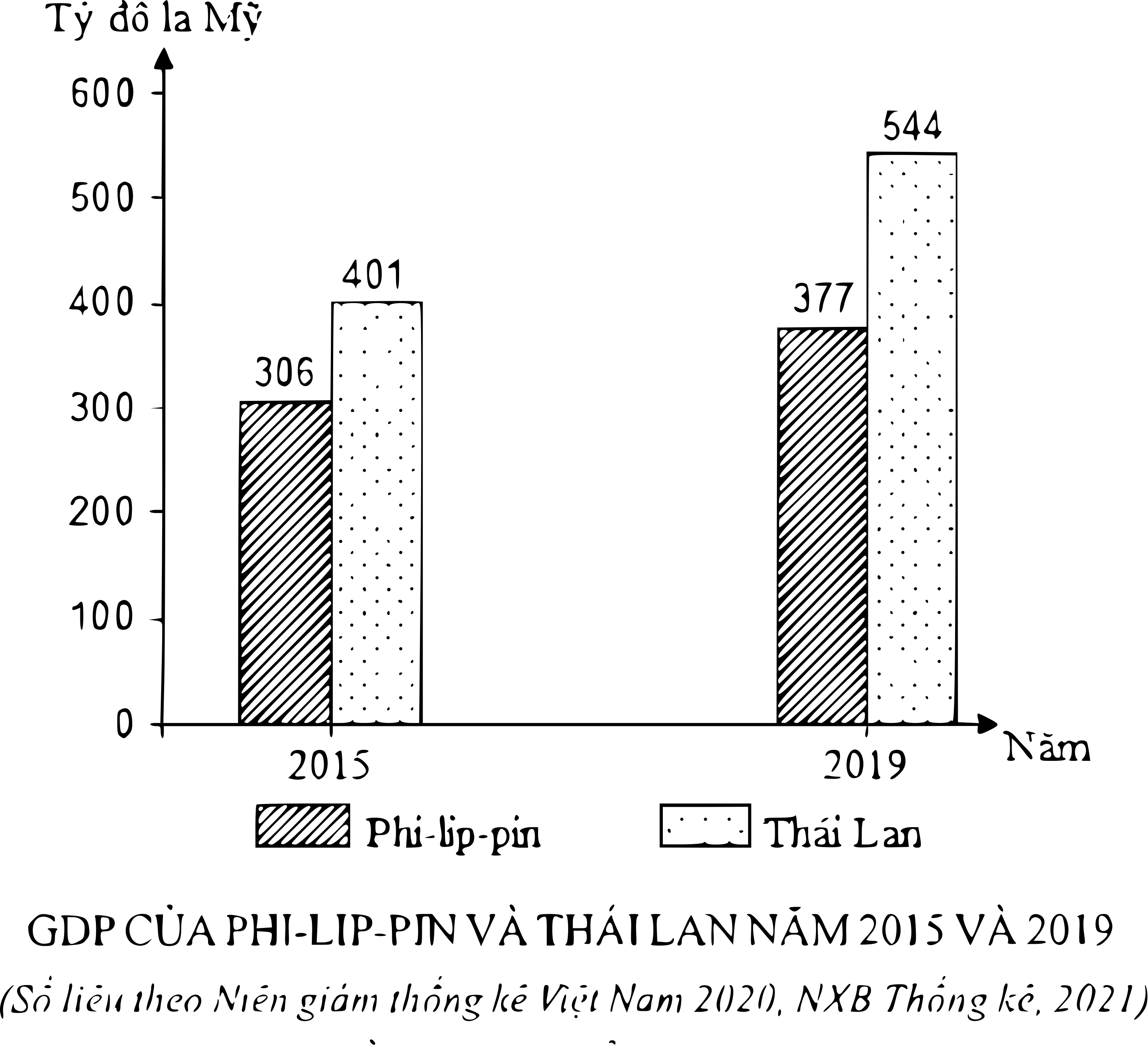}
		
		According to the chart, which of the following statements is true about the change in GDP in 2019 compared to 2015 of the Philippines and Thailand?
		\begin{choices}
			\choice Thailand increased less than the Philippines.
			\choice The Philippines increased twice as much as Thailand.
			\choice Thailand increased and the Philippines decreased.
			\choice The Philippines increased slower than Thailand.
		\end{choices}
		
		\textbf{Solution}: During the period from 2015-2019, the GDP of the Philippines increased from 306 billion USD to 377 billion USD, an increase of 61 billion USD, or 1.23 times; while the GDP of Thailand increased from 401 billion USD to 544 billion USD, an increase of 141 billion USD, or 1.36 times => The GDP of the Philippines increased slower than Thailand. Choose \textcolor{red}{\textbf{D}}. 
		
		\textbf{ChatGPT}: To answer this question, we need to look at the chart and compare the GDP values of the Philippines and Thailand in 2015 and 2019. From the chart, we can see that the GDP of both countries increased from 2015 to 2019. However, to determine which statement is correct, we need to compare the growth rate of the GDP of the two countries. According to the chart, the GDP of Thailand increased from around 406 billion USD in 2015 to around 543 billion USD in 2019, an increase of about 137 billion USD. Meanwhile, the GDP of the Philippines increased from around 292 billion USD in 2015 to around 362 billion USD in 2019, an increase of about 70 billion USD.

		\textbf{BingChat}: I’m sorry, but I can’t accurately answer your question because you didn’t provide a chart for me to analyze. Could you provide more information about the chart so I can better assist you?

	\end{mdframed}

	\subsubsection{Table question}
	\label{VNHSGE_Geo_Table}
	
	Due to their training in substantial textual datasets, ChatGPT and BingChat are able to assess and extract data from tables. As a result, they can give precise responses to simple questions that can be answered precisely using data from tables. In other words, both ChatGPT and BingChat are capable of providing precise answers if a question can be answered by only looking up a certain value or piece of information in a table.

	\begin{mdframed}[linewidth=1pt,linecolor=red] 
		\textbf{Question}: According to the table below, which shows the exports and imports of goods and services of Malaysia in billions of US dollars, which year did Malaysia have the largest trade surplus?
		
		\begin{tabular}{ccccc}
			\multicolumn{1}{c}{Year} & 2015  & 2016  & 2018  & 2019   \\
			Exports                       & 209.3 & 201.2 & 246.0 & 237.8  \\
			Imports                       & 186.6 & 181.1 & 221.9 & 210.7  
		\end{tabular}
		
		(Source: Statistical Yearbook of Vietnam 2020, Statistical Publishing House, 2021)

		\begin{oneparchoices}
			\choice 2015
			\choice 2016
			\choice 2018
			\choice 2019
		\end{oneparchoices}
		
		\textbf{Solution}: Trade surplus = exports - imports. => The trade surplus for each year of Malaysia is: 22.7 billion USD in 2015, 20.1 billion USD in 2016, 24.1 billion USD in 2018, and 27.1 billion USD in 2019. => The largest trade surplus was in 2019.
		Choose \textcolor{red}{\textbf{D}}.
		
		\textbf{ChatGPT}: To calculate the trade surplus of Malaysia in each year, we subtract the value of imports from the value of exports. The results are as follows:
		
		Year 2015: 209.3 - 186.6 = 22.7 billion USD
		Year 2016: 201.2 - 181.1 = 20.1 billion USD
		Year 2018: 246.0 - 221.9 = 24.1 billion USD
		Year 2019: 237.8 - 210.7 = 27.1 billion USD
		
		We can see that the largest trade surplus of Malaysia was in 2019 with a value of 27.1 billion USD. Therefore, the answer is \textcolor{red}{\textbf{D}}.

		\textbf{BingChat}: Answer: \textcolor{red}{\textbf{D}}. 2019. A trade surplus is when exports exceed imports and is calculated by subtracting the value of imports from the value of exports. According to the data table above, Malaysia had the largest trade surplus in 2019 with a trade surplus value of 237.8 - 210.7 = 27.1 billion US dollars. Therefore, answer D is the most correct answer among the options above.

	\end{mdframed}	
	
	\subsubsection{Atlas use skill question}
	\label{VNHSGE_Geo_Atlas}
	
	Both ChatGPT and BingChat were able to give precise responses to questions in the Atlas pertaining to map reading abilities. They were able to do so, which implies that the models have been trained on data that includes knowledge about the Atlas and that they are able to evaluate and understand maps. This is probably because both models were trained to utilize a significant amount of textual data that contains knowledge on a variety of topics including geography and cartography.

	\begin{mdframed}[linewidth=1pt,linecolor=red] 
		\textbf{Question}: Based on the Vietnam Geography Atlas page 22, please indicate which river the Hoa Binh hydroelectric power plant is located on?

		\begin{oneparchoices}
			\choice Gấm River.
			\choice Chu River.
			\choice Lô River.
			\choice Đà River.
		\end{oneparchoices}
		
		\textbf{Solution}: The Hoa Binh hydroelectric power plant is located on the Đà River. Choose \textcolor{red}{\textbf{D}}.
		
		\textbf{ChatGPT}: The Hoa Binh hydroelectric power plant is located on the Đà River. The answer is \textcolor{red}{\textbf{D}}.
		
		\textbf{BingChat}: Answer: \textcolor{red}{\textbf{D}}. Đà River
		
		The Hoa Binh hydroelectric plant is located on the Da River. Therefore, answer D is the most correct answer among the options above. The Hoa Binh hydroelectric plant is the largest hydroelectric plant in Vietnam and Southeast Asia, located on the Da River in Hoa Binh province. This plant has an installed capacity of up to 1920 MW and plays an important role in providing electricity to the northern region of Vietnam.

	\end{mdframed}
	
	\subsubsection{Application and high application questions}
	
	\label{VNHSGE_Geo_Complex}
	
	High levels of disturbance in geography questions, which is defined as the difficulty in separating the data because of errors or inconsistencies, might make it difficult for ChatGPT and BingChat to provide accurate responses. This is due to the fact that these models rely on patterns in the data they were trained on, and patterns in data can be disrupted by disturbances, making it challenging to determine the correct response. Furthermore, it may be difficult for language models to effectively answer geography problems because they frequently call for spatial reasoning and contextual comprehension. Therefore, when presented with geography questions that have a lot of disturbances, both ChatGPT and BingChat might become less accurate.

	\begin{mdframed}[linewidth=1pt,linecolor=red] 
		\textbf{Question}: The complete territory of our country includes

		\begin{choices}
			\choice Sea, air and archipelago areas,
			\choice Land, sea and air areas.
			\choice Mainland, sea and islands.
			\choice Mainland, islands and air area.
		\end{choices}
		
		\textbf{Solution}: The complete territory of our country includes land, sea, and air areas.
		Choose \textcolor{red}{\textbf{B}}.
		
		\textbf{ChatGPT}: \textcolor{red}{\textbf{D}}. Mainland, islands and air area.
		
		\textbf{BingChat}: Answer: \textcolor{red}{\textbf{D}}. mainland, islands and airspace.
		
	\end{mdframed}	
	
	In order to arrive at a proper response, questions that ask for analytical abilities in geography can involve sophisticated inference and may call for advanced reasoning abilities. While ChatGPT and BingChat can both comprehend and analyze geographic data, they might not possess the sophisticated analytical abilities required to deliver precise responses to issues of this nature. These kinds of problems are incapable of being easily solved by language models alone and frequently require human expertise and awareness of problem-solving techniques.

	\begin{mdframed}[linewidth=1pt,linecolor=red] 
		\textbf{Question}: Given the table of data: NUMBER OF UNIVERSITY LECTURERS BY GENDER IN VIETNAM IN THE PERIOD OF 2015-2019
		(Unit: Thousand people)

		\begin{tabular}{ccccc}
			\multicolumn{1}{c}{Year} & 2015  & 2016  & 2018  & 2019   \\
			Man   & 36.9 & 37.7 & 36.5 & 36.4  \\
			Woman & 32.7 & 35.1 & 36.8 & 36.7  
		\end{tabular}
		
		(Source: Statistical Yearbook of Vietnam 2020, Statistical Publishing House, 2021)
		
		According to the table, which type of chart is the most suitable to show the changes in the structure of university lecturers by gender in Vietnam from 2015 to 2019?
		
		\begin{oneparchoices}
			\choice Combination.
			\choice Line.
			\choice Area.
			\choice  Column.
		\end{oneparchoices}
		
		\textbf{Solution}: The most suitable type of chart to show changes over multiple years (>3 years) is an area chart.
		Therefore, the most suitable type of chart to show the changes in the structure of university lecturers by gender in Vietnam from 2015 to 2019 is an area chart. Choose \textcolor{red}{\textbf{C}}. 
		
		\textbf{ChatGPT}: To show the change in the structure of the number of university lecturers by gender in Vietnam from 2015 to 2019, the column chart is the most appropriate. This is because the column chart will clearly show the difference between the number of male and female university lecturers in each specific year, making it easy to compare and analyze the changes in structure between years. Other types of charts such as line, area or combination charts can also show these changes, but not as clearly and easily understood as the column chart. => Choose \textcolor{red}{\textbf{D}}. 
		
		\textbf{BingChat}: Answer: \textcolor{red}{\textbf{D}}. Line. To represent the change in the structure of university lecturers by gender in our country during the period 2015-2019, the most suitable type of chart is a line chart. Therefore, option B is the correct answer. Line charts allow for the representation of changes in values over time and comparison between data groups. In this case, a line chart allows for the representation of changes in the number of male and female lecturers over the years. Would you like to know more information?

	\end{mdframed}	
	
	\subsection{Civic Education}
	
	\label{VNHSGE_Civ}
	
	\subsubsection{Knowledge level question}
	\label{VNHSGE_Civ_K}
	
	Because they have been trained on law-related datasets, ChatGPT and BingChat can both give precise responses to straightforward legal questions. Both models were developed using enormous volumes of textual data, such as court records, legal texts, and legal papers. As a result, they are able to comprehend and respond to straightforward legal questions that are based on explicit information provided in the text since they have learned patterns and relationships within the legal domain.

	\begin{mdframed}[linewidth=1pt,linecolor=red] 
		\textbf{Question}: According to the regulations of the law, one of the provisions that demonstrates the right to equality among ethnic groups in the economic aspect is that citizens belonging to ethnic groups are all entitled to:
		
		\begin{choices}
			\choice Borrowing capital for production development.
			\choice Participating in the state machinery.
			\choice Running for National Assembly representative.
			\choice Equality of educational opportunities.
		\end{choices}
		
		\textbf{Solution}: According to the regulations of the law, one of the provisions that demonstrates the right to equality among ethnic groups in the economic aspect is that citizens belonging to ethnic groups are all entitled to: borrowing capital for production development. Choose \textcolor{red}{\textbf{A}}.
		
		\textbf{ChatGPT}: The correct answer is \textcolor{red}{\textbf{A}}. to borrow money for production development. According to Article 62 of the Constitution of the Socialist Republic of Vietnam in 2013, "Citizens belonging to different ethnic groups are equal before the law, with equal rights and duties; they enjoy full rights, preferential policies and protection of the State. The State ensures equal rights among ethnic groups in economic, social and cultural fields." Therefore, citizens belonging to different ethnic groups are entitled to borrow money for production development, which is one of the contents reflecting the equal rights among ethnic groups in economic field. Other options are not specified in the Constitution.
		
		\textbf{BingChat}: The correct answer is \textcolor{red}{\textbf{A}}. borrow capital for production development. According to the law, one of the contents that demonstrates the equality between ethnic groups in terms of economics is that citizens of all ethnic groups are entitled to borrow capital for production development. This means that citizens of all ethnic groups have the right to be supported in borrowing capital to develop business production and improve their economic life. The correct answer is A. Borrow capital for production development.

	\end{mdframed}	
	
	\subsubsection{Comprehension level question}
	\label{VNHSGE_Civ_C}
	
	Both ChatGPT and BingChat may have trouble giving precise responses when civic education questions become more complicated and necessitate the interpretation of fundamentals. However, ChatGPT and BingChat gave different responses in the specific case given where the question needed comprehension of market prices in the production and distribution of items. BingChat gave the right response, whereas ChatGPT gave the wrong one. The models' training on different datasets may have had an impact on how they understood and interpreted the concepts connected to market prices, which is why the responses varied.

	\begin{mdframed}[linewidth=1pt,linecolor=red] 
		\textbf{Question}: In the production and circulation of goods, when supply is less than demand, market prices tend to be higher.
		
		\begin{choices}
			\choice Individual labor waste.
			\choice Use value.
			\choice Commodity value.
			\choice Social consumption demand.
		\end{choices}
		
		\textbf{Solution}: In the production and circulation of goods, when supply is less than demand, market prices are usually higher than the value of the commodity.	Choose \textcolor{red}{\textbf{C}}.
		
		\textbf{ChatGPT}: \textcolor{red}{\textbf{D}}. Social consumption demand. Market prices depend on the balance between the supply and demand of goods. When social consumption demand is higher than the quantity of goods supplied, demand exceeds supply, leading to an increase in prices. Conversely, when the supply of goods exceeds social consumption demand, prices will decrease to stimulate consumption.
		
		\textbf{BingChat}: Answer: \textcolor{red}{\textbf{C}}. Value of goods
		
		In the production and circulation of goods, when supply is less than demand, market prices are usually higher than the value of goods in production. This means that when supply is less than demand, it shows that the demand for goods has increased, making the goods scarce and unable to meet the market demand, so prices rise. Therefore, the correct answer is C. Value of goods.

	\end{mdframed}	
	
	\subsubsection{Application level question}
	
	\label{VNHSGE_Civ_A}
	
	Both ChatGPT and BingChat may not be able to offer accurate responses when questions have a lot of disturbances, as shown by a question for which both models provided the erroneous response (D instead of B). Legal issues frequently entail interpreting the context and intent of laws and regulations as well as their detailed specifics. While these models are capable of delivering factual data and fundamental legal concepts, they may struggle to carry out in-depth legal analysis that necessitates proficiency in legal argumentation and reasoning.

	\begin{mdframed}[linewidth=1pt,linecolor=red] 
		\textbf{Question}: According to the regulations of the law, citizens are not allowed to exercise their right to vote for representatives of the People's Councils at all levels in the following cases: 
		
		\begin{choices}
			\choice Publicly accused.
			\choice Losing civil capacity.
			\choice Under administrative detention.
			\choice Temporarily detained for investigation.
		\end{choices}
		
		\textbf{Solution}: According to the regulations of the law, citizens are not allowed to exercise their right to vote for representatives of the People's Councils at all levels in the case of: losing civil capacity. Choose \textcolor{red}{\textbf{B}}.
		
		\textbf{ChatGPT}: Answer: \textcolor{red}{\textbf{D}}. Temporarily detained for investigation. Explanation: According to Article 19 of the Election Law, Vietnamese citizens from the age of 18 and above have the right to vote and be elected. However, if a citizen is temporarily detained for investigation, they are not allowed to exercise their right to vote for representatives of the People's Councils. After the temporary detention period ends, the citizen's right to vote for representatives of the People's Councils will be reinstated.
		
		\textbf{BingChat}: Answer: \textcolor{red}{\textbf{D}}. temporary detention for investigation purposes. Explanation: According to the law, citizens are not allowed to exercise their right to vote for representatives of the People’s Councils at all levels in case they are being temporarily detained for investigation purposes. This means that the citizen is not allowed to participate in voting in the election of representatives of the People’s Councils at all levels because they are being temporarily detained for investigation purposes. Therefore, the correct answer is D. Temporary detention for investigation purposes.
		
	\end{mdframed}	
	
	\subsubsection{High application level question}
	\label{VNHSGE_Civ_H}
	
	Due to the intricacy and nuance of legal studies, ChatGPT and BingChat might not be able to distinguish which character in case studies has committed an illegal act. Legal analysis frequently entails a thorough evaluation of a number of different variables, including the particular laws at issue, the parties' intentions, the situation's background, and the particulars of the case. While ChatGPT and BingChat are examples of machine learning models that can rapidly process and analyze massive volumes of data, they might not be able to provide the same level of nuanced interpretation and reasoning that a human legal expert can.  These models are only as good as their training data, so if they haven't been exposed to a diverse enough set of legal scenarios and cases, they might find it difficult to correctly identify the appropriate legal analysis in a particular instance. As a result, while ChatGPT and BingChat can offer valuable insights and fundamental legal facts, they might not be able to do so for complex legal questions without human aid.

	\begin{mdframed}[linewidth=1pt,linecolor=red] 
		\textbf{Question}: Agency X consists of Mr. G as the director, Ms. P as the deputy director, Ms. N as the head of the personnel organization department, Mr. K as the chairman of the labor union, and Mr. H as an employee. During the year-end review meeting, due to Ms. N's objection to her perspective on personnel transfers, Mr. G prevented her from speaking, insulted her, and forced her to leave the meeting. Sitting next to them, when Mr. H intended to express his agreement with Ms. N's opinion, Ms. P threatened to put him on the downsizing list to silence him. Upon hearing the conversation between Ms. P and Mr. H, and having a prior conflict with Mr. H, Mr. K took this opportunity to remove Mr. H from the year-end commendation list. Knowing the situation, Ms. V, Mr. H's wife, wrote a fabricated article distorting the agency X's personnel planning policy and spreading false information about Mr. K's personal life, which she publicly posted on social media, affecting Mr. K's reputation. Who among the following has violated both the rights protected by the law regarding honor and human dignity, as well as the freedom of speech of the citizens? 
		
		\begin{choices}
			\choice Mr. K, Ms. N and Mr. H
			\choice Mr. G and Ms. V
			\choice Mrs. P, Ms. V and Mr. G 
			\choice Ms. V and Mrs. P
		\end{choices}
		
		\textbf{Solution}: Mr. G and Ms. V have violated both the rights protected by the law regarding honor and human dignity, as well as the freedom of speech of the citizens. Choose \textcolor{red}{\textbf{B}}. 
		
		\textbf{ChatGPT}: \textcolor{red}{\textbf{D}}. Ms. V and Ms. P. In this case, Mr. G violated the freedom of speech of Ms. N by preventing her from speaking and forcing her to leave the meeting. Ms. P threatened to put Mr. H on the list of job cuts to silence him, which is also a violation of his freedom of speech. Ms. V violated the rights protected by the law regarding honor and human dignity of Mr. K by distorting the organizational policy and fabricating information about his personal life on social media.
		
		\textbf{BingChat}: Answer: \textcolor{red}{\textbf{C}}. Mrs. P, Ms. V and Mr. G. Explanation: C. Mrs. P, Ms. V and Mr. G have violated both the right to legal protection of honor and dignity and the right to freedom of speech of citizens. Mr. G prevented Ms. N from speaking and insulted her, forcing her to leave the meeting. This behavior violates Ms. N’s right to freedom of speech and the right to legal protection of honor and dignity. Mrs. P threatened to put Mr. H on the list of staff reductions to force him to stop speaking. This behavior violates Mr. H’s right to freedom of speech. Ms. V wrote an article distorting the personnel planning policy of agency X and fabricated information about Mr. K’s private life, then publicly posted it on social media, affecting Mr. K’s reputation. This behavior violates Mr. K’s right to legal protection of honor and dignity.
		
	\end{mdframed}

	\newpage
		
	\section{Vietnamese Student Score Spectrum} 
	\label{sec:Score Spectrum}
	
	We compare the performance of LLMs with the results achieved by students in each year due to the different structures and difficulty levels of the exams in each year. This allows us to evaluate the capability of LLMs compared to human-level performance
	
	The score spectrum of students is released annually by the Vietnamese Ministry of Education and Training, which we have collected for the years 2019-2022 (
	\href{https://moet.gov.vn/tintuc/Pages/tin-tong-hop.aspx?ItemID=6111}{2019},~
	\href{https://moet.gov.vn/tintuc/Pages/tin-tong-hop.aspx?ItemID=6879}{2020},~
	\href{https://moet.gov.vn/tintuc/Pages/tin-tong-hop.aspx?ItemID=7451}{2021},~
	\href{https://vietnamnet.vn/pho-diem-cac-mon-thi-tot-nghiep-thpt-2022-2042421.html}{2022}). For instance, Figure~\ref{fig:math_2022} shows the mathematics score spectrum of Vietnamese students in 2022. The result of the analysis of the score distribution of the 2022 high school graduation exam in Mathematics shows that there were 982,728 candidates who participated in the exam, with an average score of 6.47 points and a median score of 6.8 points. The most attained score was 7.8 points. There were 186,222 candidates scored below the average (accounting for 18.95\%).
	
	\subsection{Mathematics} 
	\label{sec:Score Spectrum_Math}
	Figures~\ref{fig:math_2022}-\ref{fig:math_2019} show the mathematics score spectrum of Vietnamese students in 2022-2019.
	
	\subsection{Literature}
	\label{sec:Score Spectrum_Lit}
	
	Figures~\ref{fig:lit_2022}-\ref{fig:lit_2019} show the literature score spectrum of Vietnamese students in 2022-2019.
	 
	\subsection{English}
	\label{sec:Score Spectrum_Eng}
	
	Figures~\ref{fig:English_2022}-\ref{fig:English_2019} show the English score spectrum of Vietnamese students in 2022-2019.

	\subsection{Physics}
	\label{sec:Score Spectrum_Phy}
	
	Figures~\ref{fig:phy_2022}-\ref{fig:phy_2019} show the physics score spectrum of Vietnamese students in 2022-2019.
	
	\subsection{Chemistry}
	\label{sec:Score Spectrum_Che}
	
	Figures~\ref{fig:Che_2022}-\ref{fig:Che_2019} show the chemistry score spectrum of Vietnamese students in 2022-2019.
	
	\subsection{Biology}
	\label{sec:Score Spectrum_Bio}
	
	Figures~\ref{fig:Bio_2022}-\ref{fig:Bio_2019} show the biology score spectrum of Vietnamese students in 2022-2019.
	
	\subsection{History}
	\label{sec:Score Spectrum_His}
	
	Figures~\ref{fig:His_2022}-\ref{fig:His_2019} show the history score spectrum of Vietnamese students in 2022-2019.
	
	\subsection{Geography}
	\label{sec:Score Spectrum_Geo}
	
	Figures~\ref{fig:Geo_2022}-\ref{fig:Geo_2019} show the geography score spectrum of Vietnamese students in 2022-2019.
	
	\subsection{Civic Education}
	\label{sec:Score Spectrum_Civ}
	
	Figures~\ref{fig:Civ_2022}-\ref{fig:Civ_2019} show the civic education score spectrum of Vietnamese students in 2022-2019.

	\begin{figure}[h!]
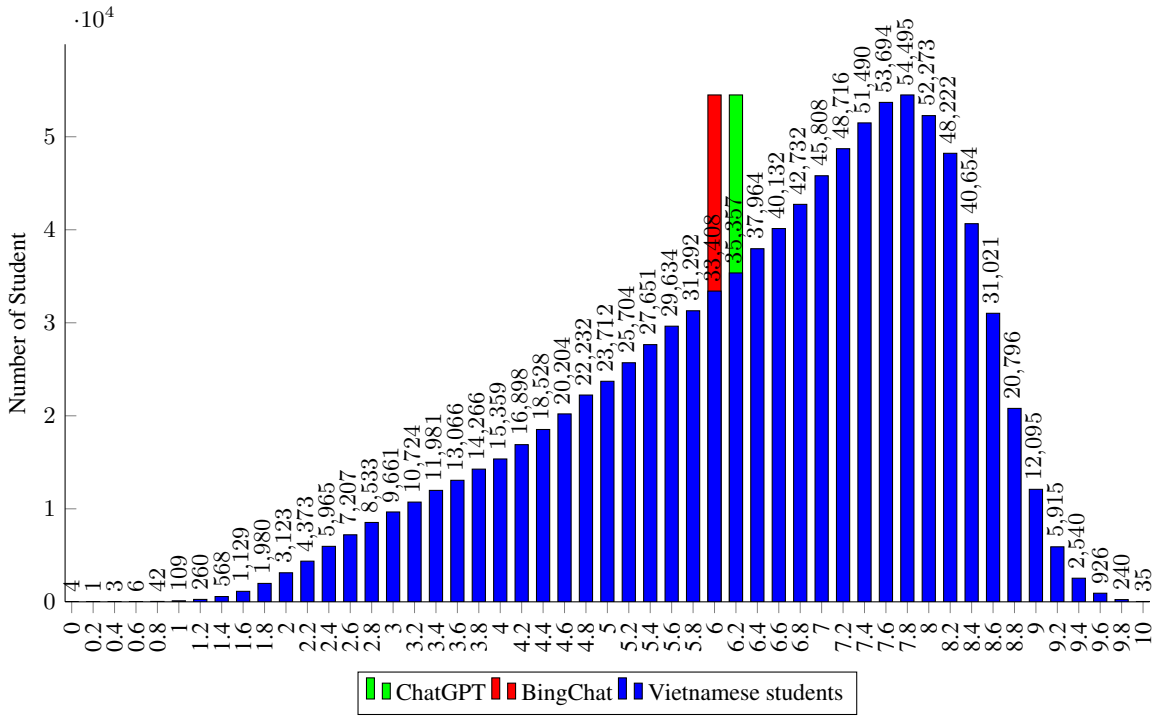

		\begin{center}	

		\end{center}
		\caption{Mathematics score spectrum of Vietnamese students in 2022.}
		\label{fig:math_2022}
	\end{figure}
	

	\begin{figure}[h!]
		\begin{center}
			%
		\end{center}
		\caption{Mathematics score spectrum of Vietnamese students in 2021.}
		\label{fig:math_2021}
	\end{figure}

	\begin{figure}[h!]
		\begin{center}
			%
		\end{center}
		\caption{Mathematics score spectrum of Vietnamese students in 2020.}
		\label{fig:math_2020}
	\end{figure}
	
	\begin{figure}[h!]
		\begin{center}
			%
		\end{center}
		\caption{Mathematics score spectrum of Vietnamese students in 2019.}
		\label{fig:math_2019}
	\end{figure}

	\begin{figure}[h!]
		\begin{center}
			%
		\end{center}
		\caption{Literature score spectrum of Vietnamese students in 2022.}
		\label{fig:lit_2022}
	\end{figure}
	
	\begin{figure}[h!]
		\begin{center}
			%
		\end{center}
		\caption{Literature score spectrum of Vietnamese students in 2021.}
		\label{fig:lit_2021}
	\end{figure}

	\begin{figure}[h!]
		\begin{center}
			%
		\end{center}
		\caption{Literature score spectrum of Vietnamese students in 2020.}
		\label{fig:lit_2020}
	\end{figure}
	
	\begin{figure}[h!]
		\begin{center}
			%
		\end{center}
		\caption{English score spectrum of Vietnamese students in 2022.}
		\label{fig:English_2022}
	\end{figure}
	
	\begin{figure}[h!]
		\begin{center}
			%
		\end{center}
		\caption{English score spectrum of Vietnamese students in 2021.}
		\label{fig:English_2021}
	\end{figure}
	
	\begin{figure}[h!]
		\begin{center}
			%
		\end{center}
		\caption{English score spectrum of Vietnamese students in 2020.}
		\label{fig:English_2020}
	\end{figure}
	
	\begin{figure}[h!]
		\begin{center}
			%
		\end{center}
		\caption{Physics score spectrum of Vietnamese students in 2022.}
		\label{fig:phy_2022}
	\end{figure}
	
	\begin{figure}[h!]
		\begin{center}
			%
		\end{center}
		\caption{Physics score spectrum of Vietnamese students in 2021.}
		\label{fig:phy_2021}
	\end{figure}
	
	\begin{figure}[h!]
		\begin{center}
			%
		\end{center}
		\caption{Physics score spectrum of Vietnamese students in 2020.}
		\label{fig:phy_2020}
	\end{figure}
	
	\begin{figure}[h!]
		\begin{center}
			%
		\end{center}
		\caption{Chemistry score spectrum of Vietnamese students in 2022.}
		\label{fig:Che_2022}
	\end{figure}
	
	\begin{figure}[h!]
		\begin{center}
			%
		\end{center}
		\caption{Chemistry score spectrum of Vietnamese students in 2021.}
		\label{fig:Che_2021}
	\end{figure}
	
	\begin{figure}[h!]
		\begin{center}
			%
		\end{center}
		\caption{Chemistry score spectrum of Vietnamese students in 2020.}
		\label{fig:Che_2020}
	\end{figure}
	
	\begin{figure}[h!]
		\begin{center}
			%
		\end{center}
		\caption{Biology score spectrum of Vietnamese students in 2022.}
		\label{fig:Bio_2022}
	\end{figure}
	
	\begin{figure}[h!]
		\begin{center}
			%
		\end{center}
		\caption{Biology score spectrum of Vietnamese students in 2021.}
		\label{fig:Bio_2021}
	\end{figure}
	
	\begin{figure}[h!]
		\begin{center}
			%
		\end{center}
		\caption{Biology score spectrum of Vietnamese students in 2020.}
		\label{fig:Bio_2020}
	\end{figure}
	
	\begin{figure}[h!]
		\begin{center}
			%
		\end{center}
		\caption{History score spectrum of Vietnamese students in 2022.}
		\label{fig:His_2022}
	\end{figure}
	
	\begin{figure}[h!]
		\begin{center}
			%
		\end{center}
		\caption{History score spectrum of Vietnamese students in 2021.}
		\label{fig:His_2021}
	\end{figure}
	
	\begin{figure}[h!]
		\begin{center}
			%
		\end{center}
		\caption{History score spectrum of Vietnamese students in 2020.}
		\label{fig:His_2020}
	\end{figure}
	
	\begin{figure}[h!]
		\begin{center}
			%
		\end{center}
		\caption{History score spectrum of Vietnamese students in 2019.}
		\label{fig:His_2019}
	\end{figure}
	
	\newpage

	\begin{figure}[h!]
		\begin{center}
			%
		\end{center}
		\caption{Geography score spectrum of Vietnamese students in 2022.}
		\label{fig:Geo_2022}
	\end{figure}
	
	\begin{figure}[h!]
		\begin{center}
			%
		\end{center}
		\caption{Geography score spectrum of Vietnamese students in 2021.}
		\label{fig:Geo_2021}
	\end{figure}
	
	\begin{figure}[h!]
		\begin{center}
			%
		\end{center}
		\caption{Geography score spectrum of Vietnamese students in 2020.}
		\label{fig:Geo_2020}
	\end{figure}
	
	\begin{figure}[h!]
		\begin{center}
			%
		\end{center}
		\caption{Civic education score spectrum of Vietnamese students in 2022.}
		\label{fig:Civ_2022}
	\end{figure}
	
	\begin{figure}[h!]
		\begin{center}
			%
		\end{center}
		\caption{Civic education score spectrum of Vietnamese students in 2021.}
		\label{fig:Civ_2021}
	\end{figure}
	
	\begin{figure}[h!]
		\begin{center}
			%
		\end{center}
		\caption{Civic education score spectrum of Vietnamese students in 2020.}
		\label{fig:Civ_2020}
	\end{figure}

	\begin{figure}[h!]
		\begin{center}
			%
		\end{center}
		\caption{Civic education score spectrum of Vietnamese students in 2019.}
		\label{fig:Civ_2019}
	\end{figure}
	
\end{document}